\documentclass[sn-basic,iicol]{sn-jnl}%

\usepackage{xspace}
\usepackage{eqnarray}
\usepackage{algorithm}
\usepackage{algorithmic}
\usepackage{booktabs}
\usepackage{mathtools}
\usepackage{graphics}
\usepackage{enumitem}
\usepackage{natbib}

\usepackage[caption=false]{subfig}
\newcommand{\M}[1]{\mathtt{#1}}
\newcommand{\V}[1]{\mathbf{#1}}

\newcommand{\comment}[1]{}

\newcommand{\gb}{Gr\"{o}bner basis }

\newcommand{\ie}{\textit{i}.\textit{e}.}
\newcommand{\eg}{\textit{e}.\textit{g}.}
\def\gf{Gr{\"o}bner fan\xspace}                                                                 

\jyear{2021}%

\theoremstyle{thmstyleone}%
\newtheorem{theorem}{Theorem}
\newtheorem{proposition}[theorem]{Proposition}%

\theoremstyle{thmstyletwo}%
\newtheorem{example}{Example}%

\theoremstyle{thmstylethree}%
\newtheorem{definition}{Definition}%

\numberwithin{theorem}{section}

\def\Hk{Heikkil{\"{a}}\xspace}  
\def\gb{Gr{\"o}bner basis\xspace}

\def\gbs{Gr{\"o}bner bases\xspace}

\def\rs{resultant\xspace}

\def\srs{sparse resultant\xspace}
\def\srss{sparse resultants\xspace}
\makeatletter
\renewcommand*\env@matrix[1][*\c@MaxMatrixCols c]{%
  \hskip -\arraycolsep
  \let\@ifnextchar\new@ifnextchar
  \array{#1}}
\makeatother
\newcommand{\mon}[1]{\V{x}^{\V{#1}}}

\newcommand{\Mres}{\M{M}(u_0)}
\newcommand{\coeff}[2]{c_{#1, \V{#2}}}
\newcommand{\F}{\mathcal{F}} 
\newcommand{\prfx}[1]{{}^{#1}} 

\begin{document}

\title[Article Title]{Sparse resultant based minimal solvers in computer vision and their connection with the action matrix}

\author*[1]{\fnm{Snehal} \sur{Bhayani}}\email{snehal.bhayani@oulu.fi}

\author[1]{\fnm{Janne} \sur{\Hk}}\email{janne.heikkila@oulu.fi}

\author[2]{\fnm{Zuzana} \sur{Kukelova}}\email{kukelova@cmp.felk.cvut.cz}

\affil[1]{\orgdiv{Center for Machine Vision and Signal Analysis}, \orgname{Faculty of Information Technology and Electrical Engineering}, \orgaddress{\city{Oulu}, \country{Finland}}}
\affil[2]{\orgdiv{Visual Recognition Group}, \orgdiv{Department of Cybernetics}, \orgname{Czech Technical University in Prague}, \orgaddress{\country{Czech Republic}}}

\abstract{Many computer vision applications require robust and efficient estimation of camera geometry from a minimal number of input data measurements, \textit{i.e.}, solving minimal problems in a RANSAC framework. Minimal problems are usually formulated as complex systems of sparse polynomial equations. The systems usually are overdetermined and consist of polynomials with algebraically constrained coefficients. Most state-of-the-art efficient polynomial solvers are based on the action matrix method that has been automated and highly optimized in recent years. On the other hand, the alternative theory of sparse resultants and Newton polytopes has been less successful for generating efficient solvers, primarily because the polytopes do not respect the constraints on the coefficients. To tackle this challenge, in this paper, we propose a simple iterative scheme to test various subsets of the Newton polytopes and search for the most efficient solver. Moreover, we propose to use an extra polynomial with a special form to further improve the solver efficiency via a Schur complement computation. We show that for some camera geometry problems our extra polynomial-based method leads to smaller and more stable solvers than the state-of-the-art \gb-based solvers. The proposed method can be fully automated and incorporated into existing tools for automatic generation of efficient polynomial solvers. It provides a competitive alternative to popular \gb-based methods for minimal problems in computer vision. Additionally, we study the conditions under which the minimal solvers generated by the state-of-the-art action matrix-based methods and the proposed extra polynomial resultant-based method, are equivalent. Specifically we consider a step-by-step comparison between the approaches based on the action matrix and the sparse resultant, followed by a set of substitutions, which would lead to equivalent minimal solvers.

}


\maketitle
\section{Introduction} 
The robust estimation of camera geometry, one of the most important tasks in computer vision, is usually based on solving so-called minimal problems~\citep{Nister-5pt-PAMI-2004,Kukelova-ECCV-2008, Kukelova-thesis}, \ie, problems that are solved from minimal samples of input data, inside a RANSAC framework~\citep{Fischler-Bolles-ACM-1981,Chum-2003, DBLP:journals/pami/RaguramCPMF13}. Since the camera geometry estimation has to be performed many times inside RANSAC~\citep{Fischler-Bolles-ACM-1981}, fast solvers to minimal problems are of high importance. Minimal problems often result in complex polynomial systems, in several variables. Further, the systems tend to be overdetermined and consist of polynomials whose coefficients are non-generic, \textit{i.e.,} algebraically constrained~\citep[p.~109]{Cox-Little-etal-05}.
A popular approach for solving minimal problems is to design procedures that can efficiently solve only a special class of systems of equations, e.g., systems resulting from the 5-pt relative pose problem~\citep{Nister-5pt-PAMI-2004}, and move as much computations as possible from the ``online'' stage of solving equations to an earlier pre-processing ``offline'' stage.

Most of the state-of-the-art specific minimal solvers are based on \gbs and the action matrix method~\citep{Cox-Little-etal-05,Stetter1996}. The \gb method was popularized in computer vision by Stewenius~\citep{DBLP:phd/basesearch/Stewenius05}. The first efficient \gb-based solvers were mostly handcrafted~\citep{Stewenius-ISPRS-2006,Stewenius-CVPR-2005} and sometimes very unstable~\citep{stewenius2005hard}. However, in the last $15$ years much effort has been put into making the process of constructing the solvers more automatic~\citep{Kukelova-ECCV-2008,larsson2017efficient,larsson2017polynomial} and the solvers stable~\citep{byrod2007improving,byrod2008column} and more efficient~\citep{larsson2017efficient,larsson2017polynomial,larsson2016uncovering,Bujnak-CVPR-2012,DBLP:conf/cvpr/LarssonOAWKP18,MartyushevVP2022}. There are now powerful tools available for the automatic generation of efficient \gb-based solvers \citep{Kukelova-ECCV-2008,larsson2017efficient}. 

Note that while all these methods are in the computer vision community known as \gb methods, most of them do not generate solvers that directly compute \gbs. They usually compute polynomials from a border basis~\citep{MourrainP2008} or go ``beyond'' Gr{\"o}bner and border bases~\citep{DBLP:conf/cvpr/LarssonOAWKP18}. However, all these methods are based on computing an action matrix\footnote{In the field of mathematics, an action matrix is also known as a multiplication matrix.}.
Therefore, for the sake of simplicity, in this paper, we will interchangeably write ``action matrix method'' and ``\gb method''  to refer to the same class of methods~\citep{Stetter1996,Kukelova-ECCV-2008,larsson2017efficient,larsson2017polynomial,larsson2016uncovering,Bujnak-CVPR-2012,DBLP:conf/cvpr/LarssonOAWKP18,MartyushevVP2022}. 

The first step in such action matrix methods is to compute a linear basis $B$ of the quotient ring $A = \mathbb{C}[X]/I$ where $I$ denotes the ideal generated by input polynomial system. An action matrix method based on the \gbs computes the basis of $A$ by defining the division w.r.t. some \gb of the ideal $I$. In general computing the \gb requires to \textit{fix} some monomial ordering. Alternatively, a heuristic basis sampling method~\citep{DBLP:conf/cvpr/LarssonOAWKP18} that goes beyond \gbs and monomial orderings in order to compute $B$ can be used. The basis sampling method is related to the border basis methods~\citep{MourrainP2008,MourrainSM2021}, which generalize the concept of monomial ordering and even propose a non-monomial basis of $A$. The second step in these action matrix methods is to compute a linear map $T_f$, representing the multiplication in $A$, w.r.t. some polynomial $f$.
The representative matrix $\M{M}_f$ of the map $T_f$ is what we call the action matrix. Recently,~\citep{TelenM2018,TelenMB2018} generalized the method used for generating the basis $B$. The proposed methods are known as normal form methods, developed for general systems of polynomial equations. They are not tailored to problems that appear in computer vision.

While the action matrix method for generating efficient minimal solvers has been thoroughly studied in computer vision and all such recently generated solvers are highly optimized in terms of efficiency and stability, little attention has been paid to an alternative algebraic method for solving systems of polynomial equations, i.e. the \rs method~\citep[Chapter~3]{Cox-Little-etal-05}. The \rs method was manually applied to several computer vision problems~\citep{Kukelova-PolyEig-PAMI-2012,Hartley-PAMI-2012,DBLP:conf/wacv/KastenGB19,Kukelova-thesis,Kukelova-PolyEig-PAMI-2012}. However, in contrast to the action matrix method, there is no method for automatically generating \srs-based minimal solvers. The most relevant works in this direction are the methods using the subdivision of the Minkowski sum of all the Newton polytopes~\citep{emiris-general,DBLP:journals/jacm/CannyE00,Checa2022MixedSS} and the iterative method based on the Minkowski sum of a subset of Newton polytopes~\citep{DBLP:conf/iccv/Heikkila17}. However, the subdivision methods are not applicable to polynomial systems with non-generic coefficients, whereas the method in~\citep{DBLP:conf/iccv/Heikkila17} leads (due to linearizations) to larger and less efficient solvers than the action matrix solvers.

Our first objective in this paper is to study the theory of \srss and propose an approach for generating efficient solvers for solving camera geometry problems, by using an extra polynomial with a special form.
Our approach, based on the Newton polytopes and their Minkowski sums, is  used to generate \srs matrices, followed by a Schur complement computation, and a conversion of the resultant constraint into an eigenvalue problem.

{Note that our approach differs from the subdivision methods~\citep{DBLP:journals/jacm/CannyE00,emiris-general} in how the Newton polytopes are used for generating minimal solvers. The subdivision methods compute the Minkowski sum of the Newton polytopes of all the polynomials and divide its lattice interior into cells, which are used to construct the \srs matrices. Whereas in this paper, we make this process iterative, computing the Minkowski sum of each subset of Newton polytopes (of the input polynomials) and avoid computing a subdivision. We directly use the lattice interior to generate a candidate \srs matrix. For systems with non-generic coefficients, such an iterative approach has been effective in generating \srs matrices for the minimal problems in computer vision.} The \srs matrix and the eigenvalue problem together represent our minimal solver. The solvers based on our proposed approach have achieved significant improvements over the state-of-the-art \srs-based solvers~\citep{DBLP:conf/iccv/Heikkila17} and achieved comparable or better efficiency and/or accuracy to state-of-the-art \gb-based solvers. Moreover, our proposed approach has been fully automated, and can be incorporated in the existing tools for automatic generation of efficient minimal solvers~\citep{Kukelova-ECCV-2008,larsson2017efficient,DBLP:conf/cvpr/LarssonOAWKP18}.

There is a similarity between the solvers obtained using our \srs-based approach and the solvers based on the action matrix methods~\citep{Kukelova-ECCV-2008,larsson2017efficient,DBLP:conf/cvpr/LarssonOAWKP18,Stetter1996,byrod-etal-ijcv-2009,MartyushevVP2022,martyushev2023automatic}. 
Therefore the second objective in this paper is to investigate this similarity. In a step-by-step manner we demonstrate that for a minimal solver generated based on the action matrix method, we can change the steps performed by the \srs method such that it leads to an equivalent solver. Similarly, we also demonstrate that for a given minimal solver generated based on the \srs method, we can change the steps performed by the action matrix method such that it leads to an equivalent solver.
Specifically, our contributions are:
\begin{enumerate}
    \item A novel approach (Sec.~\ref{sec:extra_poly_sparse_res}), of adding an extra polynomial with a special form for generating a \srs matrix whose \srs constraint can be decomposed (Sec.~\ref{subsec:block_partn}) into an eigenvalue problem.
        \begin{itemize}
        \item A scheme (Sec.~\ref{subsec:basis_using_polytopes}) to iteratively test the Minkowski sum of the Newton polytopes of each subset of polynomials, searching for the most efficient minimal solver, in the presence of algebraically constrained coefficients.
        \item Two procedures (Sec.~\ref{subsec:col_red}) to reduce the \srs matrix size, leading to comparable or better solver speeds than those generated by many state-of-the-art \gb-based methods.
        \item A general method for automatic generation of efficient minimal solvers.
        The automatic generator is publicly available at~\citep{genautsparseresultantsolver}.
    \end{itemize}
    \item A study of the constraints (Sec.~\ref{sec:am_vs_res}) to be satisfied so that the solvers based on our \srs method as well as the action matrix method are equivalent, \ie~they involve eigenvalue decomposition of exactly the same matrix, and the steps performed for constructing that matrix can be interchanged.
\end{enumerate}   

This paper is an extension of our work~\citep{BhayaniKH20}, where we proposed an extra polynomial \srs method  for generating efficient minimal solvers.

\section{Theoretical background and related work}\label{sec:theory}
\noindent In this section, we summarize the basic concepts and notation used in the paper. This notation is based  on the book by~\cite{Cox-Little-etal-05}, to which we refer the reader for more details. 

Our goal is to compute the solutions to a system of $m$ polynomial equations,
\begin{equation}\label{eq:eq_system}
 f_1(x_1,\dots,x_n) = 0,\dots, f_m(x_1,\dots,x_n)=0,
\end{equation}  in $n$ variables, $X = \lbrace x_1,\dots,x_n \rbrace$. Let us denote the set of polynomials from~\eqref{eq:eq_system} as $\F=\{f_1,\dots,f_m\}$.
The variables in $X$ can be ordered and formed into a vector. W.l.o.g. let us assume that this vector has the form, $\V{x} =  \begin{bmatrix}
  x_1 & \dots & x_n
\end{bmatrix}^\top$.
Then for a vector $\V{\alpha} =\begin{bmatrix}
   \alpha_1 & \dots & \alpha_n
\end{bmatrix}^\top \in \mathbb{N}^{n}$, a monomial $\V{x}^\V{\alpha}$ is a product  $\V{x}^\V{\alpha} = \prod_{i=1}^{n} x_i^{\alpha_i}$. Each polynomial $f_i \in \F$ is expressed as a linear combination of a finite set of monomials
\begin{equation}\label{eq:poly_expr}
    f_{i}(\V{x}) = \underset{\V{\alpha} \in \mathbb{N}^{n}}{\Sigma} c_{i,\V{\alpha}} \mon{\alpha}.
\end{equation}
We collect all monomials present in $\F$ and denote the set as $\text{mon}(\F)$. Let $\mathbb{C}[X]$ denote the set of all polynomials in unknowns $X$ with coefficients in $\mathbb{C}$. The ideal $I = \langle f_1,\dots,f_m \rangle \subset \mathbb{C}[X]$ is the set of all polynomial combinations of  generators $f_1,\dots,f_m$. The set $V$ of all solutions of the system~\eqref{eq:eq_system}
is called the affine variety. Each polynomial $f \in I$ vanishes on the solutions of \eqref{eq:eq_system}. Here we assume that the ideal $I$  generates a zero dimensional variety, \ie, the system~\eqref{eq:eq_system} has a finite number of solutions (say $r$). Using the ideal $I$ we can define the quotient ring $A = \mathbb{C}[X]/I$ which is the set of equivalence classes, denoted over $\mathbb{C}[X]$ defined by the relation $a \sim b 
\iff (a - b) \in I$. 
If $I$ has a zero-dimensional variety then the quotient ring $A$ is a finite-dimensional vector space over $\mathbb{C}$~\citep{Cox-Little-etal-05}. For an ideal $I$ there exist special sets of generators called \gbs which have the nice property that the remainder after division is unique. Using a \gb we can define a linear basis for the quotient ring $A$.\\

\noindent \textbf{Non-generic polynomial system: } One of the important property of the polynomial systems in computer vision is that of non-genericity of its coefficients. The genericity of a property of a system is formally defined in~\citep[p.~109]{Cox-Little-etal-05}, which we adapt here to define \textit{non-genericity} as,
\begin{definition}\label{def:non_genericity}
    The coefficients $\lbrace c_{i,\V{\alpha}} \rbrace$ of the polynomial system $\F$ in~\eqref{eq:poly_expr}, are non-generic, if and only if there exists a non-vanishing polynomial in $\lbrace c_{i,\V{\alpha}} \rbrace $ as variables.  
\end{definition}

\noindent \textbf{Matrix form: } In this paper we will need to express a given set of polynomials $\F$ via matrix multiplication. To achieve this we fix an order for the polynomials in $\F$, and also do the same for the monomials in $B = \text{mon}(\F)$.
The matrix representation of $\F$ has the form
\begin{equation}\label{eq:matrix_form}
    \M{C}([\coeff{i}{\alpha}]) \ \V{b},
\end{equation}
where $\V{b}$ is a  column vector of the monomials in $B$ ordered w.r.t. the given monomial ordering and $\M{C}([\coeff{i}{\alpha}])$ is the matrix of coefficients $\coeff{i}{\alpha}$ of $\F$. The entries of $ \M{C}([\coeff{i}{\alpha}])$ are row-indexed by the ordered set of polynomials in $\F$ and column-indexed by the monomials in $\V{b}$.\\

\noindent \textbf{Monomial multiples of a polynomial set: } 
We begin with a polynomial system $\F$ and the set of monomials in $\F$, \ie, $B = \text{mon}(\F)$. Let $B^\prime \supset B$ be another set of monomials. 

Then an important step in our proposed method is to extend $\F$ to the largest possible set, say $\F^\prime$, such that $\text{mon}(\F^\prime) \subseteq  B^\prime$. This extension is done by multiplying each $f_i \in \F$ with monomials. \ie, we are generating polynomials from $I = \langle f_1,\dots, f_m \rangle$. For this, we compute the set $T_i$ of all possible monomials $T_i = \lbrace \mon{\alpha} \rbrace$ for each $f_i \in \F$, such that $\text{mon}(\mon{\V{\alpha}} f_i) \subset B^\prime, \  \mon{\alpha} \in T_i $. We will use the following shorthand notation to express such an operation
\begin{equation}\label{eq:poly_extn}
    \F \overset{B^\prime}{\rightarrow} (\F^\prime, T),
\end{equation}
where $T = \lbrace T_1,\dots,T_m \rbrace$.
Subsequently, in this operation we assume to have removed all monomials in $B^\prime$ which are not present in the extended set $\F^\prime$ and denote the modified set as $B^\prime$ by abuse of notation. In other words, we will assume that $B^\prime = \text{mon}(\F^\prime)$.

\subsection{The action matrix method}\label{subsec:action_matrix_outline}
\noindent One of the well known SOTA methods for polynomial solving is the \gb-based action matrix method~\citep{Cox-Little-etal-05,Stetter1996,Sturmfels-CBMS-2002}. It has been recently used to efficiently solve many minimal problems in computer vision~\citep{Kukelova-ECCV-2008,Kukelova-thesis,Stetter1996,larsson2017efficient,byrod2007improving,byrod-etal-ijcv-2009,DBLP:conf/cvpr/LarssonOAWKP18}. It transforms the problem of finding the solutions to~\eqref{eq:eq_system}, to a problem of eigendecomposition of a special matrix. We list the steps performed by an action matrix method in Appendix~\ref{app:am_steps}, and note here that the algorithm can essentially be distilled in a sequence of three important steps, viz. construct the set $T_j$ of monomial multiples for each of the input polynomials $f_j \in \F$, extend $\F$ via monomial multiplication to $\F^{\prime}$ and linearize the resulting system as a matrix product, $\M{C} \V{b}$. A G-J elimination of $\M{C}$ is then used to extract the required action matrix, whose eigenvalues give us the roots of $\F$.

\subsection{Resultants} 
Alternative to the action matrix method, we have the method of resultants for polynomial solving. Originally, resultants were used to determine whether a system of $n+1$ polynomial equations in $n$ unknowns has a common root or not. Let us have a system of polynomial equations as defined in~\eqref{eq:eq_system} and assume that $m=n+1$. Classically, a resultant is defined to be an irreducible polynomial, constraining the coefficients of the polynomials in $\F$ (see Eq.~\eqref{eq:poly_expr}) to have a non-trivial solution. Then the resultant is a polynomial $Res([c_{i,\V{\alpha}}])$ with $\coeff{i}{\alpha}$ as variables. A more formal treatment of the theory of resultants can be obtained from~\citep[Chapter~3]{Cox-Little-etal-05}.

\subsubsection{Polynomial solving using resultants}\label{subsubsec:poly_sol_using_res}
A step in a \srs method is to expand the given system of polynomials $\F$ to a set of linearly independent polynomials. This is usually done by adding some monomial multiples of the original polynomials, \ie, using the operation~\eqref{eq:poly_extn}. The expanded set of polynomials, say $\F^\prime$, can be expressed in a matrix form as
\begin{equation}
    \M{C}([\coeff{i}{\alpha}]) \ \V{b}.
    \label{eq:Cb=0}
\end{equation} 
Usually,  $\M{C}([\coeff{i}{\alpha}])$ in~\eqref{eq:Cb=0} is a rectangular matrix (tall or wide). 

The resultant-based method here, requires the coefficient matrix $\M{C}([\coeff{i}{\alpha}])$ to be a square invertible matrix for randomly assigned values to the coefficients, $\coeff{i}{\alpha} \in \mathbb{C}_{\neq 0}$
\ie, $\det \M{C}([\coeff{i}{\alpha}]) \neq 0$. A matrix with such properties is called the \textit{\rs matrix} and in this paper we will denote it as $\M{M}([\coeff{i}{\alpha}])$\footnote{E.g. if $\M{C}([\coeff{i}{\alpha}])$ in~\eqref{eq:Cb=0} is a tall matrix, i.e. matrix with more rows than columns, with full column rank, then $\M{M}([\coeff{i}{\alpha}])$ can be constructed as a full rank square submatrix of $\M{C}([\coeff{i}{\alpha}])$}. If $\M{C}([\coeff{i}{\alpha}])$ is a square invertible matrix, we can rewrite~\eqref{eq:Cb=0} as
\begin{equation}\label{eq:res_mat_form}
    \M{M}([\coeff{i}{\alpha}]) \ \V{b}.
\end{equation}
Now, $\F = 0$ implies $\F^\prime = 0$ and it leads to
\begin{equation}\label{eq:res_mat_sing}
    \M{M}([\coeff{i}{\alpha}]) \ \V{b} = \V{0}.
\end{equation} Thus the requirement for $\F=0$ to have common solutions is the following condition on the coefficients of $\F$,
\begin{equation}\label{eq:res_cnst}
    \det \M{M}([\coeff{i}{\alpha}]) = 0. 
\end{equation} We call this the \textit{resultant constraint}. It is a polynomial with $\coeff{i}{\alpha}$ as variables. 
From the definition of a resultant $Res([\coeff{i}{\alpha}])$~\citep[Theorem~2.3]{Cox-Little-etal-05} we have that $Res([\coeff{i}{\alpha}])$ is a polynomial with $\coeff{i}{\alpha}$ as variables and that it vanishes iff the system of equations $\F=0$ has common roots. This gives us the necessary condition for the existence of roots the system $\F=0$, that $\det \M{M}([\coeff{i}{\alpha}])$ must vanish if the resultant vanishes, \ie,
\begin{equation}\label{eq:res_mat}
    Res([\coeff{i}{\alpha}]) = 0 \implies \det \M{M}([\coeff{i}{\alpha}]) = 0.
\end{equation}
This implies that given a polynomial system $\F$, the resultant constraint~\eqref{eq:res_cnst} is a non-trivial multiple of its resultant $Res([\coeff{i}{\alpha}])$.\\

While resultants are defined for a system of one more polynomial than the number of variables, we can employ them for solving a system of $n$ polynomials in $n$ variables. One way to do this, is to hide one of the variables to the coefficient field (in other words, consider it to be a constant), another way is to add an extra polynomial by introducing a new variable, and then hide this variable to the coefficient field. In both these approaches, we end up with a system where we have one more polynomial than the number of variables. 

\subsubsection{Hiding a variable}\label{subsubsec:hidden_var_intro}
By \textit{hiding} one variable, say $x_{n}$, to the coefficient field, we are left with $n$ polynomials $\F$ in $n - 1$ variables. This gives us a way to use the concept of resultants and compute $Res([\coeff{i}{\alpha}], x_{n})$ which now becomes a function of $\coeff{i}{\alpha}$ and $x_{n}$. In this case,~\eqref{eq:res_mat_form} becomes 
\begin{equation} \label{eq:hid_var_sp_res_const}
    \M{M}([\coeff{i}{\alpha}],x_{n}) \V{b},
\end{equation}
where the symbols have their usual meaning. For simplicity we will denote $\M{M}([\coeff{i}{\alpha}],x_{n})$  as $\M{M}(x_{n})$ in the rest of this paper. Its determinant $\det \M{M}(x_{n})$, is a multiple of the resultant $Res(x_n)$. This is known as a \textit{hidden variable resultant} and it is a polynomial in $x_n$ whose roots are the $x_{n}$-coordinates of the solutions of the system of polynomial equations. For theoretical details and proofs we refer to \citep[Chapter~7]{Cox-Little-etal-05}. Such a hidden variable approach has been used in the past to solve various minimal problems~\citep{Kukelova-thesis,Kukelova-PolyEig-PAMI-2012,Hartley-PAMI-2012,DBLP:conf/wacv/KastenGB19}. 

This approach leads to computing the roots of the polynomial, $\det \M{M}(x_{n}) = 0 $. However, computing the determinant of a polynomial matrix $\det \M{M}(x_n)$ and then its roots, may be slow and unstable. Therefore, the most common way to solve the original system of polynomial equations is to first transform the following matrix equation
\begin{equation}
\M{M}(x_{n}) \V{b} = \V{0},
\end{equation}
to a polynomial eigenvalue problem (PEP)~\citep{Cox-IVA-2015}, which is then expressed as,
\begin{equation}\label{eq:pepformulation}
(\M{M}_{0} + \M{M}_{1} \ x_{n}+...+ \M{M}_{l} \ x^{l}_{n})\V{b} = \V{0},
\end{equation}
where $l$ is the degree of the matrix $\M{M}(x_{n})$ in the hidden variable $x_n$ and $\M{M}_{0},\dots,\M{M}_{l}$ are matrices that depend only on the coefficients $\coeff{i}{\alpha}$ of the original system of polynomials. The PEP~\eqref{eq:pepformulation} can be easily converted to a generalized eigenvalue problem (GEP), written as,
\begin{equation}\label{eq:GEP}
\M{A} \V{y} =  x_{n} \M{B} \V{y},
\end{equation}
and solved using standard efficient eigenvalue algorithms~\citep{Kukelova-PolyEig-PAMI-2012}. Basically, the eigenvalues give us the solutions to $x_{n}$ and the rest of the variables can be extracted from the corresponding eigenvectors, $\V{y}$. Such a transformation to a GEP relaxes the original problem of finding the solutions to the input polynomial system since the eigenvectors in general do not satisfy the monomial dependencies induced by the monomial vector $\V{b}$ as well the monomial vector $\V{y}$. Moreover, this relaxation may also introduce extra parasitic (zero) eigenvalues leading to slower polynomial solvers.

\subsubsection{Adding an extra polynomial}\label{subsubsec:u_res_intro}
Alternatively, we can add a new polynomial 
\begin{equation}
f_{n+1} = u_0 + u_1 x_1 + \dots + u_n x_n
\label{eq:u-res_eq}
\end{equation} to $\F$ and compute the so called \textit{$u$-resultant} (see~\citep{waerden_modern_1950} and \citep[Chapter~3]{Cox-Little-etal-05}) by hiding $u_0,\dots,u_n$. In general, random values are assigned to $u_1,\dots,u_n$ (and $u_0$ is a new unknown). Just like in the hidden variable method, this method hides the variable $u_0$ to the coefficient field and generates the resultant matrix $\Mres$, \ie, in this case the matrix form of the extended system of polynomials~\eqref{eq:res_mat_form} has the following form
    \begin{equation} 
        \M{M}([\coeff{i}{\alpha}], u_0) \ \V{b},
        \label{eq:u-res_Mb}
    \end{equation} 
    where $u_0$ is a new unknown variable and $\V{b}$ is a vector of all monomials in this extended system. We will denote $\M{M}([\coeff{i}{\alpha}], u_0)$ as $\Mres$ in order to simplify the notation. Again, it holds that $\text{det} \ \Mres$ is a multiple of the resultant $Res(u_0)$. Thus, in order to compute the common zeros of $\F=0$, we need to solve $\text{det} \Mres =0$, similar to the case of the hidden variable resultant. 
    Instead of computing the determinant of a polynomial matrix, here we also solve it as a GEP described in Sec.~\ref{subsubsec:hidden_var_intro}. However, as mentioned in the hidden variable method, such a method introduces spurious solutions, arising from the linearization of the original system in~\eqref{eq:u-res_Mb}, \ie, not considering monomial dependencies in $\V b$, as well as from transforming PEP to GEP and not considering the monomial dependencies in $\V y$ in~\eqref{eq:GEP}. Some of these spurious solutions (eigenvalues) can be removed by using Schur complement, described in the section below or by applying the method for removing zero eigenvalues, similar to~\citep{Kukelova-thesis,BhayaniKHHiddenVarRes}.

\subsubsection{Schur complement}\label{subsec:schur_complement}
\noindent One way to remove the spurious solutions introduced by the linearization of a hidden variable resultant matrix $\Mres$ is to use the Schur complement of one its submatrices. Here, we briefly review this method. Let us first consider \textit{some} partition of the set of monomials $B$ as
\begin{equationarray}{l}\label{eq:B_partn}
    B = B_1 \sqcup B_2.
\end{equationarray} Note that $\V{b} = \text{vec}(B) = \begin{bmatrix}
   \V{b}_1 & \V{b}_2
\end{bmatrix}^\top$, where $\V{b}_1=\text{vec}(B_1)$ and $\V{b}_2=\text{vec}(B_2)$.  This imposes a column partition on $\Mres$ in~\eqref{eq:u-res_Mb}. Morever, we can order the rows of $\Mres$ such that its upper block is independent of $u_0$. Together, we obtain the following block partition of $\Mres$:
\begin{equation}\label{eq:M_block_partn}
    \Mres = \begin{bmatrix}
       \M{M}_{11} & \M{M}_{12} \\ \M{M}_{21}(u_0) & \M{M}_{22}(u_0)
    \end{bmatrix}.
\end{equation}
Here, the upper block $\begin{bmatrix}
   \M{M}_{11} & \M{M}_{12}
\end{bmatrix}$ is independent of $u_0$. 
Thus we can write~\eqref{eq:u-res_Mb} as
\begin{equationarray}{l}\label{eq:lambda_res_mat_const_block_partn}
   \begin{bmatrix}
       \M{M}_{11} & \M{M}_{12} \\ \M{M}_{21}(u_0) & \M{M}_{22}(u_0)
    \end{bmatrix} \ \begin{bmatrix}
   \V{b}_1 \\ \V{b}_2
\end{bmatrix}.
\end{equationarray}
The requirement for existence of solutions of $\F = 0$ is that the vector in~\eqref{eq:lambda_res_mat_const_block_partn} should vanish. We thus obtain the following two vector equations
\begin{equationarray}{l}
    \M{M}_{11} \V{b}_{1} +  \M{M}_{12} \V{b}_{2} = \V{0} \\
    \M{M}_{21}(u_0) \V{b}_{1} +  \M{M}_{22}(u_0) \V{b}_{2} = \V{0}.
\end{equationarray}
If we were able to partition $\Mres$ such that $\M{M}_{12}$ is a square invertible matrix, we can eliminate $\V{b}_2$ from these two equations, to obtain
\begin{equationarray}{l}\label{eq:res_schur_complement}
   (\underbrace{\M{M}_{21}(u_0) -  \M{M}_{22}(u_0) \M{M}^{-1}_{12} \M{M}_{11}}_{\M{X}(u_0)}) \V{b}_{1} = \V{0}.
\end{equationarray}
The matrix $\M{X}(u_0)$ is the Schur complement of the block $\M{M}_{12}$ of $\Mres$, which has the following property:
\begin{equationarray}{l}
    \text{det} (\Mres) = \text{det}(\M{A}_{12}) \ \text{det}(\M{X}),
\end{equationarray} where, $\text{det}(\M{A}_{12}) \neq 0$ by our assumption. Therefore, for a generic value of $u_0$, $\text{det}(\Mres) \neq 0 \Leftrightarrow \text{det}(\M{X}(u_0)) \neq 0$, which means that $\M{X}(u_0)$ is a square invertible matrix and its determinant is also a multiple of the resultant. However, $\M{X}(u_0)$  corresponds to a smaller eigenvalue problem as compared to the the sparse resultant matrix $\Mres$.

Both the \rs based methods, described in Secs.~\ref{subsubsec:hidden_var_intro} and~\ref{subsubsec:u_res_intro} are proposed for square generic systems~\citep{emiris-general} via mixed subdivision of polytopes. However, the polynomial systems studied here, are usually sparse, overdetermined and consist of polynomials with non-generic coefficients (see Def.~\ref{def:non_genericity}). The sparsity of the systems can be exploited to obtain more compact resultants using specialized algorithms. Such resultants are commonly referred to as the \textit{sparse resultants}.

\begin{figure*}[t]
\centering
\subfloat[]{\includegraphics[width=0.32\linewidth]{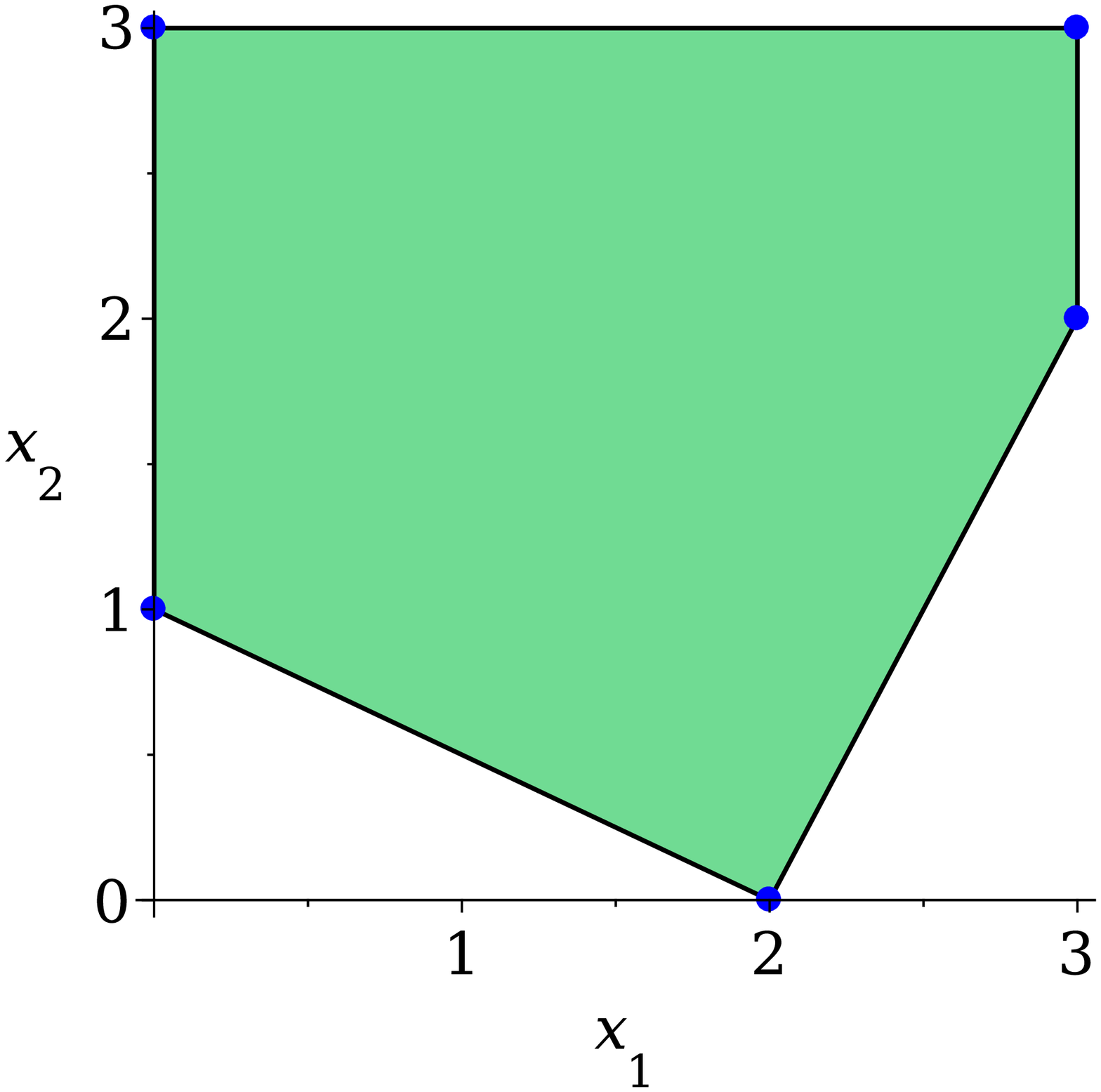}}
\subfloat[]{\includegraphics[width=0.32\linewidth]{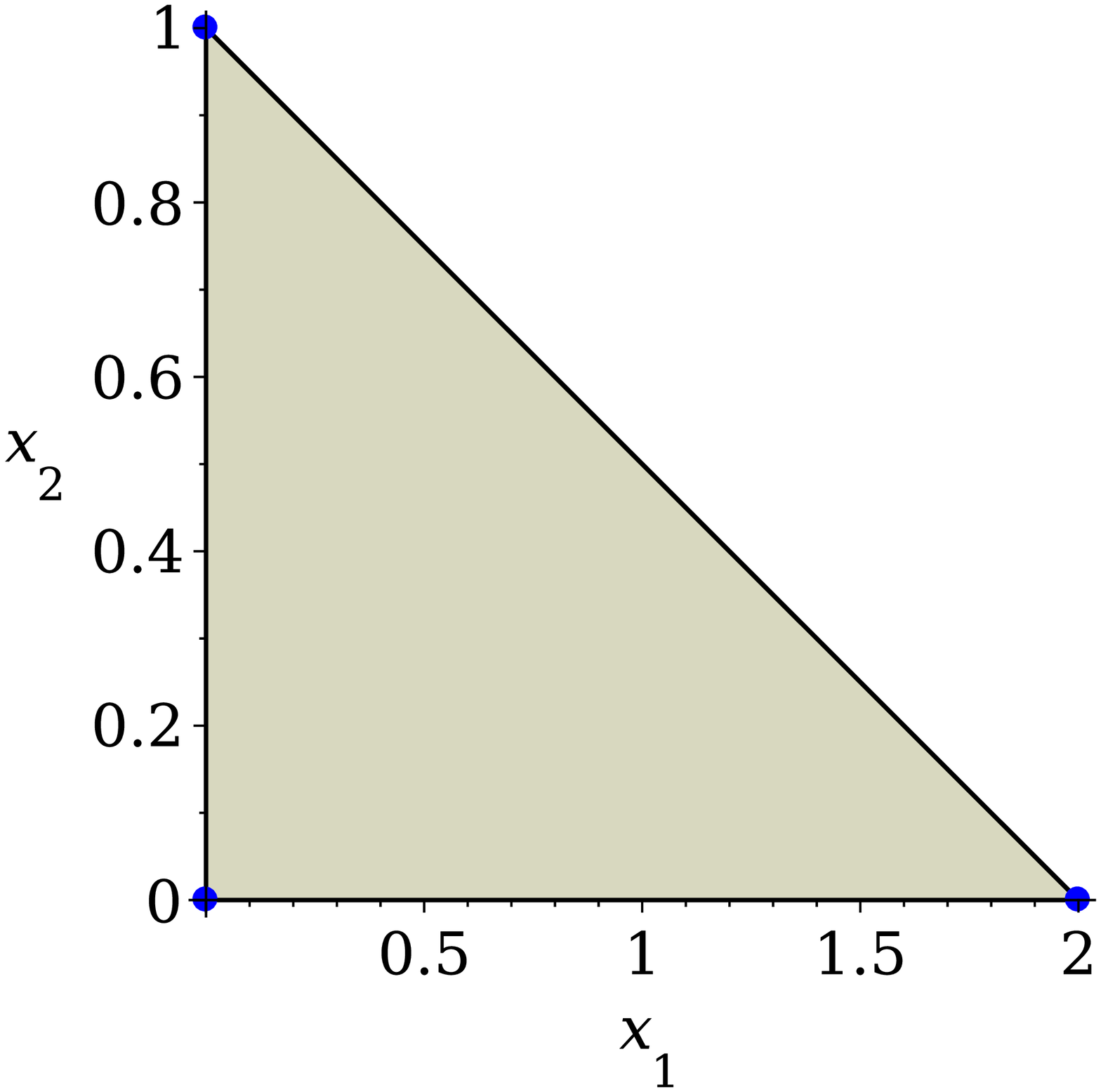}}
\subfloat[]{\includegraphics[width=0.32\linewidth]{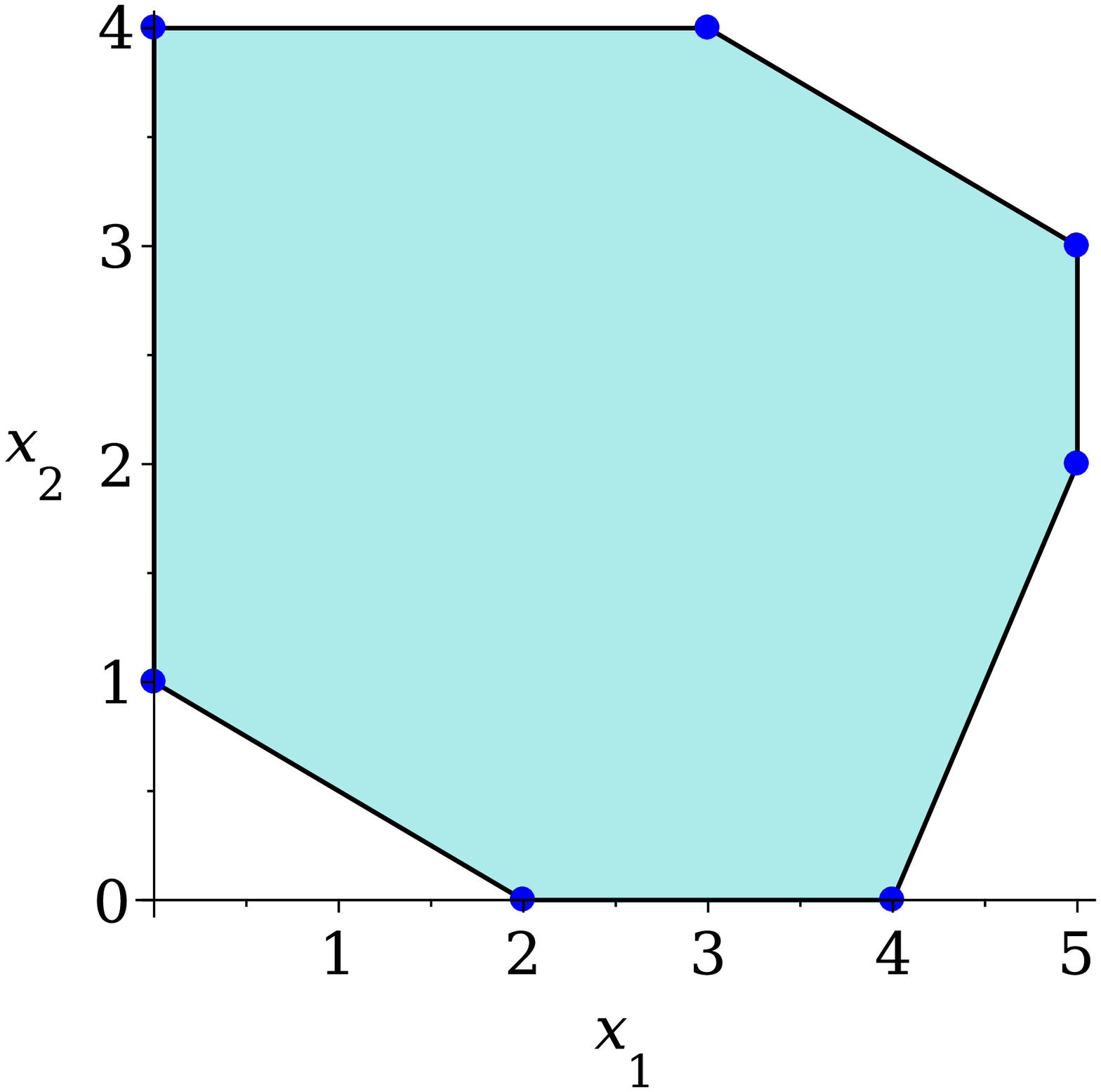}}
\caption{An example of the Newton polytopes of two polynomials as well as their Minkowski sum: \textbf{a.} $P_1 = \text{NP}(f_1)$ \textbf{b.} $P_2 = \text{NP}(f_2)$ and \textbf{c.} $Q = P_1 + P_2$}
\label{fig:poly_example_1}
\end{figure*}

\subsubsection{Polyhedral geometry}\label{subsec:polytope_terminology} 
\noindent Sparse resultants are studied via the theory of polytopes~\citep[Chapter~7]{Cox-Little-etal-05}. Therefore, we define the important terms and notations related to polytopes, which we will later use in the text. 

The \textit{Newton polytope} $\text{NP}(f)$, of a polynomial $f$ is defined as a convex hull of the exponent vectors of all the monomials occurring in the polynomial (also known as the support of the polynomial). Therefore we have $\text{NP}(f_{i}) = \text{Conv}(A_{i})$ where $A_{i}$ is the set of all integer vectors that are exponents of monomials with non-zero coefficients in $f_{i}$. A \textit{Minkowski sum} of any two convex polytopes $P_1, P_2$ is defined as $P_1+P_2 = \lbrace p_1 + p_2 \ | \ \forall p_1 \in P_1, \forall p_2 \in P_2 \rbrace$. We demonstrate the concept of a Newton polytope using a simple example.
\begin{example}\label{ex:ex_1}
 Let us consider a system of two polynomials $\F = \lbrace f_1(\V{x}), f_2(\V{x}) \rbrace$, in two variables $X = \lbrace x_1, x_2 \rbrace$
\begin{equationarray}{rl}
    f_1 =& {c_{1,1}} {x_1}^{3} {x_{2}}^{3} + {c_{1,2}} {x_{1}}^{2} {x_{2}}^{3}  +  {c_{1,3}} {x_{1}}^{3}  {x_{2}}^{2}+ \nonumber \\
    & {c_{1,4}} {x_{1}}^{2} {x_{2}}^{2}  +  {c_{1,5}} {x_{2}}^{3} +{c_{1,6}} {x_{1}}^{2} {x_{2}} +\nonumber \\
    & {c_{1,7}} {x_{2}}^{2}  + {c_{1,8}} {x_{1}} {x_{2}}  +{c_{1,9}} {x_{1}}^{2}  +{c_{1,10}}  {x_{2}} \\
    f_2 =& {c_{2,1}} {x_{1}}^{2}  +{c_{2,2}} {x_{2}}  +{c_{2,3}} {x_{1}}  + {c_{2,4}}.
 \end{equationarray}
 Here, $\lbrace c_{1,1},\dots, c_{1,10},c_{2,1},\dots,c_{2,4} \rbrace \subset \mathbb{C} $ is the set of coefficients of $\F$. The supports of $\F$ are
\begin{equationarray}{rl}
    A_{1} =& \lbrace [3,3],[2,3],[3,2], [2,2], [0,3], \nonumber \\
    & [2,1], [0,2], [1,1], [2,0], [0,1] \rbrace \subset \mathbb{Z}^2 \\
    A_{2} =& \lbrace [2,0], [0,1], [1,0], [0,0] \rbrace \subset \mathbb{Z}^2.
 \end{equationarray}
 The Newton polytopes $P_1 = \text{NP}(f_1)$, $P_2 = \text{NP}(f_2)$ as well as the Minkowski sum $Q = P_1 + P_2$ are depicted in Fig.~\ref{fig:poly_example_1}.  
\end{example}

\subsubsection{Sparse resultants}\label{subsubsec:sparse_resultants_intro}
\noindent Consider the SOTA methods in~\citep{DBLP:journals/jacm/CannyE00,DBLP:conf/issac/EmirisC93,emiris-general},  for computing the sparse resultant for a well-determined generic system $\F$.  The main idea in all such methods is to compute the Minkowski sum of the Newton polytopes of all the polynomials, $f_i \in \F$, and divide its interior into cells. Each cell determines a multiple of one of the input polynomials, $f_i \in \F$ with a monomial. All such monomial multiples collectively lead to an extended polynomial system, $\F^\prime$. The extended polynomial system, via linearization into a matrix product, then leads to a sparse resultant matrix $\M{M}(x_n)$, if using the hidden-variable technique in Sec.~\ref{subsubsec:hidden_var_intro}, or $\M{M}(u_0)$, if using the $u$-resultant of the general form in Sec.~\ref{subsubsec:u_res_intro}. As such the resulting solvers are usually quite large and not very efficient. 

However, if $\F$ has  non-generic coefficients (see Def.~\ref{def:non_genericity}), these methods may fail, as the algebraic constraints on the coefficients leads to algebraic constraints on the elements of the \srs matrices $\M{M}(x_n)$ or $\M{M}(u_0)$ and hence they may not satisfy the necessary rank conditions. 
For such systems,~\cite{DBLP:conf/iccv/Heikkila17} recently proposed an iterative approach based on the hidden variable resultant, to test and extract $\M{M}(x_n)$ in~\eqref{eq:res_mat_form}. Thereafter, it transforms~\eqref{eq:res_mat_form} to a GEP~\eqref{eq:GEP} and solves for eigenvalues and eigenvectors to compute the roots. Our proposed approach here, extends this iterative scheme for generating a \srs matrix $\M{M}(u_0)$, specifically for the extra polynomial approach in Sec.~\ref{subsubsec:u_res_intro}.

\section{Proposed extra polynomial resultant method} \label{sec:extra_poly_sparse_res} 
\noindent 
In this paper, we propose an iterative approach based on~\citep{DBLP:conf/iccv/Heikkila17} and apply it to a modified version of the $u$-resultant method described in Sec.~\ref{subsubsec:u_res_intro}.
Specifically, we propose the following modifications to the $u$-resultant method.
\begin{enumerate}
    \item Instead of the general form~\eqref{eq:u-res_eq} of the extra polynomial in the $u$-resultant based method, in this paper, we propose to use the following special form of the extra polynomial
    \begin{equation}\label{eq:special_form_xtr_poly}
        f_{m+1} = x_{k} - u_0,
    \end{equation}
    where $u_0$ is a new variable and $x_k \in X$ is one of the input variables. In general, we can select any $x_k \in X$. However, since in practice, selecting different $x_k$'s leads to solvers of different sizes and with different numerical stability, we test all $x_k \in X$ when generating the final solver.
     \item For non-generic and overdetermined polynomial systems $\F$, we avoid computing the Minkowski sum of all the Newton polytopes $NP(f), \forall f \in \F_a$, as proposed in the methods in~\citep{DBLP:journals/jacm/CannyE00,DBLP:conf/issac/EmirisC93,emiris-general}. Instead, we iterate through each subset, $\F_{\text{sub}} \subset \F_a$, and compute the Minkowski sum of $NP(f), \forall f \in \F_{\text{sub}}$. Instead of dividing the Minkowski sum into cells, we simply use its lattice interior to determine the monomials $B$ in the extended polynomial system $\F^\prime$, \textit{i.e.} $B=\text{mon}(\F^\prime)$.
      \item We exploit the form  of the extra polynomial~\eqref{eq:special_form_xtr_poly} and propose a block partition of $\Mres$~\eqref{eq:u-res_Mb}, which facilitates its decomposition using Schur complement directly into a regular eigenvalue problem. This regular eigenvalue problem, compared to GEP that arises for general $u$-resultant polynomial~\eqref{eq:u-res_eq}, leads to fewer spurious eigenvalues and hence a faster solver.
\end{enumerate}
      
Note that our method of iteratively testing various polynomial subsets, has been quite effective in generating efficient solvers, as demonstrated on many minimal problems (see Sec.~\ref{sec:experiments}). The generated solvers are comparable in efficiency and speed with those generated by the SOTA action matrix-based methods~\citep{larsson2017efficient,DBLP:conf/cvpr/LarssonOAWKP18,MartyushevVP2022}.

\subsection{Method outline}\label{subsec:sparse_res_outline}
\noindent In the following, we first go through the important steps performed by our method.
\begin{enumerate}
    \item Let us consider a system of $m$ polynomial equations, $\F=0$~\eqref{eq:eq_system}, in $n$ variables, $X = \lbrace x_1,\dots,x_n \rbrace$. For all variables $x_k \in X$, we perform the steps $2-5$ in the offline phase.
    \item \textbf{[Offline]} We fix the form of the extra polynomial to $f_{m+1} = x_k - u_0$~\eqref{eq:special_form_xtr_poly} and augment $\F$ with $f_{m+1}$.
    \begin{equation}\label{eq:inputaug}
        \F_a = \F \sqcup \lbrace f_{m+1} \rbrace.
    \end{equation} We hide the new variable $u_0$ to the coefficient field which means that, $\F_a \subset \mathbb{C}[X]$.
    \item \textbf{[Offline]} We execute steps $3$(a)-$3$(c), for each subset of polynomials $\F_{\text{sub}} \subset \F_a$ and for every variable $x_k \in X$.
    \begin{enumerate}
        \item From the set $\F_{\text{sub}}$, we attempt to construct a \textit{favourable monomial set} $B$, using a polytope-based method, described in Sec.~\ref{subsec:basis_using_polytopes}.
        \item We extend the polynomial system, $\F_a$, using the computed monomial set $B$, represented as
        \begin{equation}
            \F_a \overset{B}{\rightarrow} (\F_a^\prime, T),
        \end{equation}
        where $T=\lbrace T_1,\dots,T_{m+1} \rbrace$~\eqref{eq:poly_extn}.
        \item The set of equations $\F_a^\prime = 0$, can be expressed in a form of a matrix equation as $\M{C}(u_0) \V{b} = \V{0}$, where $\V{b} = \text{vec}(B) $.
    \end{enumerate}
   The output of this step described in detail in Sec.~\ref{subsec:basis_using_polytopes}, is all favourable monomial sets $B$ and the corresponding coefficient matrices $\M{C}(u_0)$.
    \item \textbf{[Offline]} For each favourable monomial set $B$ and the corresponding coefficient matrix $\M{C}(u_0)$ we perform the following:
        \begin{enumerate}
            \item  We partition $B=B_1 \sqcup B_2$ in two different ways,
        \begin{equationarray}{rl}
            B_{1} =& B \cap T_{m+1} \text{  or}  \label{eq:des_mon_partn} \\
            B_{1}  =& \lbrace \mon{\alpha} \in  B  \mid  \dfrac{\mon{\alpha}}{x_k}  \in  T_{m+1} \rbrace. \label{eq:des_mon_partnalt}
        \end{equationarray} Note that $B_2 = B \setminus B_1$.
         \item Based on the two set-partitions of $B$, we attempt to partition the matrix $\M{C}(u_0)$ in the following two ways.
        \begin{equationarray}{l}
           \M{C}(u_0) \V{b}\! = \! \begin{bmatrix} \M{A}_{11} & \M{A}_{12} \\ \M{A}_{21}-u_0 \M{I} & \M{A}_{22} \end{bmatrix}\! \begin{bmatrix} \V{b}_1 \\ \V{b}_2 \end{bmatrix}\! = \! \V{0}, \ \ \label{eq:des_C_partn} \\
          \text{or} \ \nonumber \\
          \M{C}(u_0) \V{b} \! = \! \begin{bmatrix} \M{A}_{11}  & \M{A}_{12} \\ \M{I} + u_0 \M{B}_{21} & u_0 \M{B}_{22}  \end{bmatrix}\! \begin{bmatrix} \V{b}_1 \\ \V{b}_2 \end{bmatrix}\! = \! \V{0}, \ \ \label{eq:des_C_partnalt}
    \end{equationarray} and the matrix $\M{A}_{12}$ has the full column rank.
        \end{enumerate}
       The output of this step is the the coefficient matrix $\M{C}(u_0)$, for which a partitioning as~\eqref{eq:des_C_partn} or~\eqref{eq:des_C_partnalt}, with the full column rank matrix $\M{A}_{12}$ is possible, and which corresponds to smallest set $B_1$. If we have more than one such choice of $\M{C}(u_0)$, we select the smallest matrix $\M{C}(u_0)$. This step is described in more detail in Sec.~\ref{subsec:block_partn}.
        
    \item \textbf{[Offline]} In this step we aim to reduce the size of the matrix $\M{C}(u_0)$, selected in the previous step.
        \begin{enumerate}
            \item We first try to remove a combination of rows and columns from $\M{C}(u_0)$ and the corresponding monomials from the favourable monomial set $B$, such that the resulting monomial set is still a favourable monomial set (Sec.~\ref{subsec:basis_using_polytopes}) and that the coefficient matrix $\M{C}(u_0)$ can still be partitioned as in~\eqref{eq:des_C_partn} or in~\eqref{eq:des_C_partnalt}.
            \item We next remove the extra rows from $\M{C}(u_0)$ to obtain a \srs matrix $\Mres$~\eqref{eq:res_mat_form}, while still respecting its block partition, as in~\eqref{eq:des_C_partn} or~\eqref{eq:des_C_partnalt}.
        \end{enumerate}
        This step is described in Sec.~\ref{subsec:col_red}. 
    \item \textbf{[Online]} In the final online solver, we fill the precomputed \srs matrix $\Mres$ with the coefficients coming from the input data/measurements. Then we compute the Schur complement of a block of $\Mres$, as described in~\ref{subsec:schur_complement}, which is then formulated as an eigenvalue problem of the matrix $\M{X}$~\eqref{eq:eig_prb} (or~\eqref{eq:eig_prb_alt}). Finally, the eigendecomposition of the matrix $\M{X}$ gives us the solutions to the input system of polynomial equations $\F$, \ie~to the given instance of the minimal problem.
\end{enumerate}

\subsection{Computing a favourable monomial set}\label{subsec:basis_using_polytopes}
\noindent  
Let the extra polynomial $f_{m+1}$ have the form $f_{m+1} = x_k - u_0$~\eqref{eq:special_form_xtr_poly} for $x_k \in X$, and let us assume a subset 
 $\F_{\text{sub}} \subset \F_a$ of the augmented system  $\F_a =  \F \sqcup \lbrace f_{m+1} \rbrace $~\eqref{eq:inputaug}.

In the step $3$(a) of our algorithm (Sec.~\ref{subsec:sparse_res_outline}) we attempt to generate so-called favourable monomial sets. 
Our method for constructing these sets is based on the polytope-based algorithm proposed in~\citep{DBLP:conf/iccv/Heikkila17}, where such sets were generated for the original system $\F$ and its subsets. 
Here we describe our method that constructs such favourable sets for subsets $\F_{\text{sub}}$ of the augmented system $\F_a$~\eqref{eq:inputaug}. In this subsection, we refer to the basic terminology and notations described in Sec.~\ref{subsec:polytope_terminology}.

\begin{figure}[t]
\centering
\includegraphics[width=0.80\linewidth]{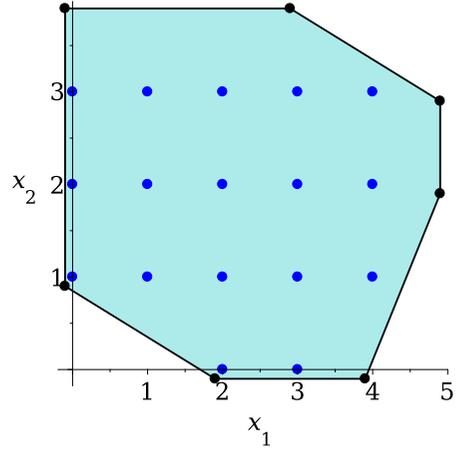}
\caption{The integer points (shown in blue) in the interior of the Minkowski sum of the two Newton polytopes, $Q = P_{1} + P_{2}$ after shifting it $\V{\delta} = [-0.1, -0.1]$.}
\label{fig:poly_example_2}
\end{figure}

In our method, we begin by computing the support $A_{j}=\text{supp}(f_{j})$ and the Newton polytope $\text{NP}(f_{j}) = \text{conv}(A_{j})$ for each polynomial $f_{j} \in \F_a$. We also compute the unit simplex $\text{NP}_{0} \subset \mathbb{Z}^{n}$ and the Minkowski sum, $Q = \text{NP}_{0} + \Sigma_{f \in \F_{\text{sub}}} \text{NP}(f)$. Let us construct a \textit{small} displacement vector $\V{\delta} \in \mathbb{R}^n$, such that each of its elements is assigned one of the three values, $\lbrace -10^{-3}, 0, 10^{-3} \rbrace$. In other words
\begin{equation}\label{eq:def_delta}
    \V{\delta} \in \lbrace \begin{bmatrix} \delta_{1} & \dots & \delta_{n}  \end{bmatrix} \mid \delta_{i} \in \lbrace -10^{-3}, 0, 10^{-3} \rbrace; i = 1,\dots ,n \rbrace.
\end{equation}
Then using this value of $\V{\delta}$, we compute the set of integer points inside $Q+\V{\delta}$, from which we compute a set of monomials, $B = \lbrace \mon{\alpha} \mid \V{\alpha} \in \mathbb{Z}^n \cap (Q+\V{\delta}) \rbrace$, \ie~a potential candidate for a favourable monomial set. We demonstrate this step on a system of two polynomials in Ex.~\ref{fig:poly_example_2}. 
\begin{example}
   Let us continue with Example~\ref{fig:poly_example_1}, where
    we computed the Newton polytopes $P_1 = \text{NP}(f_1)$ and $P_2 = \text{NP}(f_2)$ as well as their Minkowski sum $Q = P_1 + P_2$. Let us assume the displacement vector to be $\V{\delta} = [-0.1,-0.1]$. In Fig.~\ref{fig:poly_example_2} we depict the integer points in the interior of $Q+\V{\delta}$, \ie~the points in the set $\mathbb{Z}^2 \cap (Q+\V{\delta})$.  
   This set of points give us a favourable set of monomials $B$
    \begin{equationarray}{rl}
        B=&  \lbrace x_2, x_2^{2}, x_2^{3}, 
x_1^{2}, x_1^{3}, x_1 x_2, 
x_1 \,x_2^{2}, \nonumber \\
& x_1 \,x_2^{3}, 
x_1^{2} x_2, x_1^{2} x_2^{2}, 
x_1^{2} x_2^{3},
 x_1^{3} x_2, \nonumber \\
& x_1^{3} x_2^{2}, 
 x_1^{3} x_2^{3}, x_1^{4} x_2, 
x_1^{4} x_2^{2}, 
x_1^{4} x_2^{3} \rbrace.

    \end{equationarray}

\end{example}

 In the next step (see step $3$(b) in Sec.~\ref{subsec:sparse_res_outline}), we extend the input polynomial system, $\F_a$, using the set of monomials $B$ as
 \begin{equation}
    \F_a  \overset{B}{\rightarrow} (\F_a^{\prime}, T),
\end{equation}
where $T=\lbrace T_1,\dots,T_{m+1} \rbrace$ (see~\eqref{eq:poly_extn}). Each $T_i$ denotes the set of monomials to be multiplied with the polynomial $f_i \in \F$. 
The extended set of polynomials $\F_a^{\prime}$ can be written in a matrix form as
 \begin{equation}\label{eq:extra_poly_coeff_mat}
     \M{C}([c_{i,\alpha}],u_0) \ \V{b},
 \end{equation} where $\V{b}$ is a vector form of $B$ w.r.t. some ordering of monomials. In the rest of the paper we will denote the coefficient matrix $\M{C}([c_{i,\alpha}], u_0)$ as $\M{C}(u_0)$ for the sake of simplicity.

 Our approach then evaluates the following three conditions:
 \begin{equationarray}{rl}
  \Sigma_{j=1}^{m+1}|T_{j}| \geq & |B|,\label{eq:cond1}\\ 
  \min\limits_{j}|T_{j}| > & 0,\label{eq:cond2}\\
  \text{column rank of } \M{C}(u_0) = & |B|,\label{eq:cond3}
 \end{equationarray}
  for a random value of $u_0 \in \mathbb{C}$. The first condition~\eqref{eq:cond1} ensures that we have at least as many rows as the columns in the matrix $\M{C}(u_0)$. The second condition~\eqref{eq:cond2} ensures that there is at least one row corresponding to every polynomial $f_i \in \F_a$. The third condition~\eqref{eq:cond3} ensures that $\M{C}(u_0)$ has full column rank and that we can extract a \srs matrix from it.  If these conditions are satisfied, we consider the set of monomials $B$ to be a \textit{favourable set of monomials}. 
  
Note that we repeat this procedure for all variables $x_k \in X$, for each subset $\F_{\text{sub}} \subset \F_a$, and for each value of the displacement vector $\V{\delta}$~\eqref{eq:def_delta}. The output of this step are all generated favourable monomial sets $B$ and the corresponding coefficient matrices $\M{C}(u_0)$.

\subsection{Block partition of $\M{C}(u_0)$}\label{subsec:block_partn}
\noindent In the step $4$ of our method (Sec~\ref{subsec:sparse_res_outline}), we iterate through all favourable monomial sets $B$ and the coefficient matrices $\M{C}(u_0)$, which was the output of the previous step.

For each favourable monomial set $B$, we know the corresponding sets of monomial multiples $T_i$ for each polynomial $f_i \in \F_a$ and the set $T = \lbrace T_{1},\dots,T_{m},T_{m+1} \rbrace$. We have also computed the corresponding coefficient matrix $\M{C}(u_0)$. Assume the extra polynomial~\eqref{eq:special_form_xtr_poly} to be fixed as $f_{m+1} = x_k - u_0$, for $x_k \in X$, while computing this favourable monomial set.

Since $B$ was constructed such that it satisfies~\eqref{eq:cond3}, $\M{C}(u_0)$ has full column rank. Let us assume, $\varepsilon = \mid B \mid $ and $p = \Sigma_{j=1}^{m+1}|T_{j}|$. As $\M{C}(u_0)$ satisfies the condition~\eqref{eq:cond1}, we have $p \geq \varepsilon$.
Therefore, we can remove the extra $p-\varepsilon$ rows from $\M{C}(u_0)$, leading to a square invertible matrix. The algorithm of row removal approach is described in~\ref{subsubsec:rowremoval}. 
The square matrix so obtained, is a sparse resultant matrix $\Mres$ and it satisfies the resultant matrix constraint~\eqref{eq:res_mat_sing}, if the set of equations $\F_a = 0$ has solutions. 

Instead of directly solving the resultant constraint~\eqref{eq:res_cnst} or converting~\eqref{eq:res_mat_sing} to GEP, we exploit the structure of the extra polynomial $f_{m+1}$~\eqref{eq:special_form_xtr_poly} and propose a special ordering of the rows and columns of $\Mres$. This facilitates a block partition of $\Mres$, such that the Schur complement of one of its block submatrices, then helps us to reduce its matrix constraint~\eqref{eq:res_mat_sing} to a regular eigenvalue problem\footnote{Note that it may happen that neither of the two set-partitions for a given favourable monomial set $B$, lead to a block partition of $\M{C}(u_0)$ such that $\M{A}_{12}$ has full column rank. In this case, the Schur complement does not exist and our algorithm will continue to test the next favourable monomial set $B$.}. 

Rather than block partitioning the \srs matrix $\Mres$, we first fix a block partition of the coefficient matrix $\M{C}(u_0)$ and then remove its rows, while respecting this partition, to obtain the \srs matrix $\Mres$. 
Note that we obtain two different block partitions on $\V{b}$ and $\M{C}(u_0)$, in~\eqref{eq:extra_poly_coeff_mat}, based on our selected set-partition of the favourable monomial set $B =  B_{1} \sqcup B_{2}$:
\begin{equationarray}{l}
    B_{1} = B \cap T_{m+1} \text{or} \label{eq:res_mon_partn} \\
    B_{1}  = \lbrace \mon{\alpha} \in  B  \mid  \dfrac{\mon{\alpha}}{x_k}  \in  T_{m+1} \rbrace \label{eq:res_mon_partnalt}
\end{equationarray} Note that $B_{2} = B \setminus B_{1}$. We consider both these partitions in Prop.~\ref{prop:eigendecomp}. Thereafter, we will describe our method of removing the rows from $\M{C}(u_0)$ such that we can compute a Schur complement of a block of $\Mres$.

\begin{proposition}\label{prop:eigendecomp}
Consider a block partition of $\M{C}(u_0)$ and $\V{b}$ in~\eqref{eq:extra_poly_coeff_mat} as
\begin{equationarray}{l}\label{eq:blockpartition}
    \M{C}(u_0)\  \V{b}  = \begin{bmatrix} \M{C}_{11} & \M{C}_{12} \\ \M{C}_{21} & \M{C}_{22} \end{bmatrix} \begin{bmatrix} \V{b}_{1} \\ \V{b}_{2} \end{bmatrix}.
\end{equationarray} We can achieve this in two different ways, based on our choice the set-partition of $B$ in~\eqref{eq:res_mon_partn} or~\eqref{eq:res_mon_partnalt}. In either case, if $\M{C}_{12}$ has full column rank, then the resultant matrix constraint~\eqref{eq:res_mat_sing} can be converted to an eigenvalue problem, of the form $\M{X} \ \V{b}_{1} = u_0\V{b}_{1}$ if we set-partition $B$ as in~\eqref{eq:res_mon_partn}, and of the form $\M{X} \ \V{b}_{1} = -\dfrac{1}{u_0} \V{b}_{1}$ if we set-partition $B$ as in~\eqref{eq:res_mon_partnalt}.  The matrix $\M{X}$ is square and its entries are the functions of the coefficients of $\F$. 
\end{proposition}
\textit{Proof:}
Let us first consider the set-partition of $B$ as in~\eqref{eq:res_mon_partn}.
Then, the special form of the extra polynomial~\eqref{eq:special_form_xtr_poly} implies that
\begin{equationarray}{l}\label{eq:res_mon_partn2}
   B_{1} \! =\! \lbrace \V{x}^{\V{\alpha}} \in B | \  x_k  \V{x}^{\V{\alpha}}  \in B   \rbrace. 
\end{equationarray} We next order the monomials in $B$ such that $\V{b}$ can be partitioned as $\V{b}\! =  \begin{bmatrix} \text{vec}(B_{1})\; \text{vec}(B_{2}) \end{bmatrix}^{T}\! =\! \begin{bmatrix} \V{b}_1\; \V{b}_2 \end{bmatrix}^{T}$. This induces a column partition of $\M{C}(u_0)$. Moreover, we can row partition $\M{C}(u_0)$ by indexing the rows in its lower block by monomial multiples of $f_{m+1}\V{(x^{\prime})}$. All such multiples are linear in $u_0$, \ie, these multiples have the form $\V{x}^{\V{\alpha_j}} (x_{k} - u_0), \forall \V{x}^{\V{\alpha_j}}\in T_{m+1}$. The upper block is row-indexed by monomial multiples of $f_{1}\V{(x^{\prime})},\dots,f_{m}\V{(x^{\prime})}$. Such a row and column partition of $\M{C}(u_0)$ gives us a block partition as in~\eqref{eq:blockpartition}. As $\begin{bmatrix} \M{C}_{11}\! &\! \M{C}_{12} \end{bmatrix}$ contains polynomials independent of $u_0$ and $\begin{bmatrix} \M{C}_{21}\! &\! \M{C}_{22}\! \end{bmatrix}$ contains polynomials of the form $\V{x}^{\V{\alpha_j}} ( x_{k} - u_0)$ we obtain
\begin{equationarray}{lcl}\label{eq:blockmatdecom}
\M{C}_{11} = \M{A}_{11}, \; \; \; \M{C}_{12} = \M{A}_{12} \nonumber \\
\M{C}_{21} = \M{A}_{21} + u_0 \M{B}_{21}, \; \; \;  
\M{C}_{22} = \M{A}_{22} + u_0 \M{B}_{22},
\end{equationarray}
where $\M{A}_{11}, \M{A}_{12}, \M{A}_{21}$, $\M{A}_{22}$, $\M{B}_{21}$ and $\M{B}_{22}$ are matrices dependent only on the coefficients of input polynomials in $\F_a$~\eqref{eq:inputaug}. 
Based on our assumption in the statement of this proposition, $\M{C}_{12} $ and hence $\M{A}_{12}$, have full column rank. Substituting~\eqref{eq:blockmatdecom} in~\eqref{eq:blockpartition} gives 
\begin{equationarray}{lll}\label{eq:fav_partn}
\M{C}(u_0) &=  \underbrace{\begin{bmatrix} \M{A}_{11} & \M{A}_{12} \\ \M{A}_{21} & \M{A}_{22} \end{bmatrix}}_{\M{C}_0} + u_0 \underbrace{\begin{bmatrix} \M{0} & \M{0} \\ \M{B_{21}} & \M{B_{22}} \end{bmatrix}}_{\M{C}_1} &
\end{equationarray}
We can order monomials so that $\V{T}_{m+1} = \V{b}_1$. Now, the block partition of $\M{C}(u_0)$ implies that $\M{C}_{21}$ is column-indexed by $\V{b}_1$ and row-indexed by $\V{T}_{m+1}$. 
As $\begin{bmatrix} \M{C}_{21}\! & \! \M{C}_{22}  \end{bmatrix}$ has rows of form $\V{x}^{\V{\alpha_j}}(x_{k}\! -\! u_0)$, $\V{x}^{\V{\alpha_j}}\! \in\! T_{m+1}\! \implies \! \V{x}^{\V{\alpha_j}}\! \in\! B_{1}$. 
This gives us, $\M{B}_{21} = \M{-I}$, where $\M{I}$ is an identity matrix of size $\mid B_{1} \mid \times \mid B_{1} \mid $ and $\M{B}_{22}$ is a zero matrix of size $\mid B_{1}\mid \times \mid B_{2} \mid $. 
This also means that $\M{A}_{21}$ is a square matrix of the same size as $\M{B}_{21}$. Thus we have
\begin{equationarray}{ll}\label{eq:spresmatstruct}
\M{C}(u_0) &  =  \begin{bmatrix} \M{A}_{11} & \M{A}_{12} \\ \M{A}_{21} & \M{A}_{22} \\ \end{bmatrix} + u_0 \begin{bmatrix} \M{0} & \M{0} \\ - \M{I} & \M{0} \end{bmatrix},
\end{equationarray}
where $\M{C}(u_0)$ is a $p \times \varepsilon$ matrix. If $\M{C}(u_0)$ is a \textit{tall} matrix, so is $\M{A}_{12}$. Therefore, we can eliminate extra rows from the upper block $\begin{bmatrix}
  \M{A}_{11} & \M{A}_{12}
\end{bmatrix}$ so that we obtain a square invertible matrix $\hat{\M{A}}_{12}$ while preserving the above mentioned structure. Such a row-removal will also give us the \srs matrix $\Mres$ which satisfies the resultant matrix constraint~\eqref{eq:res_mat_sing}. We will describe our method for removing the extra rows in Sec.~\ref{subsec:col_red}. Let us now here, use the Schur complement technique~\ref{subsec:schur_complement}, to reduce the size of the eigenvalue problem. From~\eqref{eq:spresmatstruct}, we have
\begin{equationarray}{ll}\label{eq:spresmatstructdecomp}
  &  \underbrace{\begin{bmatrix} \hat{\M{A}}_{11} & \hat{\M{A}}_{12} \\ \M{A}_{21} & \M{A}_{22} \end{bmatrix}}_{\hat{\M{C}}_{0}} \begin{bmatrix} \V{b}_1  \\ \V{b}_2 \end{bmatrix} + u_0 \begin{bmatrix} \M{0} & \M{0} \\ - \M{I} & \M{0} \end{bmatrix} \begin{bmatrix} \V{b}_1 \\ \V{b}_2 \end{bmatrix}  =  \V{0} \nonumber \\
 \implies & \hat{\M{A}}_{11} \V{b}_1 + \hat{\M{A}}_{12} \V{b}_2 = \V{0}, \nonumber \\
  & \M{A}_{21} \V{b}_1 + \M{A}_{22} \V{b}_2 - u_0 \V{b}_1 = \V{0}.
\end{equationarray}
\noindent Eliminating $\V{b}_2$ from the above pair of equations we obtain
\begin{eqnarray}\label{eq:eig_prb}
  \underbrace{(\M{A}_{21} - \M{A}_{22} \hat{\M{A}}_{12}^{-1} \hat{\M{A}}_{11})}_{\M{X}} \V{b}_1 = u_0 \V{b}_1.
\end{eqnarray} The matrix $\M{X}$ is the Schur complement of $\hat{\M{A}}_{12}$ w.r.t. $\hat{\M{C}}_{0}$.

If we select the alternative set-partition of $B$, as in~\eqref{eq:res_mon_partnalt}, we obtain a different block partition of $\V{b}$ and $\M{C}(u_0)$. Specifically, in~\eqref{eq:fav_partn}, we have $\M{A}_{21}\! =\! \M{I}$ and $\M{A}_{22}\! =\! \M{0}$. By assumption, we have $\M{A}_{12}$ has full column rank. Therefore, we can remove the extra rows from $\M{C}(u_0)$ in~\eqref{eq:spresmatstruct} and obtain
\begin{equationarray}{ll}\label{eq:spresmatstructalt}
\Mres  & =  \begin{bmatrix} \hat{\M{A}}_{11} & \hat{\M{A}}_{12} \\ \M{I} & \M{0} \\ \end{bmatrix} + u_0 \begin{bmatrix} \M{0} & \M{0} \\ \M{B}_{21} & \M{B}_{22} \end{bmatrix}.
\end{equationarray}
 Substituting the value of $\Mres$, in~\eqref{eq:res_mat_sing}, we get $\hat{\M{A}}_{11} \V{b}_1 + \hat{\M{A}}_{12} \V{b}_2 = \V{0}$ and $u_0(\M{B}_{21} \V{b}_1  + \M{B}_{22} \V{b}_2) +  \V{b}_1  =  \V{0}$. Eliminating $\V{b}_2$ from these equations, we obtain an alternate eigenvalue formulation 
\begin{eqnarray}\label{eq:eig_prb_alt}
    \underbrace{(\M{B}_{21} - \M{B}_{22} \hat{\M{A}}_{12}^{-1} \hat{\M{A}}_{11})}_{\M{X}} \V{b}_1 = -(1/u_0) \V{b}_1. \;\; \qedsymbol
\end{eqnarray} The matrix $\M{X}$ here, represents an alternative representation to the one in~\eqref{eq:eig_prb}, of the Schur complement.

Note that this proposition allows us to test the existence of the Schur complement $\M{X}$~\eqref{eq:eig_prb} (or~\eqref{eq:eig_prb_alt}), for a given favourable monomial set $B$ and the corresponding coefficient matrix $\M{C}(u_0)$. This condition is tested for each $B$ and corresponding $\M{C}(u_0)$, for both choices of set-partitions~\eqref{eq:res_mon_partn} or~\eqref{eq:des_mon_partnalt}. Out of those that succeed, we select the coefficient matrix $\M{C}(u_0)$, which led to the smallest possible Schur complement $\M{X}$. If we have more than one such choice, we choose the smallest coefficient matrix $\M{C}(u_0)$. Note that our iterative approach is crucial in increasing the chances of obtaining a minimal solver, even when the polynomials in $\F$ have non-generic coefficients. 
 
\subsection{Row-column removal}\label{subsec:col_red}
\noindent The next step in our method is to attempt to reduce the size of the matrix $\M{C}(u_0)$ selected in the previous step. For this, we employ a method, inspired by~\citep{Kukelova-thesis}. Here we select columns of $\M{C}(u_0)$, one by one in a random order to test for their removal. For each such column, we select rows (say $r_{1},\dots,r_{s}$) that contain non-zero entries in the column and also consider all columns (say $c_{1},\dots,c_{l}$) that have non-zero entries in $r_{1},\dots,r_{s}$. Then we can remove these $s$ rows and $l$ columns from $\M{C}(u_0)$ only if the following conditions hold true for the resulting reduced matrix $\M{C}_{\text{red}}(u_0)$. 
\begin{enumerate}
    \item After eliminating the monomials from $T$, we require that there is at least one monomial left in each $T_{i}$.
    \item If $\M{C}(u_0)$ is of size $p \times \varepsilon$, the reduced matrix $\M{C}_{\text{red}}(u_0)$ would be of size $(p-s) \times (\varepsilon-l)$. Then we require $p-s \geq \varepsilon-l$ and $\text{rank}(\M{C}_{\text{red}}(u_0)) = \varepsilon-l$.
    \item $\M{C}_{\text{red}}(u_0)$ must be block partitioned and decomposed as in Proposition~\ref{prop:eigendecomp}. 
\end{enumerate} 

We repeat the above process until there are no more columns that can be removed. Note that the last condition is important as it ensures that at each stage, the reduced matrix can still be partitioned and decomposed into an eigenvalue formulation~\eqref{eq:eig_prb} (or alternately~\eqref{eq:eig_prb_alt}). Now by abuse of notation, let us denote $\M{C}(u_0)$ to be the reduced matrix and denote $B$ and $T$ to be the reduced set of monomials and the set of monomial multiples, respectively. 

If $\M{C}(u_0)$ still has more rows than columns, we transform it into a square matrix by removing extra rows (say $q_{1},\dots,q_{j}$) and the monomials from $T$ indexing these rows, chosen in a way such that the three conditions mentioned above are still satisfied. Note that it is always possible to choose rows, such that these three conditions are satisfied.  Moreover, our proposed approach first, tries to remove as many rows as possible from the lower block, indexed by $T_{m+1}$. This is to reduce $|T_{m+1}|$($=|B_{1}|$) as much as possible and ensure that the matrix $\M{A}_{21}$ and hence $\M{X}$~\eqref{eq:eig_prb} (or~\eqref{eq:eig_prb_alt}) for eigenvalue problem has as small size as possible. Then if there are more rows still to be removed, the rest are randomly chosen from the upper block indexed by $\lbrace T_{1},\dots,T_{m} \rbrace$. Algorithms for this step of matrix reduction are provided in the Appendix~\ref{subsec:polytope_extra_poly_alg}. 

The \srs matrix $\Mres$ is constructed offline through these three steps. In the online stage, we fill in the coefficients of $\Mres$ from the input measurements and compute a Schur complement $\M{X}$~\eqref{eq:eig_prb} (or~\eqref{eq:eig_prb_alt}). Then, we extract the  solutions to $x_{1},\dots ,x_{n}$ by computing the eigenvectors of $\M{X}$. The speed of execution of the solver depends on the size of $\V{b}_1$(=$\lvert B_{1} \rvert $) as well the size of $\hat{\M{A}}_{12}$ while the accuracy of the solver largely depends on the condition number as well as the size of the matrix to be inverted \ie, $\hat{\M{A}}_{12}$.

\section{Action matrix vs \srs}\label{sec:am_vs_res}
\noindent In this section, we study the connection between a solver generated using an action matrix method~\citep{Kukelova-ECCV-2008,Kukelova-thesis,Stetter1996,larsson2017efficient,byrod2007improving,byrod-etal-ijcv-2009,DBLP:conf/cvpr/LarssonOAWKP18}, and a solver based on the \srs method proposed in Sec.~\ref{sec:extra_poly_sparse_res}. Observe that, our \srs method exploits only the structure of the input polynomials, via the geometry of their Newton polytopes, whereas the SOTA action matrix method algebraically investigate the input polynomial system via the \gb. Seemingly different, the generated solvers exhibit the same properties, and hence in this section our objective is to throw some light on the similarities. Some of the symbols used in both the methods are the same. Therefore, to distinguish them, in this section we will use the prefixes $\prfx{a}$ and $\prfx{r}$ to respectively denote the symbols used in the action matrix method in Sec.~\ref{subsec:action_matrix_outline} and the \srs method in Sec.~\ref{subsec:sparse_res_outline}.
Let us assume that both these methods are used to generate a solver for a system $\F=0$~\eqref{eq:eq_system} of $m$ polynomial equations  in $n$ variables with $r$ solutions. An action matrix method results in a solver that is performing G-J elimination (or alternatively LU/QR decomposition) of an elimination template $\prfx{a}\!\M{C}$~\eqref{eq:am_C} and the eigendecomposition of an action matrix $\M{M}_f$~\eqref{eq:def_action_matrix} of size $r \times r$~\footnote{The size of the action matrix in all state-of-the art action matrix methods~\citep{Kukelova-ECCV-2008,Kukelova-thesis,Stetter1996,larsson2017efficient,byrod2007improving,DBLP:conf/cvpr/LarssonOAWKP18} is equal to the number of solutions to the input system. The only exceptions are the methods proposed in~\citep{byrod-etal-ijcv-2009} and~\citep{martyushev2023automatic}, which can result in larger action matrices.}.  The proposed \srs method, on the other hand, leads to a solver, which computes a Schur complement $\M{X}$~\eqref{eq:eig_prb} (or \eqref{eq:eig_prb_alt}) of size $N \times N$, from the \srs matrix $\Mres$, followed by its eigendecomposition. 
In general, $N \geq r$. 

While the final solvers generated by these two methods are in general different, \ie, they perform different operations, at the same time, they demonstrate some similar properties, \eg, both these solvers extract solutions from eigenvectors of some matrices, and both involve computing a matrix inverse (G-J elimination can be replaced by matrix inverse).
This motivates us to study the similarities of these solvers. Let us first define the \textit{equivalence} of a solver generated using an action matrix method (\textbf{AM}) and a solver generated using the sparse resultant method from Sec.~\ref{sec:extra_poly_sparse_res} (\textbf{SRes}).
\begin{definition}\label{def:solver_equiv}
For a given system of polynomial equations $\F=0$~\eqref{eq:eq_system}, an action matrix-based solver and a \srs-based solver are defined to be equivalent, iff the following two conditions are satisfied:
\begin{enumerate}
    \item The action matrix $\M{M}_f$~\eqref{eq:def_action_matrix} is the same as the Schur complement $\M{X}$~\eqref{eq:eig_prb} (or~\eqref{eq:eig_prb_alt}), \ie~$\M{M}_f = \M{X}$.
    \item The step of G-J elimination of the template matrix $\prfx{a}\!\hat{\M{C}}$ in the online stage (step $7$) of the action matrix method (Sec.~\ref{subsec:action_matrix_outline}) can be replaced with the matrix product $\hat{\M{A}}_{12}^{-1} \hat{\M{A}}_{11}$, used to compute the Schur complement in the online stage (step $6$) of our \srs method (Sec.~\ref{subsec:sparse_res_outline}), or vice-versa.
    The size of the matrix $\prfx{a}\!\hat{\M{C}}$ is the same as the size of the upper block $\begin{bmatrix} \hat{\M{A}}_{11} & \hat{\M{A}}_{12}\end{bmatrix}$ of the \srs matrix $\Mres$~\eqref{eq:spresmatstructdecomp} (or~\eqref{eq:spresmatstructalt}).
    
\end{enumerate}
\end{definition}

Now, our goal is to define conditions under which the final online solvers generated by these two methods are \textit{equivalent}.

We next demonstrate that if we have an action matrix-based solver (\textbf{AM}) for a given system $\F=0$~\eqref{eq:eq_system}, we can modify the steps performed by the \srs method proposed in the Sec.~\ref{subsec:sparse_res_outline}, such that both the solvers are equivalent according to Def.~\ref{def:solver_equiv}. 

\subsection{$\M{X}$ (\textbf{SRes}) from $\M{M}_f$ (\textbf{AM})}\label{subsec:res_frm_actn_mat}
\noindent Let us assume that we have generated a solver for a system of polynomial equations, $\F=0$~\eqref{eq:eq_system} using the action matrix method described in Sec.~\ref{subsec:action_matrix_outline}. The action polynomial is assumed to be $f=x_1$, where $ x_1 \in X$.

We first propose the following changes to the steps performed by the \srs method in the offline stage,
in Sec.~\ref{subsec:sparse_res_outline}. In what follows, we will assume that step $i$ actually means the step $i$ in Sec.~\ref{subsec:sparse_res_outline}.
\begin{enumerate}[label=\textbf{C\arabic*R}]
    \item The step $1$ is skipped, as there is no need to test each variable $x_k \in X$.
    \item In the step $2$, the extra polynomial is assumed to be $f_{m+1} = x_1 - u_0$. 
    \item In the step $3(a)$, the favourable set of monomials $\prfx{r}\!B$ is directly constructed from $\prfx{a}\!B$, as $\prfx{r}\!B = \prfx{a}\!B$. Moreover, the set of monomial multiples is constructed as $\prfx{r}T_i = \prfx{a}T_i, i=1,\dots,m$ and $\prfx{r}T_{m+1}$ = $B_a$. 
    
    The step $3(b)$ is to be skipped, as the coefficient matrix $\prfx{r}\M{C}(u_0)$ will be directly obtained from the action matrix-based solver, in subsequent steps. The output of this step then is the favourable monomial set $\prfx{r}\! B$.
    
    \item Replace the step $4(a)$ by directly determining the set partition of $\prfx{r}\!B=B_1\sqcup B_2$~\eqref{eq:des_mon_partn}-\eqref{eq:des_mon_partnalt}, as $B_1 = B_a $ and $ B_2 = \hat{B}_e \sqcup B_r$. Moreover, the monomial ordering in $\prfx{a}\V{b}$ determines the vectors, $\V{b}_1 = \V{b}_a$ and $\V{b}_2 = \begin{bmatrix}
       \V{\hat{b}}_e \\
       \V{b}_r
    \end{bmatrix} $. The action matrix-based solver corresponds to the first version of set partition of $\prfx{r}\!B$, considered in our sparse resultant method, \ie~the set partition as in~\eqref{eq:des_mon_partn}.
    \item In the step $4(b)$, the upper block $\begin{bmatrix}
    \M{A}_{11} & \M{A}_{12}
    \end{bmatrix}$ of $\prfx{r}\M{C}(u_0)$~\eqref{eq:des_C_partn} is directly obtained from $\prfx{a}\hat{\M{C}}$, as
            \begin{equationarray}{rl}\label{eq:invertible_left_block_C}
                \M{A}_{11} =& \begin{bmatrix} \M{C}_{13} \\ \M{C}_{23} \end{bmatrix}, \nonumber \\
                \M{A}_{12} =& \begin{bmatrix}
               \hat{\M{C}}_{11} & \M{C}_{12} \\
               \hat{\M{C}}_{21} & \M{C}_{22} 
           \end{bmatrix}. 
           \end{equationarray}

     \noindent Multiplying $f_{m+1}$ with the monomials in $\prfx{r}T_{m+1}$, the expanded set of polynomials is obtained as $\lbrace \mon{\alpha} \left( x_1 - u_0 \right) \mid \mon{\alpha} \in \prfx{r}T_{m+1} \rbrace$. It is possible to order the polynomials in this set, such that its matrix form is
    \begin{equationarray}{l}
        \begin{bmatrix}
           \M{A}_{21} - u_0 \M{I} & \M{A}_{22}
       \end{bmatrix} \begin{bmatrix}
           \V{b}_1 \\ \V{b}_2
       \end{bmatrix}.
    \end{equationarray} This determines the lower block of $\prfx{r}\M{C}(u_0)$.
    Here, $\M{A}_{21}$ and $\M{A}_{22}$ are binary matrices, with entries either $0$ or $1$. The upper and the lower blocks, together give us 
    \begin{equationarray}{l}\label{eq:des_C_partn_repeated} 
    \prfx{r}\M{C}(u_0) \ \prfx{r}\V{b} =  \begin{bmatrix}  \M{A}_{11} & \M{A}_{12} \\ \M{A}_{21} - u_0 \M{I} & \M{A}_{22} \end{bmatrix} \begin{bmatrix}
           \V{b}_1 \\ \V{b}_2
       \end{bmatrix}. \end{equationarray} 
       The output of the modified step $4$, is the coefficient matrix $\prfx{r}\M{C}(u_0)$. 
       
     \item By construction, $\prfx{r}\M{C}(u_0)$ is already a square invertible matrix and has no extra rows to be removed. As we do not need to attempt row-column removal from $\prfx{r}\M{C}(u_0)$, the step $5$ is to be skipped, to directly construct the \srs matrix as $\Mres = \prfx{r}\M{C}(u_0)$. In this case, the submatrices $\hat{\M{A}}_{11}$ and $\hat{\M{A}}_{12}$ of $\Mres$ are equal to the submatrices $ \M{A}_{11}$ and $ \M{A}_{12}$ of $\prfx{r}\M{C}(u_0)$, respectively.
    \end{enumerate}
    
\vspace{10px}
\noindent Applying these changes to the steps 1-5 in Sec.~\ref{subsec:sparse_res_outline}, \ie~the offline stage, we obtain the \srs matrix $\Mres$. After that, the online solver computes the Schur complement $\M{X}$ of the block submatrix $\hat{\M{A}}_{12}$ of $\Mres$, followed by an eigenvalue decomposition~\eqref{eq:eig_prb} of $\M{X}$ as described in the step $6$ in Sec.~\ref{subsec:sparse_res_outline}. In the proposition~\ref{prop:X_frm_Mf} we prove that the \srs-based solver so obtained, is equivalent to the action matrix-based solver.

\subsection{$\M{M}_f$ (\textbf{AM}) from $\M{X}$ (\textbf{SRes})}\label{subsec:Mf_frm_X}
\noindent Let us assume that for a system of polynomial equations, $\F=0$~\eqref{eq:eq_system}, we have a \srs-based solver, generated by following the steps in Sec.~\ref{subsec:sparse_res_outline}. In this solver the coefficient matrix $\prfx{r}\M{C}(u_0)$ is assumed to be partitioned as in~\eqref{eq:des_C_partn}, and the alternative partition~\eqref{eq:des_C_partnalt} will be considered in the next Sec.~\ref{subsec:Mf_frm_Xalt}. Moreover, the Schur complement $\M{X}$ is assumed to have as many eigenvalues as the number of solutions to $\F=0$, \ie, $N = r$\footnote{If the Schur complement $\M{X}$ has more eigenvalues $N$ than the number of solutions $r$, we can still change the steps performed by the action matrix method such that it leads to a solver equivalent to the \srs-based solver. However such a case would be too complicated, and is not discussed here. Recently,~\cite{martyushev2023automatic} have proposed an action matrix-based automatic generator which attempts to generate solvers with an eigenvalue problem larger than the number of roots of the system.}. The extra polynomial is supposed to be of the form $f_{m+1} = x_1 - u_0$, for $x_1 \in X$. 

We first propose the following changes to the steps performed by the action matrix method in the offline stage,
in Sec.~\ref{subsec:action_matrix_outline}. In the following list, we assume that step $i$ actually means the step $i$ in Sec.~\ref{subsec:action_matrix_outline}.
\begin{enumerate}[label=\textbf{C\arabic*A}]
    \item There is no change to the step $1$.
    \item In the step $2$, set the basis of the quotient ring $A$, to be $\mathcal{B}_A = \lbrace [\mon{\alpha}] \mid \mon{\alpha} \in B_1\rbrace \subset A$. By assumption, $A$ is an $r$-dimensional vector space over $\mathbb{C}$ spanned by the elements of $\mathcal{B}_A$\footnote{It was proved in~\citep[Theorem~6.17]{Cox-Little-etal-05}, that if $\F$ is a generic polynomial system, then $A$ would be spanned by the elements of the set $\mathcal{B}_A$. However even if the polynomial system is not generic, we can still follow the steps performed in the same proof. 
    }. 
    Construct the monomial set $B_a = B_1$.
    \item In the step $3$, assume the action polynomial $f = x_1$.
    \item The monomial multiples, required in the step $4$, are obtained as $\prfx{a}T_i = \prfx{r}T_i, i=1,\dots,m$. This also gives us the extended polynomial set $\F^\prime$. Note that $\prfx{a}\!B = \prfx{r}\!B = \text{mon}(\F^\prime)$.
    \item In the step $5$, the required sets of reducible and excess monomials are determined as
    \begin{equationarray}{l}
        B_r = \lbrace x_1 \mon{\alpha} \mid \mon{\alpha} \in B_1 \rbrace \setminus B_1 \nonumber  \\
        B_e = B_2 \setminus B_r.
    \end{equationarray}
    Note that by~\eqref{eq:res_mon_partn2}, $ \mon{\alpha} \in B_1 \implies x_1 \mon{\alpha} \in \prfx{r}\!B $. 

    \item In the step $6$, the vector of monomials is determined as $\prfx{a}\hat{\V{b}}=\prfx{r}\V{b}$. This will also determine the vectors $\V{b}_a, \V{b}_r$ and $\hat{\V{b}}_e$. Thus, the monomials in the action matrix method are ordered in the same way as the monomials in the \srs method.
    
    Moreover, in the step $6$, the elimination template $\prfx{a}\hat{\M{C}}$, and its block partition, are determined as
        \begin{equationarray}{rl}
         \begin{bmatrix}
               \hat{\M{C}}_{11} & \M{C}_{12} \\
               \hat{\M{C}}_{21} & \M{C}_{22} 
                \end{bmatrix} =& \hat{\M{A}}_{12}, \nonumber \\
                \begin{bmatrix}
               \M{C}_{13} \\
               \M{C}_{23} 
                \end{bmatrix} =& \hat{\M{A}}_{11}.\label{eq:C_frm_M}
        \end{equationarray} The column partition of $\prfx{a}\hat{\M{C}}$ is determined by the partition of the vector $\prfx{a}\V{b}$. By the construction of $\Mres$, $\hat{\M{A}}_{12}$ is a square invertible matrix. Therefore, we can always find a row partition such that $\M{C}_{22}$ is a square invertible matrix. As, $\hat{\M{C}}_{22}$ and $\hat{\M{A}}_{12}$ are square matrices, $\hat{\M{C}}_{11}$ is also a square matrix. Therefore, there are no extra columns to be removed and the elimination template $\prfx{a}\hat{\M{C}}$ will be such that its G-J elimination will lead to the required form, as in~\eqref{eq:LU_C}.
\end{enumerate}

\vspace{10px}
\noindent Applying these changes to the steps $1$-$6$ in Sec.~\ref{subsec:action_matrix_outline}, \ie~the offline stage, we obtain the elimination template $\prfx{a}\hat{\M{C}}$. 
Subsequently, the online action matrix-based solver (step $7$ in Sec.~\ref{subsec:action_matrix_outline}) computes the G-J elimination of $\prfx{a}\hat{\M{C}}$, from which the entries of the action matrix $\M{M}_f$ are read-off. This implies that the action matrix-based solver is equivalent to the \srs-based solver, which is proved in Proposition~\ref{prop:Mf_frm_X}.

\subsection{$\M{M}_f$ (\textbf{AM}) from $\M{X}$ (\textbf{SRes}) with alternative form}\label{subsec:Mf_frm_Xalt}
\noindent Let us assume that for a system of polynomial equations, $\F=0$~\eqref{eq:eq_system}, we have used the alternative partition of $\prfx{r}\!B$~\eqref{eq:des_mon_partnalt} and $\prfx{r}\M{C}(u_0)$~\eqref{eq:des_C_partnalt}, and generated a \srs-based solver by following the steps in Sec.~\ref{subsec:sparse_res_outline}. This means that we need to assume that no solution of $\F=0$, has $x_1=0$. The other assumptions remain the same as in the Sec.~\ref{subsec:Mf_frm_X}.

The alternative partition of the favourable monomial set $\prfx{r}\!B$~\eqref{eq:des_mon_partnalt} implies that the Schur complement $\M{X}$~\eqref{eq:eig_prb_alt} gives us a representation of each coset $\left[ \dfrac{\mon{\alpha_j}}{x_1} \right], \mon{\alpha_j} \in B_1 $ as a linear combination of  the cosets $[\mon{\alpha_j}], \forall  \mon{\alpha_j} \in B_1$. However, the approach in Sec.~\ref{subsec:Mf_frm_X} does not work, because in this case we need to set the action polynomial $f$ to be $1/x_1$. 

Therefore, we can use the so-called \textit{Rabinowitsch trick}~\citep{Cox-Little-etal-05}, and propose the following changes to the offline stage of the action matrix method in Sec.~\ref{subsec:action_matrix_outline}.  In the following list, we will assume, that step $i$ actually means the step $i$ in Sec.~\ref{subsec:action_matrix_outline}.

\begin{enumerate}[label=\textbf{C\arabic*A$^\prime$}]
    \item In the step $1$, the polynomial system $\F$, is augmented with an extra polynomial $\prfx{a}\!f_{m+1}=x_1 \lambda - 1$, where $\lambda$ is a new variable. The augmented polynomial system is denoted as $\prfx{a}\!\F_a$, and the augmented set of variables as $\prfx{a}\!X_a = X \sqcup \lbrace \lambda \rbrace$. Consider the ideal $I_a = \langle \prfx{a}\!\F_a \rangle$ and the quotient ring $A_a = \mathbb{C}[\prfx{a}\!X_a]/I_a$.

    \item Note that the number of solutions to $\F=0$ is the same as that of $\prfx{a}\!\F_a=0$, because we have assumed $x_1 \neq 0$. Therefore, $A_a$ is an $r$-dimensional vector space over $\mathbb{C}$. Its basis is constructed as $\mathcal{B}_A = \lbrace [\mon{\alpha_j}] \mid \mon{\alpha_j} \in B_1\rbrace$. Construct the monomial set $B_a = B_1$.
    
    \item In the step $3$, the action polynomial is assumed to be $f = \lambda$.
    
    \item The sets of monomial multiples required in the step $4$, are constructed as $\prfx{a}T_i = \prfx{r}T_i, i=1,\dots,m+1$. From~\eqref{eq:des_mon_partnalt}, $\mon{\alpha_j} \in T_{m+1}$ implies $x_1 \mon{\alpha_j} \in B_1$, which implies that $\lambda x_1 \mon{\alpha_j} \in B_r$. 
    The extended polynomial set $\prfx{a}\!\F_a^\prime$ is obtained by multiplying each $f_i \in \prfx{a}\!\F_a$ with the monomials in $\prfx{a}\!T_i$. In the step $4$, the set of monomials is obtained $\prfx{a}\!B = \text{mon}(\F_a^\prime)$.
    
    \item In the step $5$, the required monomial sets are
    \begin{equationarray}{l}
        B_r = \lbrace \lambda \mon{\alpha} \mid \mon{\alpha} \in B_1 \rbrace \nonumber  \\
        B_e = B_2.
    \end{equationarray} 
    Note that the set of monomials $\prfx{a}\!B = B_e \sqcup B_r \sqcup B_a$, and $B_a \cap B_r = \emptyset$.
    
    \item In the step $6$, the monomial vectors are set as $\V{b}_a=\V{b}_1, \V{b}_r=\text{vec}(B_r)$ and $\V{b}_e=\V{b}_2$. This will also fix the vector of monomials $\prfx{a}\V{b} = \begin{bmatrix}
    \V{b}_e \\ \V{b}_r \\ \V{b}_a
    \end{bmatrix}$. 
    The monomials in the vector $\prfx{a}\V{b}$ in the action matrix method are ordered in the same way as the monomials in the \srs method in Sec.~\ref{subsec:sparse_res_outline}. Moreover, as the monomials in $B_r$ are in a one-to-one correspondence w.r.t. the monomials in $B_1$, the monomial ordering in $\V{b}_1$ fixes the order of the monomials in $\V{b}_r$ as well.
   \end{enumerate} \vspace{10px}

    \noindent Applying these changes to the steps $1$-$6$ in Sec.~\ref{subsec:action_matrix_outline}, \ie~the offline stage, we obtain the elimination template $\prfx{a}\hat{\M{C}}$.
    Then, the action matrix-based solver (step $7$ in Sec.~\ref{subsec:action_matrix_outline}) computes the G-J elimination of $\prfx{a}\hat{\M{C}}$, from which the entries of the action matrix $\M{M}_f$ are read-off. This implies that the action matrix-based solver is equivalent to the \srs-based solver, which is proved in Proposition~\ref{prop:Mf_frm_X_alt}.

\section{Experiments}\label{sec:experiments}
\noindent In this section, we evaluate the performance of the minimal solvers, generated using our \srs method, proposed in Sec.~\ref{sec:extra_poly_sparse_res}. We compare the stability as well as the computational complexities of these solvers 
with the state-of-art \gb-based solvers for many interesting minimal problems. 
The minimal problems selected for comparison, represent a huge variety of relative and absolute pose problems and correspond to those studied in~\citep{DBLP:conf/cvpr/LarssonOAWKP18}. Note that for the two minimal problems, Rel. pose $f$+E+$f$ 6pt and Abs. pose refractive P5P, we generated solvers from a simplified polynomial system as described in~\citep{MartyushevVP2022}, instead of the original formulation. In comparison to the SOTA \gb-based action matrix methods such as~\citep{larsson2017efficient,DBLP:conf/cvpr/LarssonOAWKP18,MartyushevVP2022}, the recent method for laurent polynomial systems~\citep{martyushev2023automatic} generates a minimal solver by iteratively testing different elimination template matrices for rank conditions and does not rely on \gb computations. As the goal in this paper is to compare our \srs method with the \gb-based action matrix method, we have only compared the performance of our minimal solvers with those based on the three \gb methods~\citep{larsson2017efficient,DBLP:conf/cvpr/LarssonOAWKP18,MartyushevVP2022}.

\subsection{Computational complexity}\label{subsec:comp_complexity}
\noindent The biggest contributor towards computational complexity of a minimal solver is the size of the matrices that undergo crucial numerical operations. The solvers based on our \srs method in Sec.~\ref{sec:extra_poly_sparse_res}, the state-of-the-art \gb solvers as well as in the original solvers proposed by the respective authors (see column $4$) involve two crucial operations, a \textit{matrix inverse}\footnote{G-J elimination is usually performed by a computing a matrix inverse.} and an \textit{eigenvalue decomposition}. This is indicated by the size of the solver in the table, \eg~a \srs-based solver of size $11 \times 20$ means computing a matrix inverse $\hat{\M{A}}_{12}^{-1}$ of size $11 \times 11$, ultimately leading to a Schur complement $\M{X}$ of size $20-11 =9$, and an eigenvalue decomposition of this matrix $\M{X}$. Similarly an action matrix-based solver of size $11 \times 20$ means performing a G-J elimination of a $11 \times 20$ matrix $\M{C}$ and then an eigenvalue decomposition of an action matrix $\M{M}_f$ of size $20-11=9$. So in Tab.~\ref{tbl:sizecomparison}, we compare the size of the upper block $\begin{bmatrix} \hat{\M{A}}_{11} & \hat{\M{A}}_{12} \end{bmatrix}$ of the \srs matrix $\Mres$ in our extra polynomial \srs-based solvers, with the size of the elimination template $\M{C}$ used in state-of-the-art \gb solvers as well as in the original solvers proposed by the respective authors (column $4$).

The \gb-based solvers used for comparison include the solvers generated using the approach in~\citep{larsson2017efficient} (column $5$), the \gf-based and the heuristic-based approaches in~\citep{DBLP:conf/cvpr/LarssonOAWKP18} (columns $6$ and $8$ respectively), and the template optimization approach using a greedy parameter search in~\citep{MartyushevVP2022} (column $9$). Out of the two methods for generating minimal solvers, proposed in~\citep{MartyushevVP2022}, we consider the smallest solver for each minimal problem.

As we can see from the Tab.~\ref{tbl:sizecomparison}, for most minimal problems, our \srs method leads to smaller solvers compared to the \gb solvers based on~\citep{larsson2017efficient,DBLP:conf/cvpr/LarssonOAWKP18} and of exactly the same size as the solvers based on~\citep{MartyushevVP2022}. The smallest solvers where size of the eigenvalue decomposition problem is the same as the number of solutions, are written in bold in the Tab.~\ref{tbl:sizecomparison}.

Moreover, for some minimal problems, our \srs method leads to a larger eigenvalue problem than the number of solutions, written in blue in Tab.~\ref{tbl:sizecomparison}. For three such minimal problems, \ie~Rel. pose E+$f\lambda$ 7pt, Unsynch. Rel. pose, and Optimal pose 2pt v2, our \srs method leads to solvers with the smaller size as compared to the solvers based on the \gb-based methods~\citep{larsson2017efficient,DBLP:conf/cvpr/LarssonOAWKP18}, whereas for the problem, Abs. pose quivers, our \srs method leads to a smaller solver than the solvers based on~\citep{larsson2017efficient,DBLP:conf/cvpr/LarssonOAWKP18} and of comparable size w.r.t. the solver based on~\citep{MartyushevVP2022}.

Note that for the two minimal problems, Optimal pose 2pt v2 and Rel. pose E angle+4pt, the elimination template $\M{C}$ in the solvers based on the approach in~\citep{MartyushevVP2022}, underwent an extra Schur complement-based step in the offline stage. Therefore, the template sizes reported for these two solvers are smaller than those in the solvers based on the other \gb methods in~\citep{larsson2017efficient,DBLP:conf/cvpr/LarssonOAWKP18}. For the problem, Rel. pose E angle+4pt, our \srs method failed to generate a solver.

Note that here we do not compare our solvers' sizes with the resultant-based solvers generated by~\citep{DBLP:conf/iccv/Heikkila17} and~\citep{emiris-general}. These methods can not be directly applied to most of the studied minimal problems as they can not handle more equations than unknowns. Though these methods can be adjusted such that they work with more equations with unknowns, and generate full-rank matrices, even then they lead to larger solvers than the new method proposed in the Sec.~\ref{subsec:sparse_res_outline}. With the method in~\citep{DBLP:conf/iccv/Heikkila17}, we also failed to generate full rank solvers for some minimal problems while for some other problems, the generated solvers were larger than ours, and GEP involved in~\citep{DBLP:conf/iccv/Heikkila17} led also to many unwanted solutions.

\subsection{Stability comparison}\label{subsec:comp_stability}
\noindent We evaluate and compare the stability of our solvers from Tab.~\ref{tbl:sizecomparison} with the \gb-based solvers. As it is not feasible to generate scene setups for all considered problems, we instead evaluated the stability of minimal solvers using $5$K instances of random data points. Stability measures include \textit{mean} and \textit{median} of $Log_{10}$ of the normalized equation residuals for computed solutions, as well as the solver failures. The solver failures are an important measure of stability for real world applications, and are computed as the $\%$ of $5$K instances for which at least one solution has a normalized equation residual $>10^{-3}$. These measures on randomly generated inputs have been shown to be sufficiently good indicators of solver stability~\citep{larsson2017efficient}. Also, Tab.~\ref{tbl:stabcomp} shows the stability of the solvers for all the minimal problems from Tab.~\ref{tbl:sizecomparison}.
We observe that for most of the minimal problems, our proposed \srs-based solvers have no failures. Among those with some failures,  for all except two minimal problems our \srs-based solvers have fewer failures, as compared to the \gb-based solvers. 
 
\begin{figure*}[t]
\centering
\includegraphics[width=0.32\linewidth]{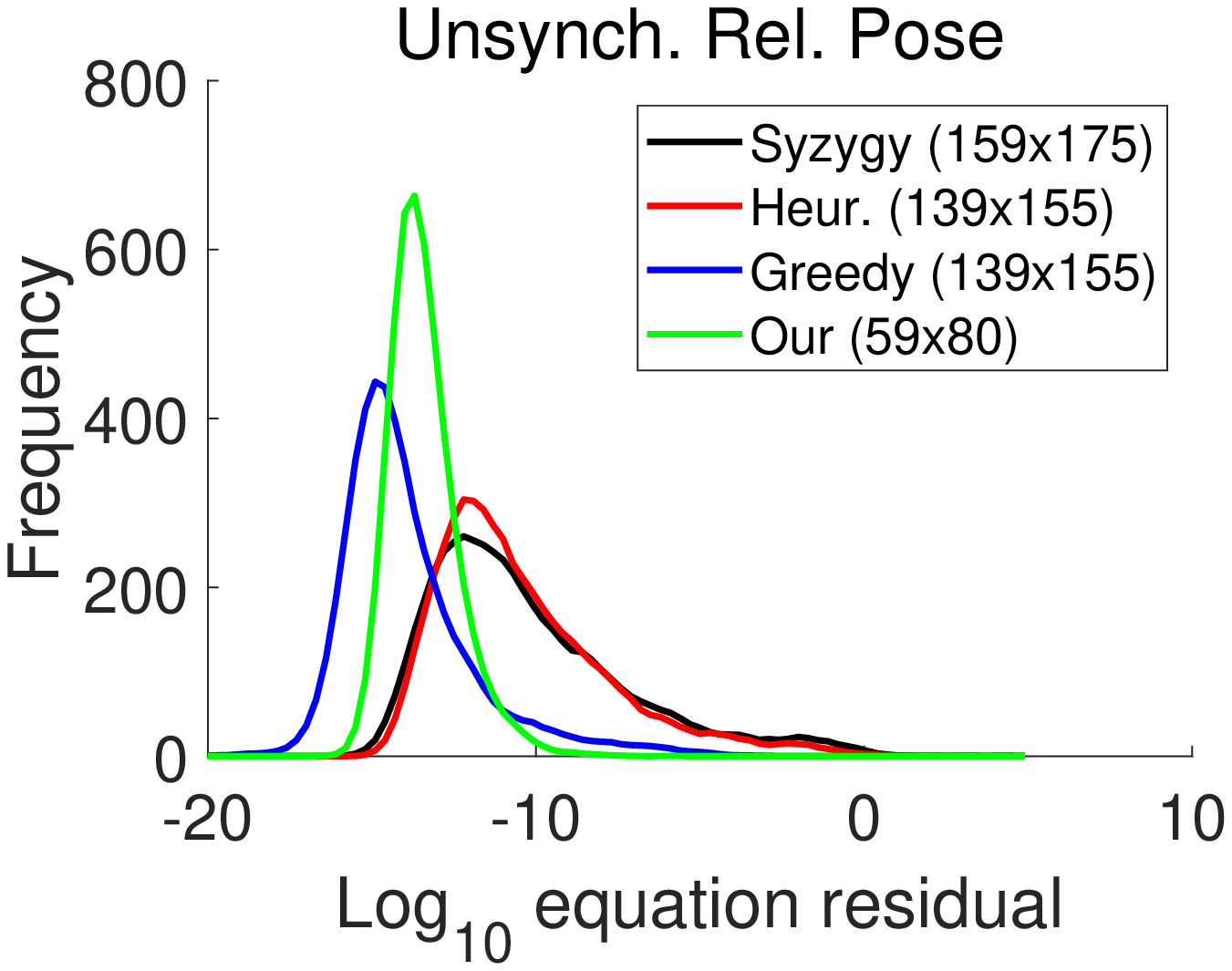}
\includegraphics[width=0.32\linewidth]{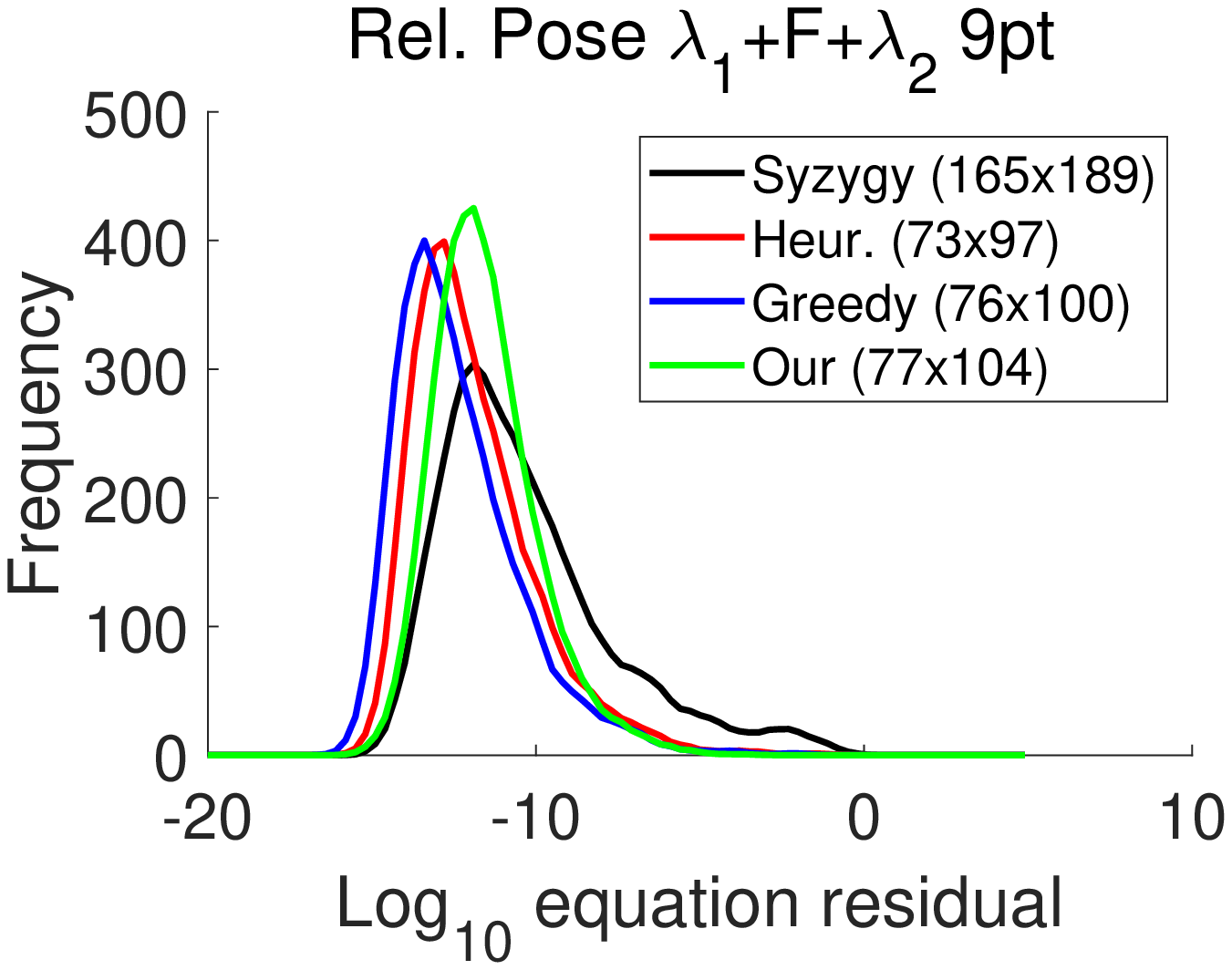}
\includegraphics[width=0.32\linewidth]{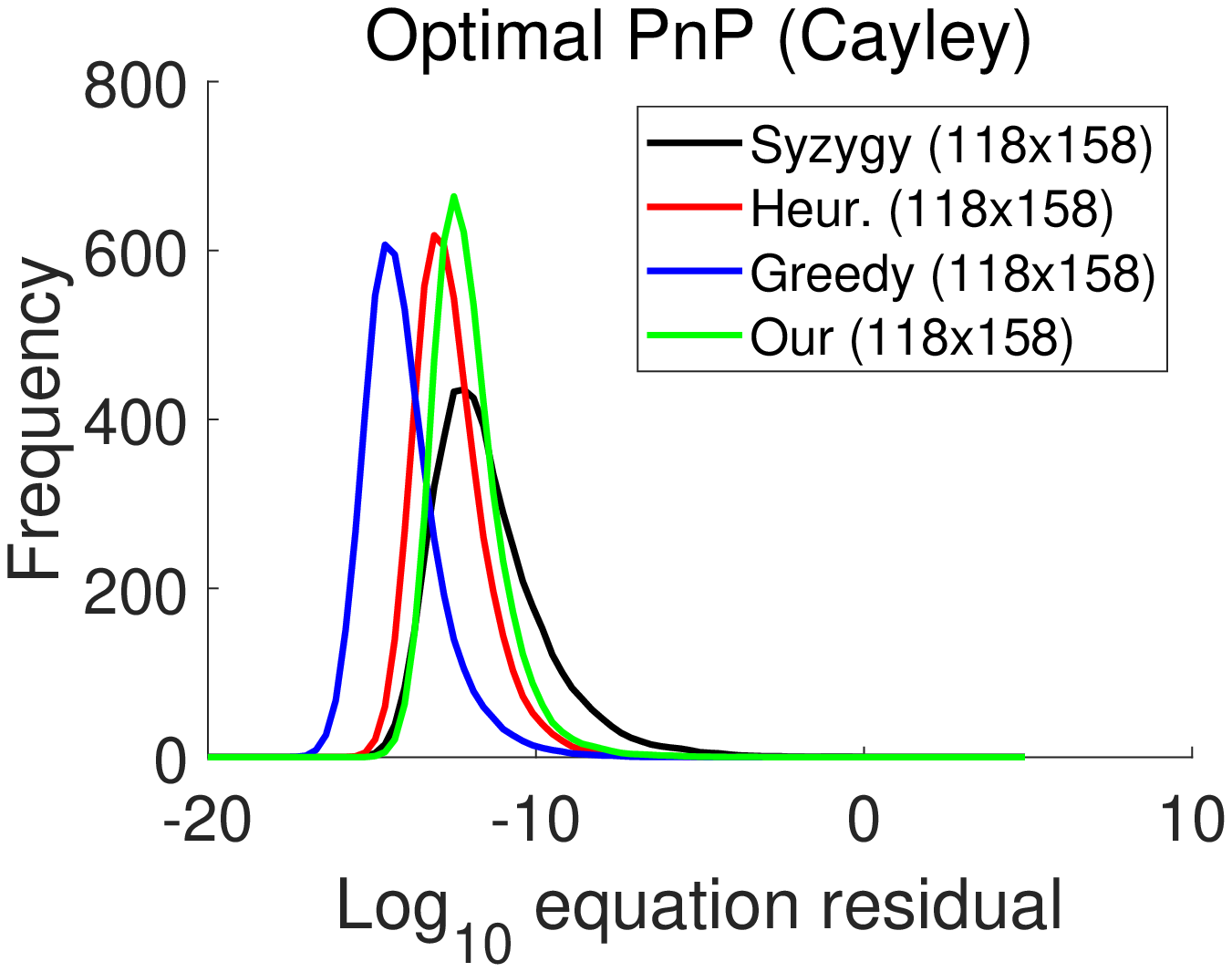} \vspace{5px}\\
\includegraphics[width=0.32\linewidth]{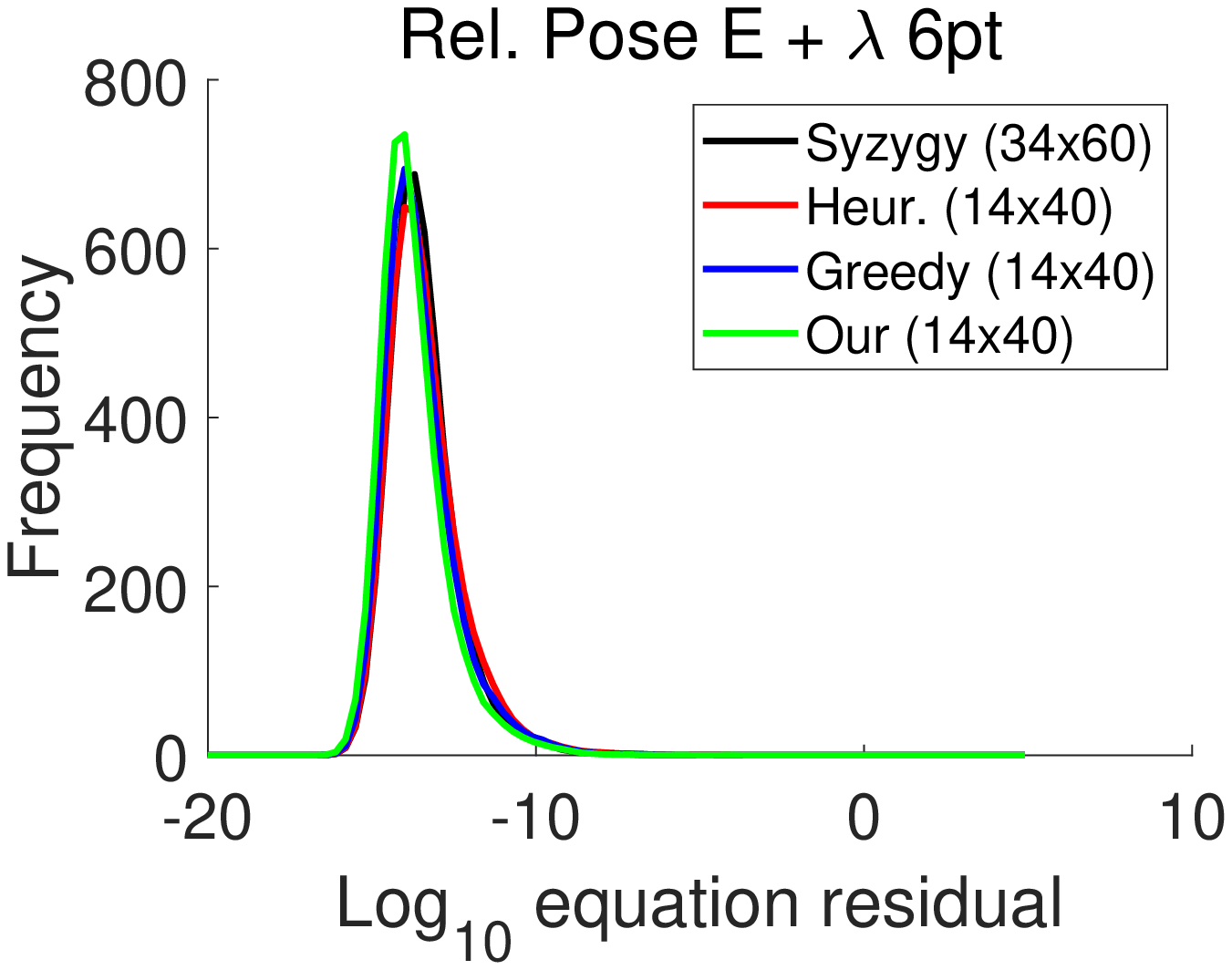}
\includegraphics[width=0.32\linewidth]{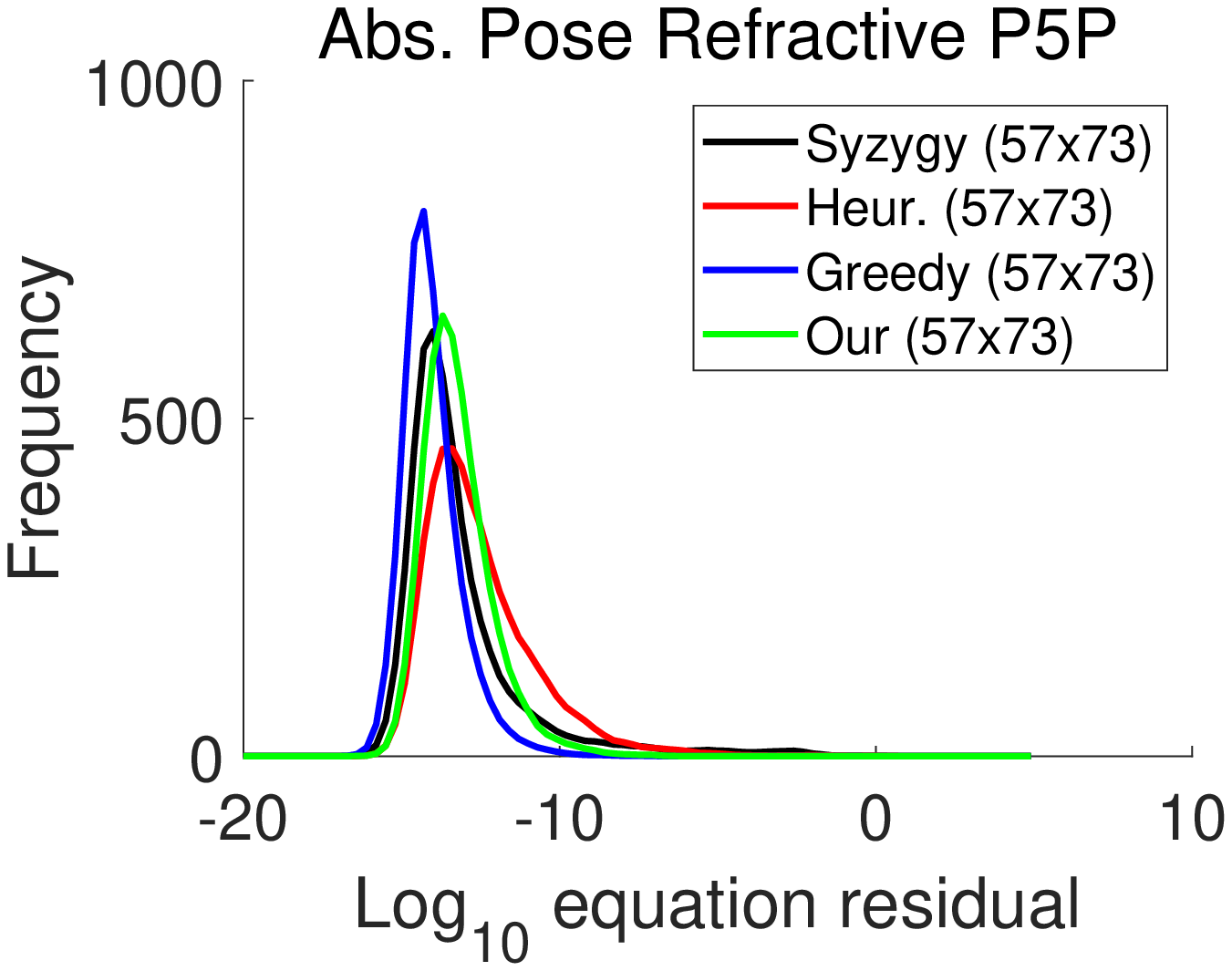}
\includegraphics[width=0.32\linewidth]{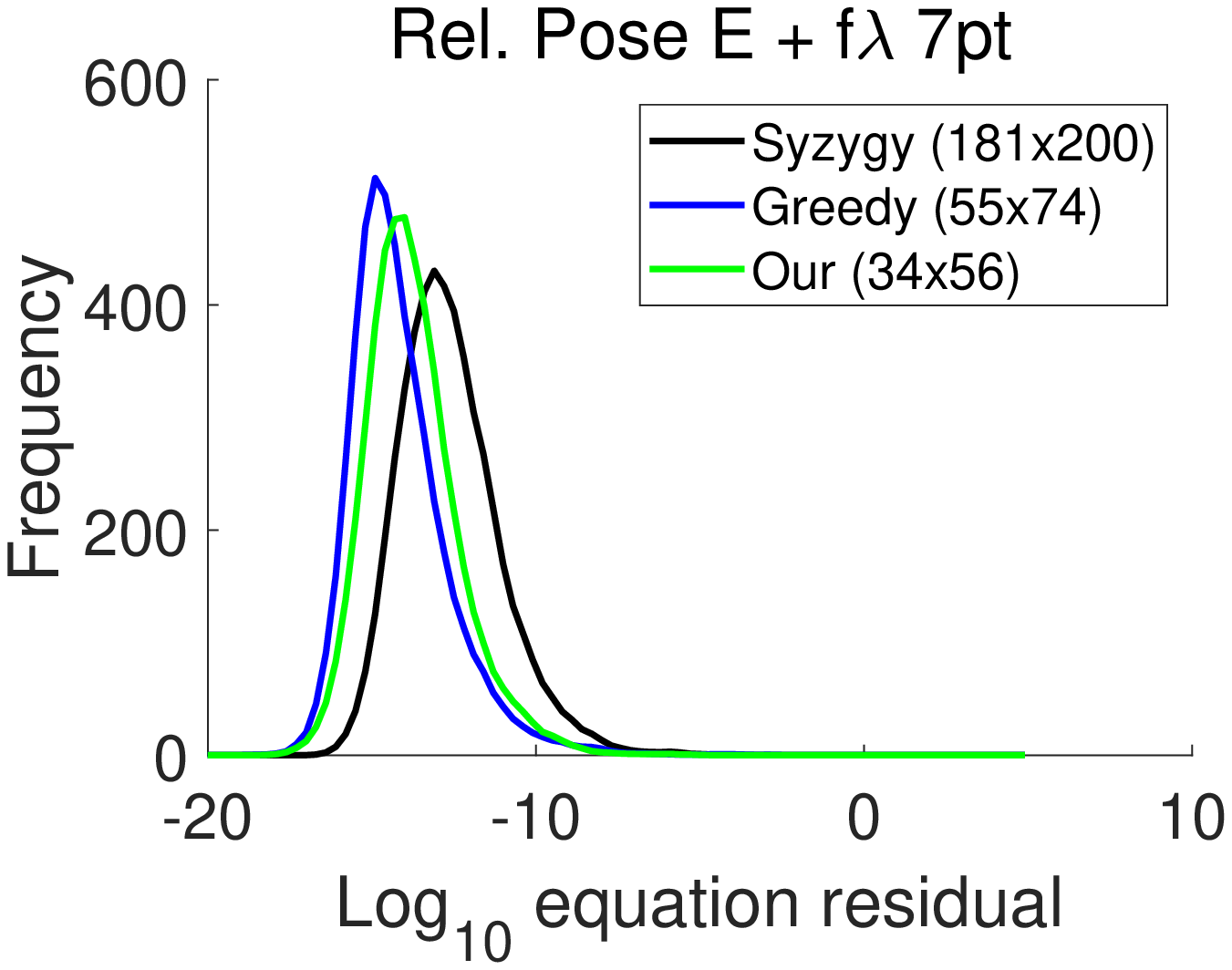} \vspace{5px}\\
\includegraphics[width=0.32\linewidth]{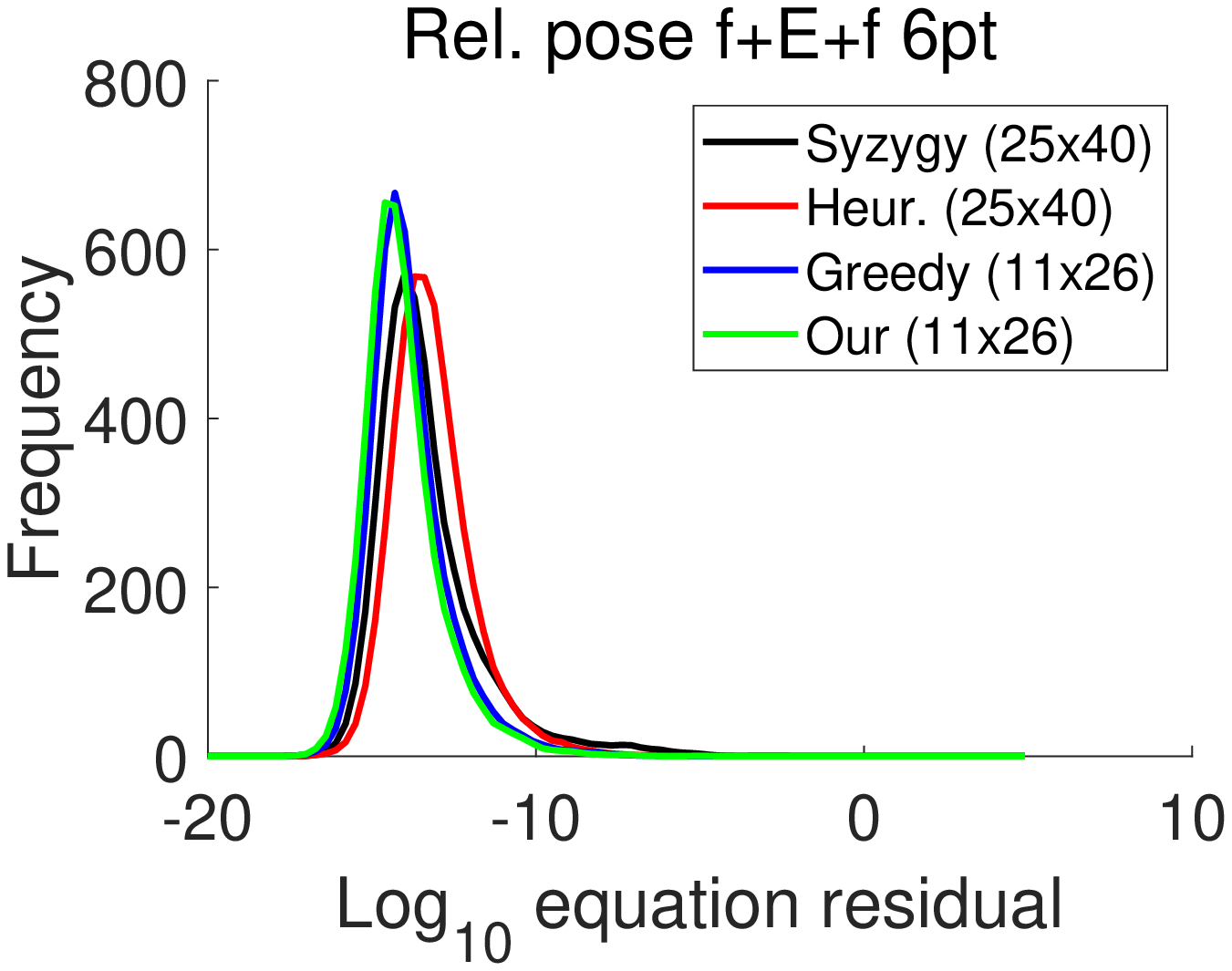}
\includegraphics[width=0.32\linewidth]{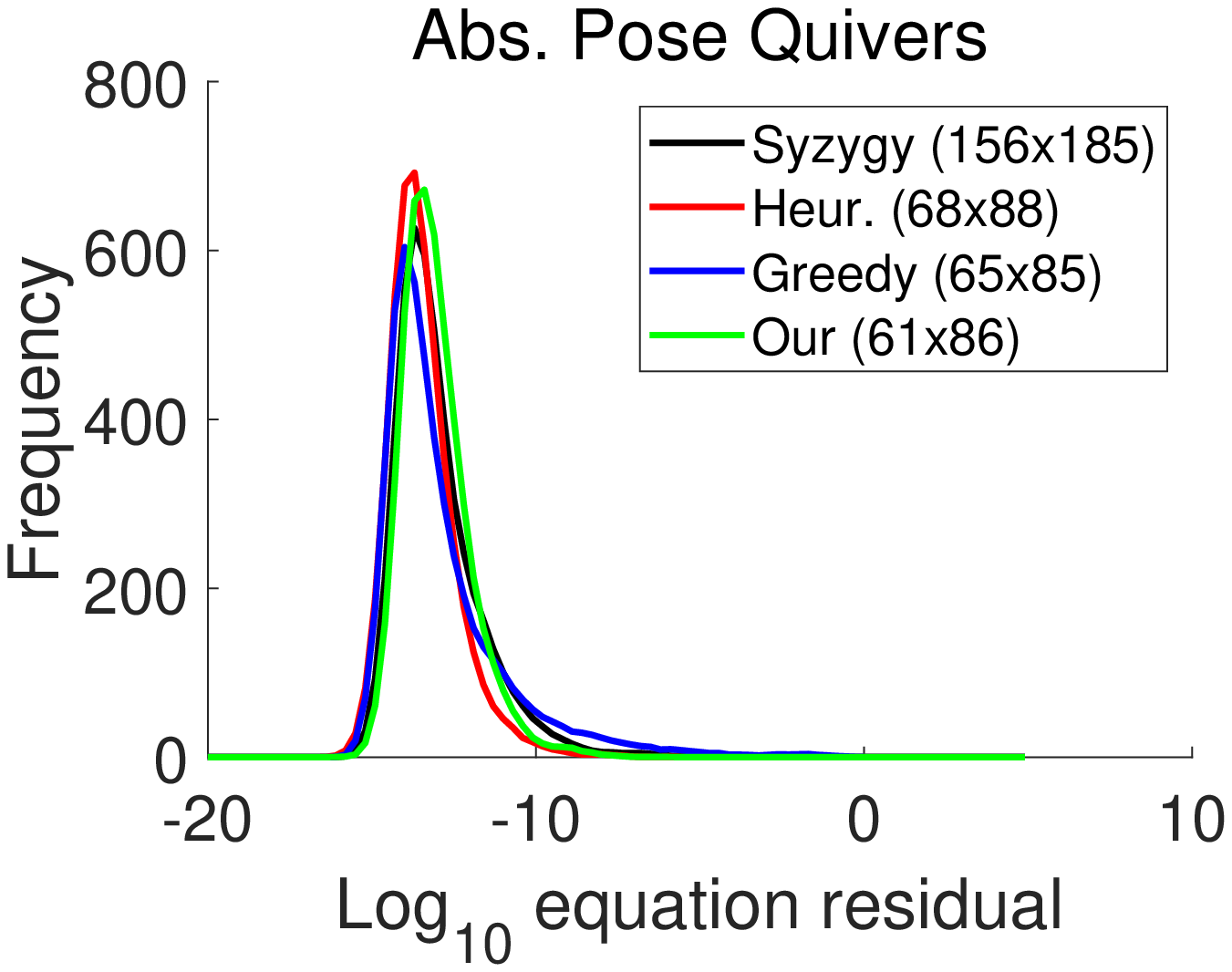}
\includegraphics[width=0.32\linewidth]{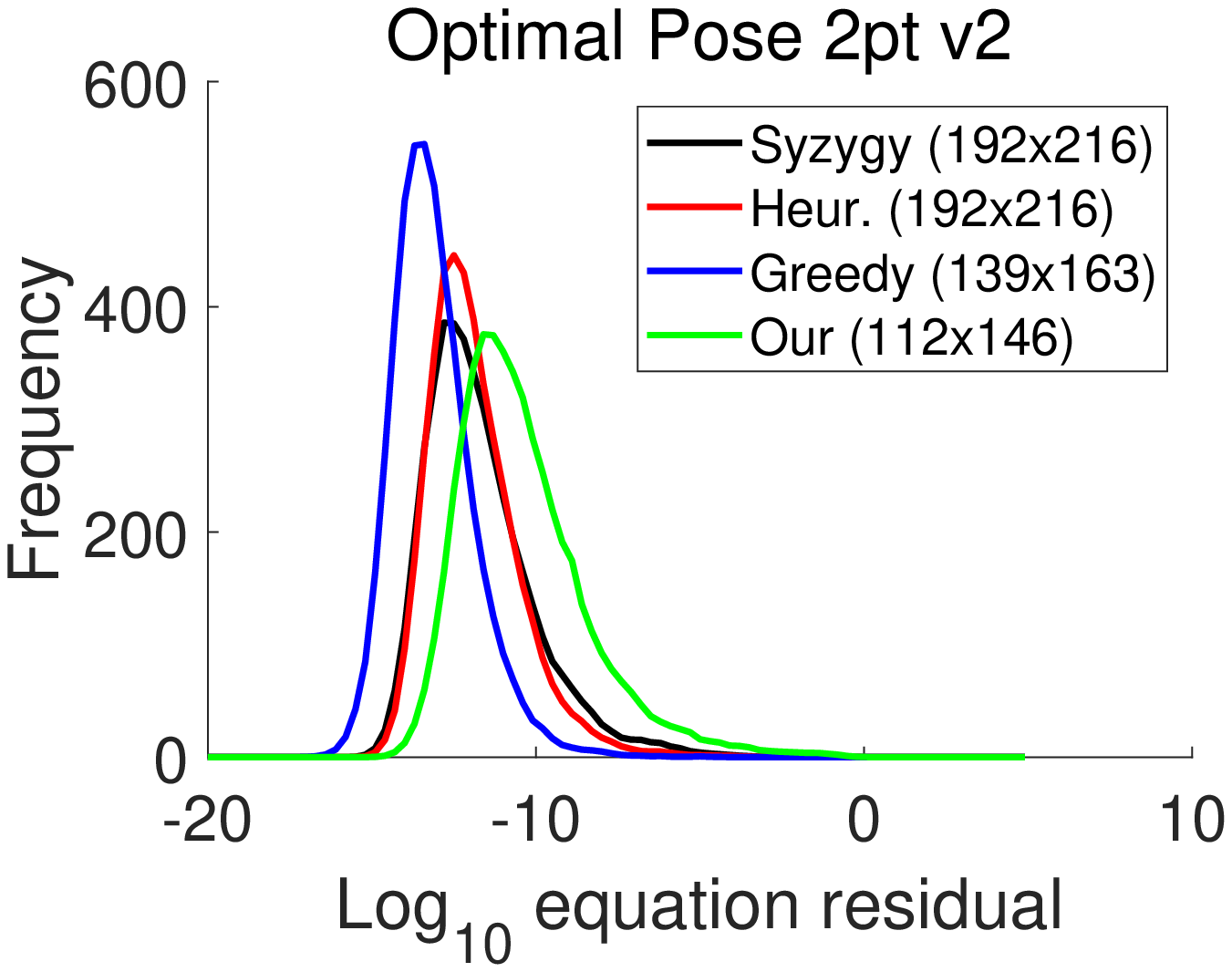} \vspace{5px}\\
\caption{Histograms of $Log_{10}$ of normalized equation residual error for nine selected minimal problems. The methods tested for comparison are based on \gb~\citep{larsson2017efficient}, heuristics~\citep{larsson2017efficient}, greedy parameter search~\citep{MartyushevVP2022} and our \srs method.   
}
\label{fig:histograms}
\end{figure*}

In general, our new method generates solvers that are stable with only very few failures. For a selected subset of nine minimal problems from the Tab.~\ref{tbl:sizecomparison}, we computed the $Log_{10}$ of the normalized equation residuals and depicted their histograms in the Fig.~\ref{fig:histograms}. The histograms agree with our observation from the Tab.~\ref{tbl:stabcomp}, that our proposed \srs method leads to solvers with only few failures.

\begin{table*}
\centering
\resizebox{0.95\linewidth}{!}{\begin{tabular}{c l c c c c c c c c}
\toprule
$\#$ & Problem & Our & Author & Original & Syzygy  & GFan & (\#GB) & Heuristic & Greedy  \\ \midrule

$1$ & Rel. pose F+$\lambda$ 8pt\textsuperscript{$(\ddagger)$} (\small{$8$ sols.})          & $\color{blue}{7 \times 16}$ & \citep{kuang2014minimal} & $12\times 24$  & $ 11\times 19 $  & $ 11\times 19 $  & $(10)$ & $\bf  7 \times 15 $ & $\bf 7 \times 15$\\ 
$2$ & Rel. pose E+$f$ 6pt (\small{$9$ sols.})       & $\bf 11 \times 20$ & \citep{bujnak20093d} &  $21 \times 30$ & $19 \times 28 $  & $\bf  11\times 20 $  & $(66)$ & $\bf  11\times 20 $  & $\bf 11 \times 20$\\ 
$3$ & Rel. pose $f$+E+$f$ 6pt\textsuperscript{*} (\small{$15$ sols.})    & $\bf  11 \times 26$ & \citep{Kukelova-ECCV-2008}  & $31 \times 46$ & $ 25\times 40 $  & $ 31\times 46 $  & $(218)$ & $ 25 \times 40 $  & $\bf 11 \times 26$\\ 
$4$ & Rel. pose E+$\lambda$ 6pt (\small{$26$ sols.})  & $\bf 14 \times 40$ & \citep{kuang2014minimal}   & $48 \times 70 $  & $ 34\times 60 $  & $ 34\times 60 $  & $(846)$ & $\bf  14\times 40 $  & $\bf 14 \times 40$\\ 
$5$ & Stitching $f\lambda$+R+$f\lambda$ 3pt (\small{$18$ sols.})  & $\color{blue}{8\times 31} $ & \citep{naroditsky2011optimizing}& $54 \times 77$ & $ 48\times 66 $  & $ 48\times 66 $  & $(26)$ & $\bf  18\times 36 $  & $\bf 18 \times 36$\\ 
$6$ & Abs. Pose P4Pfr (\small{$16$ sols.}) & $\bf  52\times 68 $ & \citep{bujnak2010new} & $136 \times 152 $ & $ 140\times 156$  & $54\times 70 $  & $(1745)$ & $54\times 70$  & $\bf  52\times 68$\\ 
$7$ & Abs. Pose P4Pfr \small{(elim. $f$)} (\small{$12$ sols.}) & $\bf  28\times 40 $ &  \citep{larsson2017making} &  $\bf  28\times 40 $ & $ 58 \times 70$  & $\bf 28 \times 40$  & $(699)$ & $\bf 28 \times 40$  & $\bf 28 \times 40$\\ 
$8$ & Rel. pose $\lambda$+E+$\lambda$ 6pt\textsuperscript{$(\ddagger)$} (\small{$52$ sols.}) & $\color{blue}{39 \times 95}$ & \citep{Kukelova-ECCV-2008} & $238 \times 290$  & $ 154\times 206$  & - & ? & $  53\times 105$  & $\color{blue}{39 \times 95}$\\ 
$9$ & Rel. pose $\lambda_1$+F+$\lambda_2$ 9pt (\small{$24$ sols.}) & $\color{blue}{77 \times 104}$ &  \citep{Kukelova-ECCV-2008} & $179 \times 203$ & $ 165\times 189$  & $  105 \times 129 $  & $(6896)$ & $\bf  73\times 97$  & $ 76\times 100$  \\ 
$10$ & Rel. pose E+$f\lambda$ 7pt (\small{$19$ sols.}) & $\color{blue}{34 \times 56}$  & \citep{kuang2014minimal}& $200\times 231 $ & $181\times 200$  & $69\times 88 $  & $(3190)$ & $  65\times 84 $  & $ 55 \times 74$\\ 
$11$ & Rel.\ pose E+$f\lambda$ 7pt \small{(elim.\ $\lambda$)} (\small{$19$ sols.}) &$\bf  22\times 41 $  & - & $ 52\times 71 $  & $ 37\times 56 $  & $(332)$ & $24\times 43 $  & $\bf 22 \times 41$\\ 
$12$ & Rel.\ pose E+$f\lambda$ 7pt \small{(elim.\ $f\lambda$)}  (\small{$19$ sols.}) & $\bf 51 \times 70$  & \citep{kukelova2017clever}  & $\bf 51 \times 70$ & $\bf  51\times 70 $  & $\bf  51\times 70 $  & $(3416)$ & $\bf  51\times 70 $  & $\bf 51 \times 70$\\ 
$13$ & Rolling shutter pose (\small{$8$ sols.}) & $\color{blue}{40 \times 52}$ & \citep{saurer2015minimal} & $48\times 56$  & $\bf  47 \times 55$   & $\bf  47 \times 55$   & ($520$) & $\bf  47 \times 55 $  & $\bf 47 \times 55$\\ 
$14$ & Triangulation from satellite im. (\small{$27$ sols.}) & $\bf 87 \times 114$ & \citep{Zheng2015satimag} & $93\times 120$  & $88 \times 115$   & $88 \times 115$   & ($837$) & $88 \times 115$  & $\bf 87 \times 114$\\ 
$15$ & Abs. pose refractive P5P\textsuperscript{*}  (\small{$16$ sols.}) & $\bf 57 \times 73$ &  \citep{haner2015absolute} & $280\times 399 $& $\bf 57 \times 73$  & $\bf 57 \times 73$  & $(8659)$ & $\bf 57 \times 73$  & $\bf 57 \times 73$\\ 
$16$ & Abs. pose quivers (\small{$20$ sols.}) & $\color{blue}{37 \times 71}$  & \citep{kuang2013pose} & $372\times 386 $ & $ 156\times 185$  & - & ? & $ 68\times 88$  & $\bf 65 \times 85$\\ 
$17$ & Unsynch. Rel. pose\textsuperscript{$(\ddagger)$} (\small{$16$ sols.}) & $\color{blue}{59 \times 80}$ & \citep{DBLP:journals/corr/AlblKFHSP17} & $633 \times 649$ & $159\times 175$ & - & ? & $139\times 155$  & $ 139 \times 155$\\ 
$18$ & Optimal PnP (Hesch) (\small{$27$ sols.}) & $\bf 87 \times 114$ & \citep{Hesch2011}& $93\times 120$  & $88 \times 115$   & $88 \times 115$   & ($837$) & $88 \times 115$  & $\bf 87 \times 114$\\ 
$19$ & Optimal PnP (Cayley) (\small{$40$ sols.}) & $\bf 118 \times 158$ & \citep{Nakano15} & $\bf 118 \times 158$  & $\bf 118 \times 158$   & $\bf 118 \times 158$   & $(2244)$ & $\bf 118 \times 158$  & $\bf 118 \times 158$\\ 
$20$ & Optimal pose 2pt v2 (\small{$24$ sols.}) & $\color{blue}{112 \times 146}$ & \citep{svarm2017city} & $192 \times 216$ & $192 \times 216$ & $-$ & ? & $192 \times 216$  & $ 139 \times 163$\textsuperscript{$\dagger$} \\ 
$21$ & Rel. pose E angle+4pt (\small{$20$ sols.})  & - &  \citep{li20134} & $ 270\times 290 $ & $ 266\times 286$  & - & ? & $ 183\times 203$ & $ 99 \times 119$\textsuperscript{$\dagger$} \\ 
\bottomrule
\end{tabular}
}
\caption{ Size comparison of solvers based on~\citep{larsson2017efficient} (\textbf{col 5}), \gf~\citep{DBLP:conf/cvpr/LarssonOAWKP18} (\textbf{col 6}), heuristic sampling~\citep{DBLP:conf/cvpr/LarssonOAWKP18} (\textbf{col 8}) and greedy parameter search~\citep{MartyushevVP2022} (\textbf{col 9}). The smallest solver sizes are written in bold, and the solver sizes with larger eigenvalue problem than the number of solutions is written in blue. Missing entries are when we failed to generate a solver. $(*)$: Using a simplified input system of polynomials, as in~\citep{MartyushevVP2022}. $(\ddagger)$: Solved using the alternate eigenvalue formulation~\eqref{eq:eig_prb_alt}. $(\dagger)$: In the offline stage in~\citep{MartyushevVP2022}, the elimination template was reduced using a Schur complement.}
\label{tbl:sizecomparison}
\end{table*}

\begin{table*}
\centering
\normalsize
\resizebox{\linewidth}{!}{\begin{tabular}{c l c c c c c c c c c c c c } \toprule
$\#$ & Problem &  \multicolumn{3} {c} {Our} &\multicolumn{3} {c} {Syzygy} & \multicolumn{3} {c} {Heuristic} &  \multicolumn{3} {c}{Greedy}  \\
\cmidrule(r){3-5} \cmidrule(r){6-8} \cmidrule(r){9-11} \cmidrule(r){12-14}
& & mean & med. & fail($\%$) & mean & med. & fail($\%$) & mean & med. & fail($\%$) & mean & med. & fail($\%$)  \\\midrule
 $1$ &Rel. pose F+$\lambda$ 8pt\textsuperscript{$(\ddagger)$}  & $-15.35$ & $-15.38$ & $\bf 0$ & $-14.31$ & $-14.63$ & $\bf 0$ &  $-14.18$ & $-14.48$ & $\bf 0$ & $-15.07$  & $-14.46$ & $\bf 0$  \\
$2$ & Rel. pose E+$f$ 6pt  & $ -13.99 $&$ -14.26  $&$\bf 0$ & $ -13.36 $&$ -13.64 $ & $\bf 0$ & $-13.05$ & $-13.34$ & $\bf 0$ & $-13.43$& $-13.79  $&$\bf 0$  \\
$3$ & Rel. pose $f$+E+$f$ 6pt\textsuperscript{*}  &$ -14.38 $&$ -14.60 $&$\bf 0 $&$ -13.60 $&$ -14.00  $&$  0.04$  & $-13.87$ & $-14.07$ & $\bf 0$ & $-14.15 $&$  -14.41  $ & $ \bf 0$  \\
$4$ & Rel. pose E+$\lambda$ 6pt  &$ -14.07$ &$   -14.27$ &$\bf 0$  &$ -13.75 $&$ -13.94 $&$\bf 0$ &$  -13.13 $&$ -13.34  $&$  0.04$ & $-13.85 $&$  -14.08 $ & $\bf 0$  \\
$5$ & Stitching $f\lambda$+R+$f\lambda$ 3pt  &$ -14.30$ &$ -14.46$ &$\bf 0 $&$ -14.60$ &$ -14.37$ &$ \bf 0 $&$ -13.20$ &$ -13.46$ &$ \bf 0$   & $-14.30 $&$  -14.51 $ & $\bf 0$  \\
$6$ & Abs. Pose P4Pfr \small{(elim. $f$)} & $-13.62$ & $-13.82$ & $\bf 0$ & $ -13.53 $ & $ -13.77$ & $\bf 0 $ & $ -12.73$ & $ -13.00 $ & $\bf 0$   & $-13.87 $ & $ -14.13$ & $\bf 0$  \\
$7$ & Rel. pose $\lambda$+E+$\lambda$ 6pt  &$ -11.86 $&$ -12.33$ &$0.06$ &$ -7.50 $&$-7.78$&$5.16$ &$-10.53$&$-10.89$ & $0.40$& $-13.50 $&$ -13.95 $ & $\bf 0$  \\
$8$ & Rel. pose $\lambda_1$+F+$\lambda_2$ 9pt  &$ -11.91 $ & $-12.14$ &$\bf 0.02 $&$-10.63 $&$ -11.22 $&$   1.94$ &$-12.67 $&$ -13.08  $&$0.10$& $-12.56 $&$ -13.16  $&$  0.40$  \\
$9$ & Rel. pose E+$f\lambda$ 7pt  &$ -14.05 $ & $  -14.23$ & $\bf 0$   &$ -12.90 $&$ -13.11$&$ \bf    0$ & $-$ & $-$ & $-$ & $-14.45$ & $-14.75$ & $\bf 0$  \\ 
$10$ & Rel. pose E+$f\lambda$ 7pt \small{(elim.\ $\lambda$)}   &$ -12.53$ &$ -12.95$ &$\bf 0.02$   &$-12.59 $&$ -12.92  $&$  0.04$ &$-12.53  $&$ -12.76 $ & $\bf  0.02$& $-12.52 $&$ -12.80  $&$\bf  0.02$  \\
$11$ & Rel. pose E+$f\lambda$ 7pt \small{(elim.\ $f\lambda$)}  &$-13.37$ & $-13.66$ &$ 0.04$   &$-12.60 $&$ -12.98$&$   0.06$ &$-13.59 $ & $ -13.78 $ & $\bf0$ & $-13.10$ & $  -13.66$ & $ 0.1$  \\
$12$ & Rolling shutter pose &$-13.67 $ & $-13.83$ &$\bf  0 $ &$ -14.52$ &$ -14.72$ &$\bf 0$ &$-12.43$&$-12.65$&$\bf 0$   & $-$ & $-$ & $-$  \\
$13$ & Triangulation from satellite im. &$-13.12$&$-13.03$ &$\bf 0 $ &$-12.67 $&$ -13.06  $&$\bf 0$ &$-11.61$&$-11.93$&$0.50$   & $-13.25 $  & $  -13.50$ & $\bf 0$  \\
 $14$ & Abs. pose refractive P5P\textsuperscript{*} &$-13.54 $&$  -13.77 $ &$\bf 0$ &$ -13.42 $&$ -14.05   $ & $0.48$ & $-12.82$ & $-13.26 $&$   0.20$& $-14.35$ & $-14.55$ & $\bf 0$  \\
 $15$ & Abs. pose quivers  &$  -13.35$&$ -13.57$ &$\bf 0 $ &$-13.29  $ & $-13.62 $ &$ 0.32$ &$ -13.74$ &$ -13.95$ &$\bf 0$& $-13.20$ & $-13.83$ & $0.24$  \\
 $16$ & Unsynch. Rel. pose\textsuperscript{$(\ddagger)$} &$-13.59 $ & $  -13.78$ &$\bf 0 $ &$-10.53$ & $-11.29$ & $3.43$&$-12.78 $&$  -13.36 $&$    0.36$ & $-12.28$ & $ -12.84$ & $0.44$  \\
 $17$ & Optimal PnP (Hesch) &$ -13.14 $&$-13.32$ &$ \bf 0 $ &$  -12.89  $&$-13.23   $&$ \bf 0$  &$-13.20 $&$ -13.48$&$ \bf 0$ & $-14.60$ &$-14.70$ & $\bf 0$  \\
 $18$ & Optimal PnP (Cayley) &$-12.36 $&$ -12.56$ &$\bf 0$ &$-11.33 $&$ -11.69  $&$\bf 0$ &$-12.88 $&$ -13.08$&$ \bf 0$&  $-14.30 $& $-14.52$ & $\bf 0$  \\
$19$ & Optimal pose 2pt v2 &$-10.68$ & $ -11.06$ & $0.52$  &$-12.01 $&$ -12.32  $&$ \bf 0$ &$-12.44$ & $-12.54$ & $ 0.05$& $ -12.01$ & $-12.21$ & $\bf 0$  \\
\bottomrule
\end{tabular}}
\caption{Stability comparison for solvers for the minimal problems from Tab.~\ref{tbl:sizecomparison}, generated by our new method, solvers generated using~\citep{larsson2017efficient} (\textbf{col 4}), heuristic-based solvers~\citep{DBLP:conf/cvpr/LarssonOAWKP18} (\textbf{col 5}) and greedy parameter search~\citep{MartyushevVP2022} (\textbf{col 6}). For each minimal problem, the solvers which led to minimum number of failures are written in bold. Mean and median are computed from $Log_{10}$ of normalized equation residuals. $(\dagger)$: In the offline stage in~\citep{MartyushevVP2022}, the elimination template was reduced using a Schur complement. $(-)$: Failed to extract solutions to all variables. $(*)$: Using a simplified input system of polynomials, as in~\citep{MartyushevVP2022}.}
\label{tbl:stabcomp}
\end{table*}

Note that as our new solvers are solving the same formulations of problems as the existing state-of-the-art solvers, the performance on noisy measurements and real data would be the same as the performance of the state-of-the-art solvers. The only difference in the performance comes from numerical instabilities that already appear in the noise-less case and are detailed in Tab.~\ref{tbl:stabcomp} (fail$\%$). For performance of the solvers in real applications we refer the reader to papers where the original formulations of the studied problems were presented (see Tab.~\ref{tbl:sizecomparison}, column $3$). Here we select two interesting problems, \ie, one relative and one absolute pose problem, and perform experiments on synthetically generated scenes and on real images, respectively.

\begin{figure}[h!]
\centering
\includegraphics[width=0.49\linewidth]{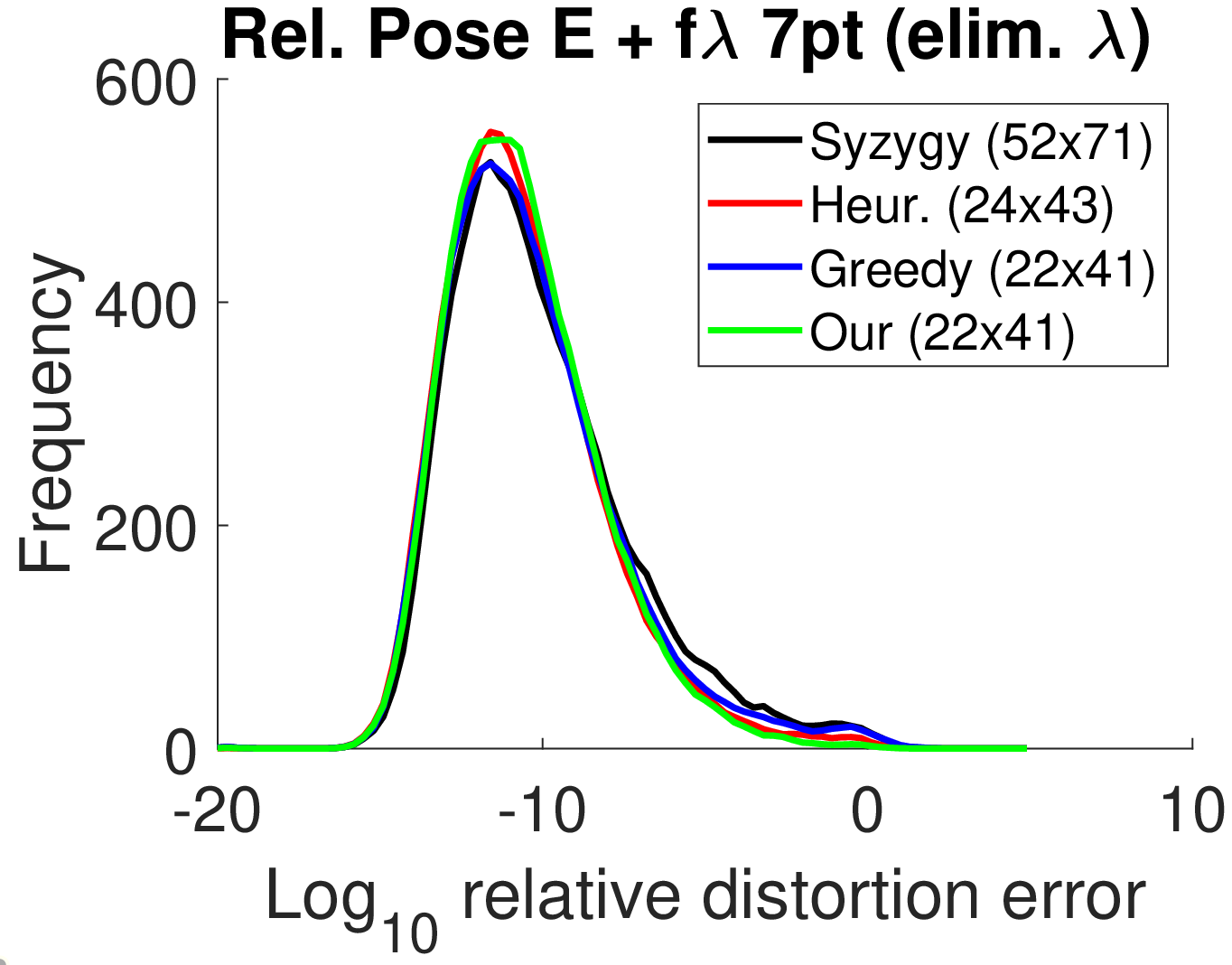}
\includegraphics[width=0.49\linewidth]{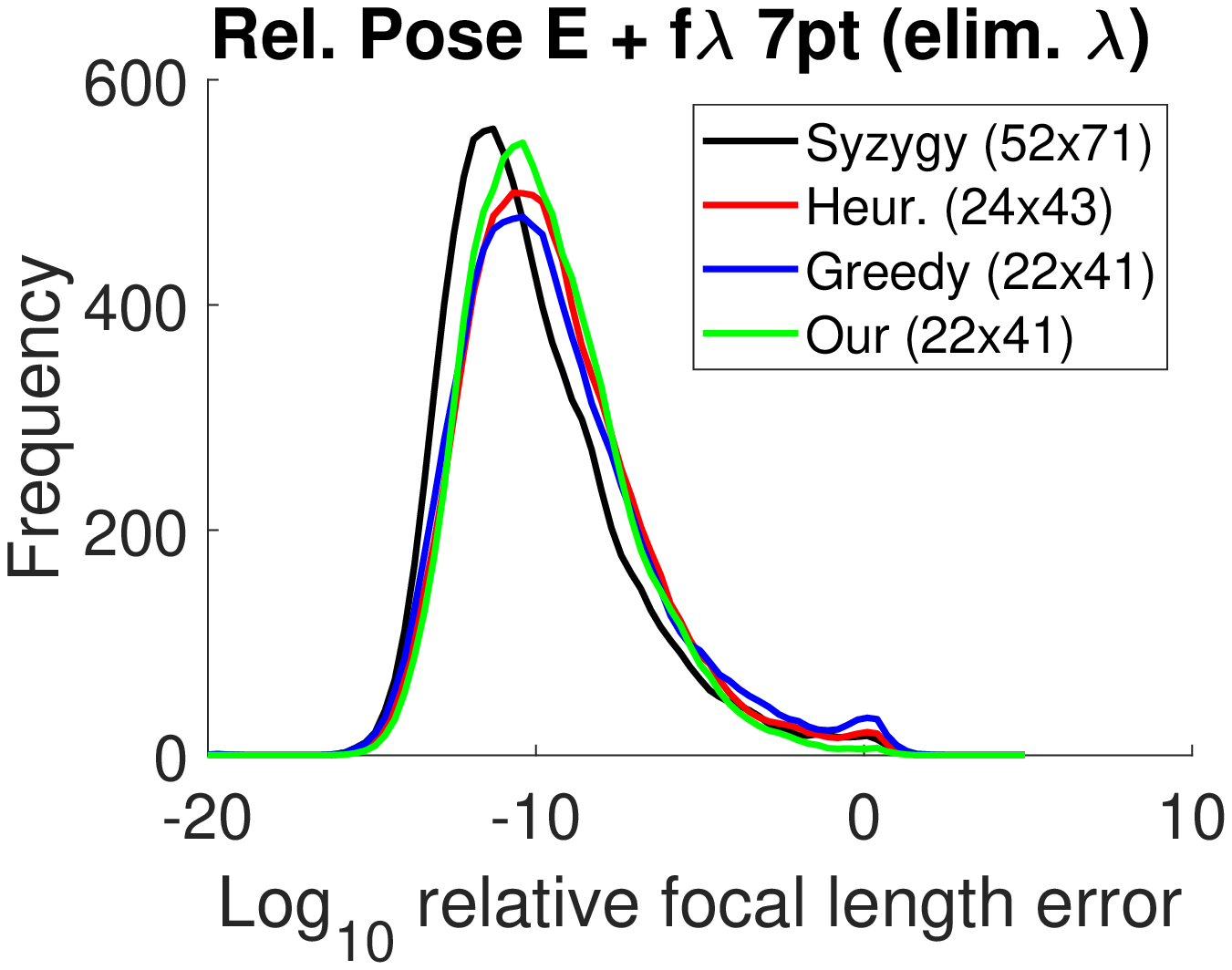}
\caption{Histograms of $Log_{10}$ relative error in radial distortion (\textbf{left}) and focal length (\textbf{right}) for Rel. pose E+$f\lambda$ 7pt (elim $\lambda$) problem 
for 10K randomly generated synthetic scenes. These scenes represent cameras with different radial distortions, poses and focal lengths. For comparison, we generated solvers using our \srs method (green), \gb method~\citep{larsson2017efficient} (black), heuristic method~\citep{DBLP:conf/cvpr/LarssonOAWKP18} (red) and greedy parameter search method~\citep{MartyushevVP2022} (blue).
}
\label{fig:7pt}
\end{figure}

\textbf{E+$f\lambda$ solver on synthetic scenes:} 
We study the numerical stability of our \srs-based solver for the problem of estimating the relative pose of one calibrated camera, and one camera with unknown focal length and radial distortion from 7-point correspondences, i.e.\ the Rel.\ pose E+$f\lambda$ 7pt problem from Tab.~\ref{tbl:sizecomparison}. We consider the formulation ``elim.\ $\lambda$'' proposed in~\citep{DBLP:conf/cvpr/LarssonOAWKP18} that leads to the smallest solvers. We study the performance on noise-free data and compare it to the results of \gb solvers from Tab.~\ref{tbl:sizecomparison}.

We generated 10K scenes with 3D points drawn uniformly from a $\left[-10,10\right]^3$ cube. Each 3D point was projected by two cameras with random feasible orientation and position. The focal length of the first camera was randomly drawn from the interval $f_{gt} \in \left[0.5, 2.5\right]$ and the focal length of the second camera was set to $1$, i.e., the second camera was calibrated. The image points in the first camera were corrupted by radial distortion following the one-parameter division model. The radial distortion parameter $\lambda_{gt}$ was drawn at random from the interval $[-0.7, 0]$ representing distortions of cameras with a small distortion up to slightly more than GoPro-style cameras. 

Fig.~\ref{fig:7pt} shows $Log_{10}$ of the relative error  of the  distortion parameter $\lambda$ (left) and the focal length $f$ (right), obtained by selecting the real root closest to the ground truth. All tested solvers provide stable results with only a small number of runs with larger errors. The solver based on our \srs method (green) is not only smaller, but also slightly more stable than the heuristic-based solver~\citep{DBLP:conf/cvpr/LarssonOAWKP18} (red) and the solver based on~\citep{larsson2017efficient} (black). For most minimal problems, the solver based on our proposed \srs method has fewer failures, as compared to the solver based on the greedy parameter approach in~\citep{MartyushevVP2022} (blue).

\begin{figure}[t]
\centering
\includegraphics[width=0.34\linewidth]{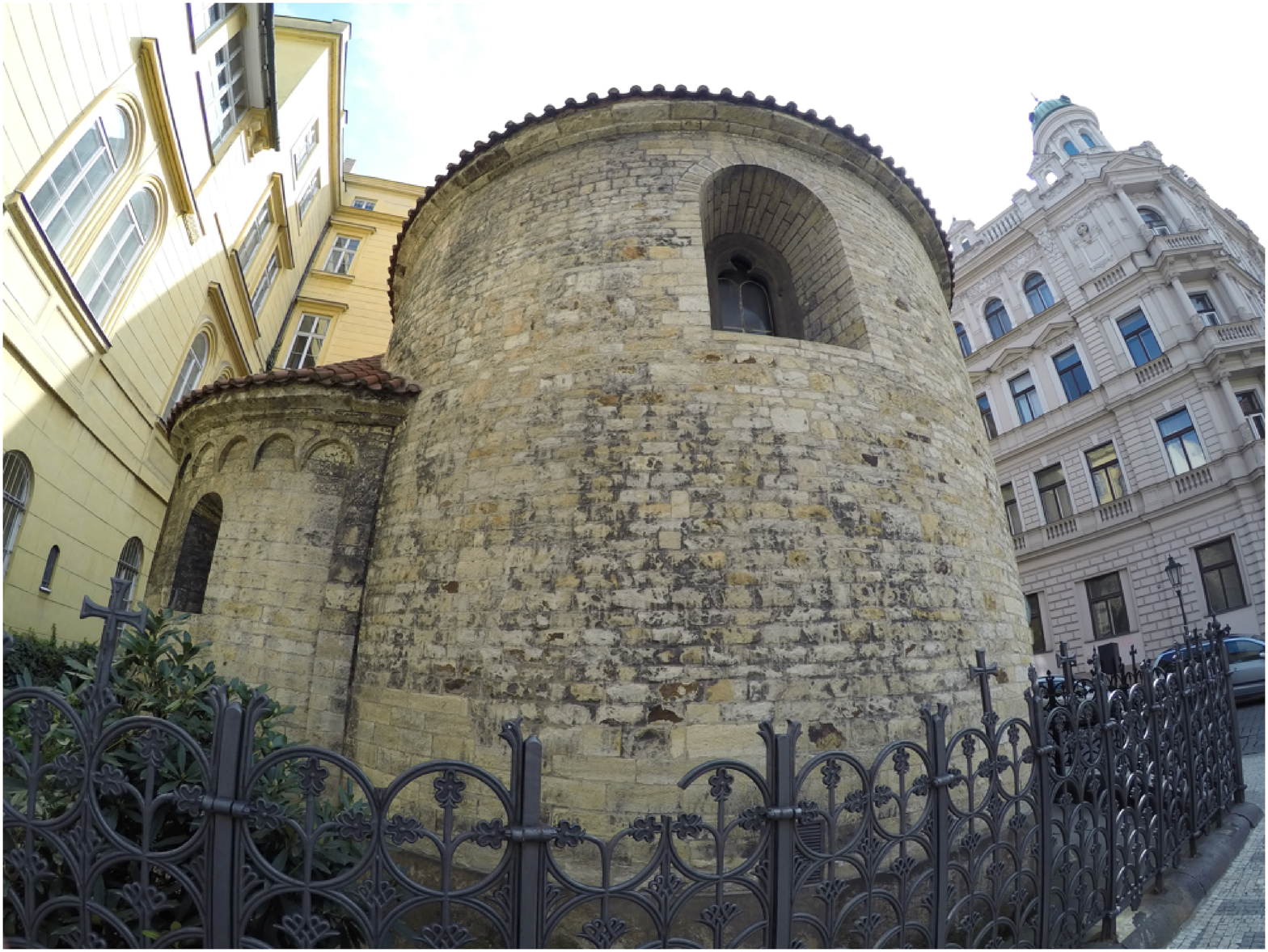}
\includegraphics[width=0.34\linewidth]{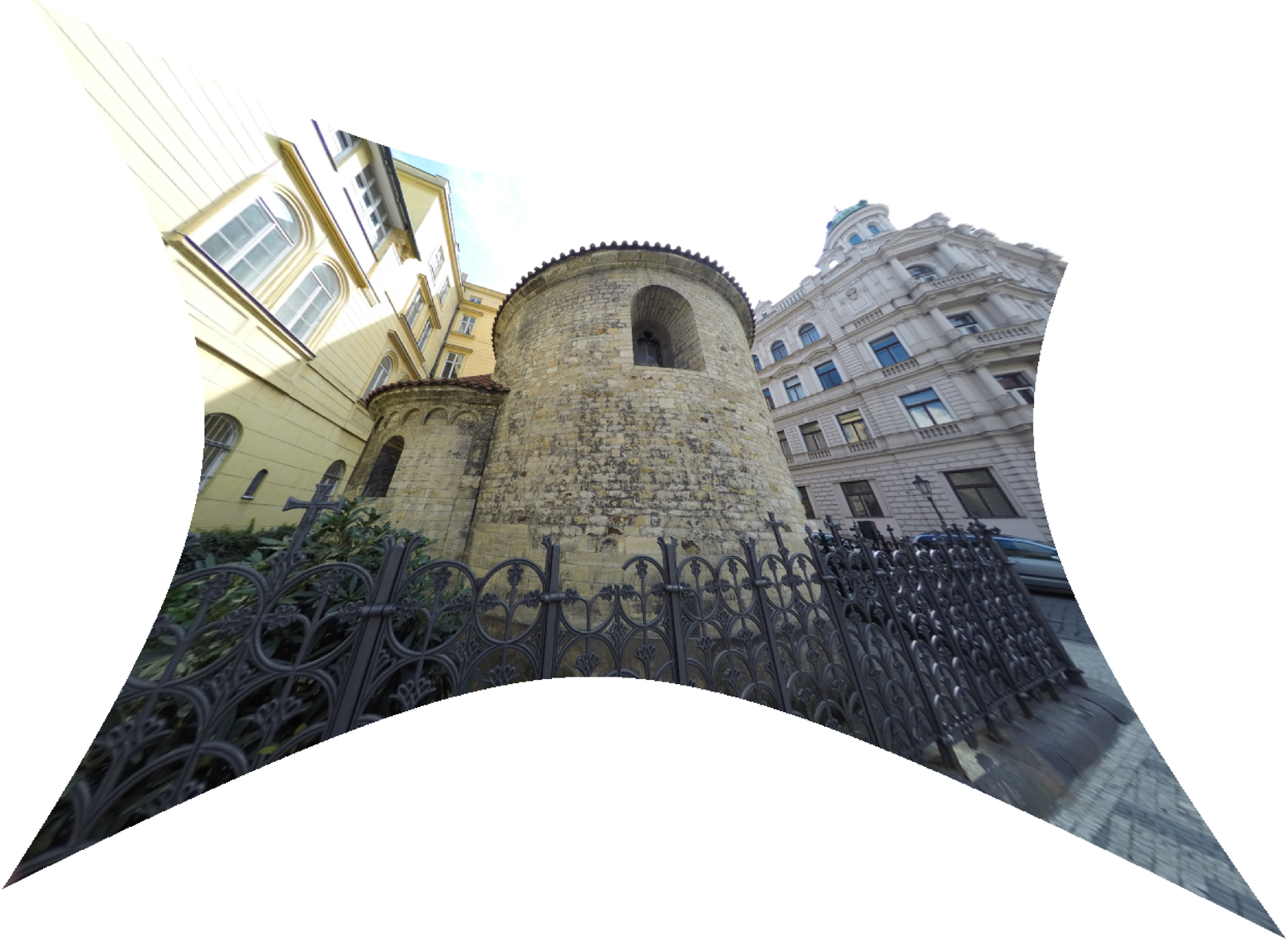}
\includegraphics[width=0.27\linewidth]{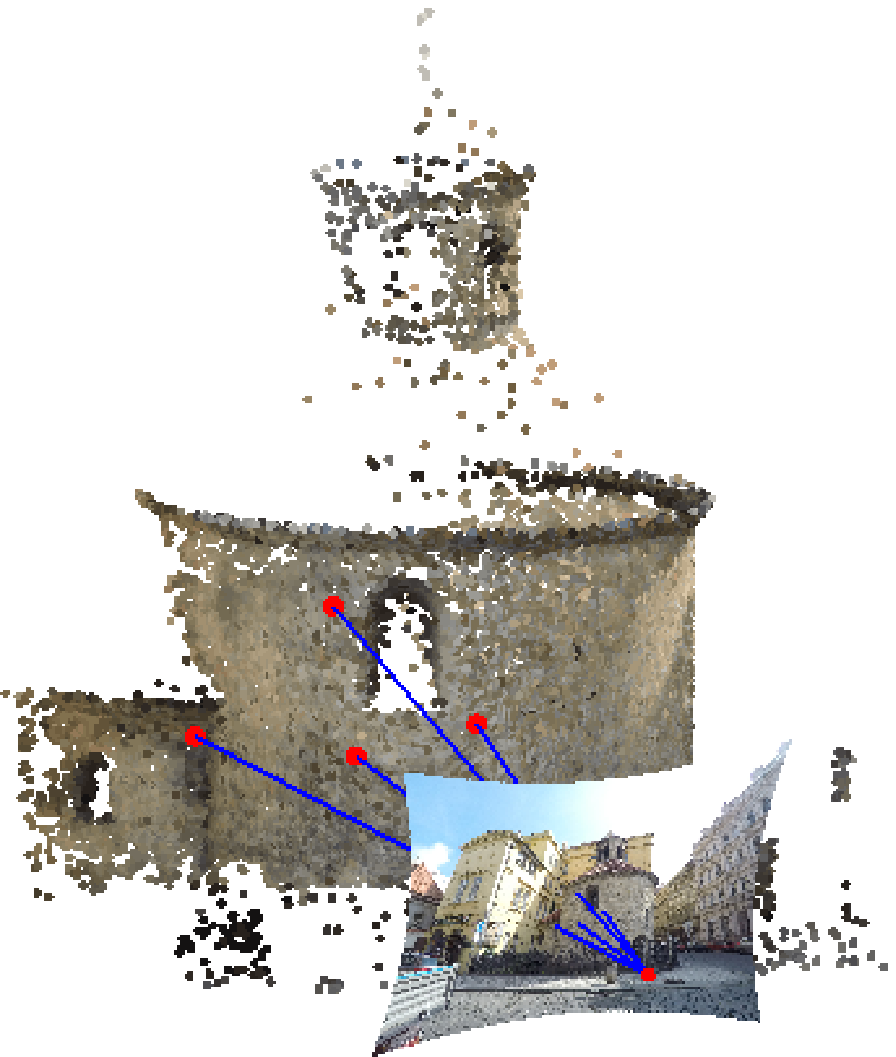} \\
\includegraphics[width=0.440\linewidth]{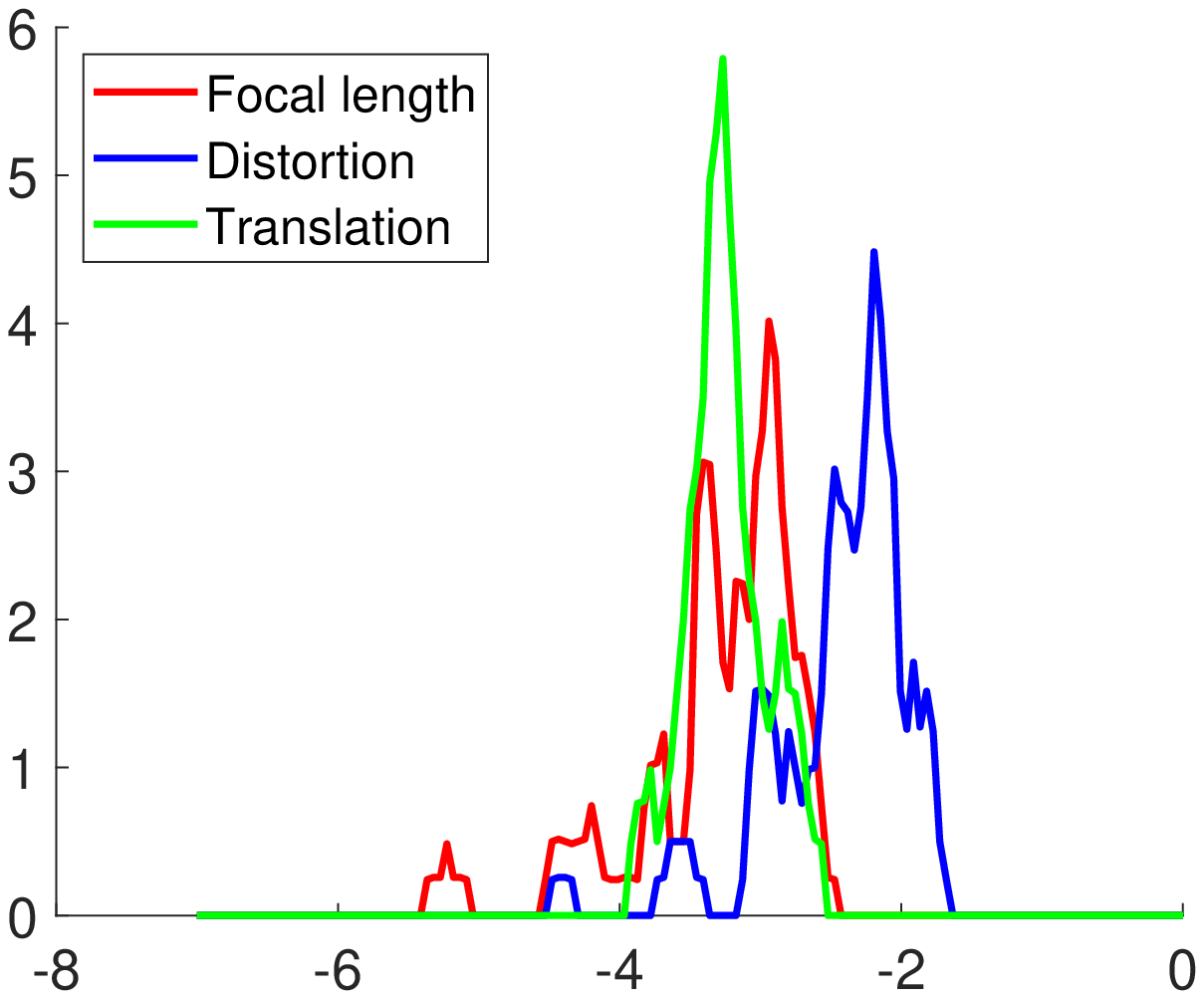}
\includegraphics[width=0.440\linewidth]{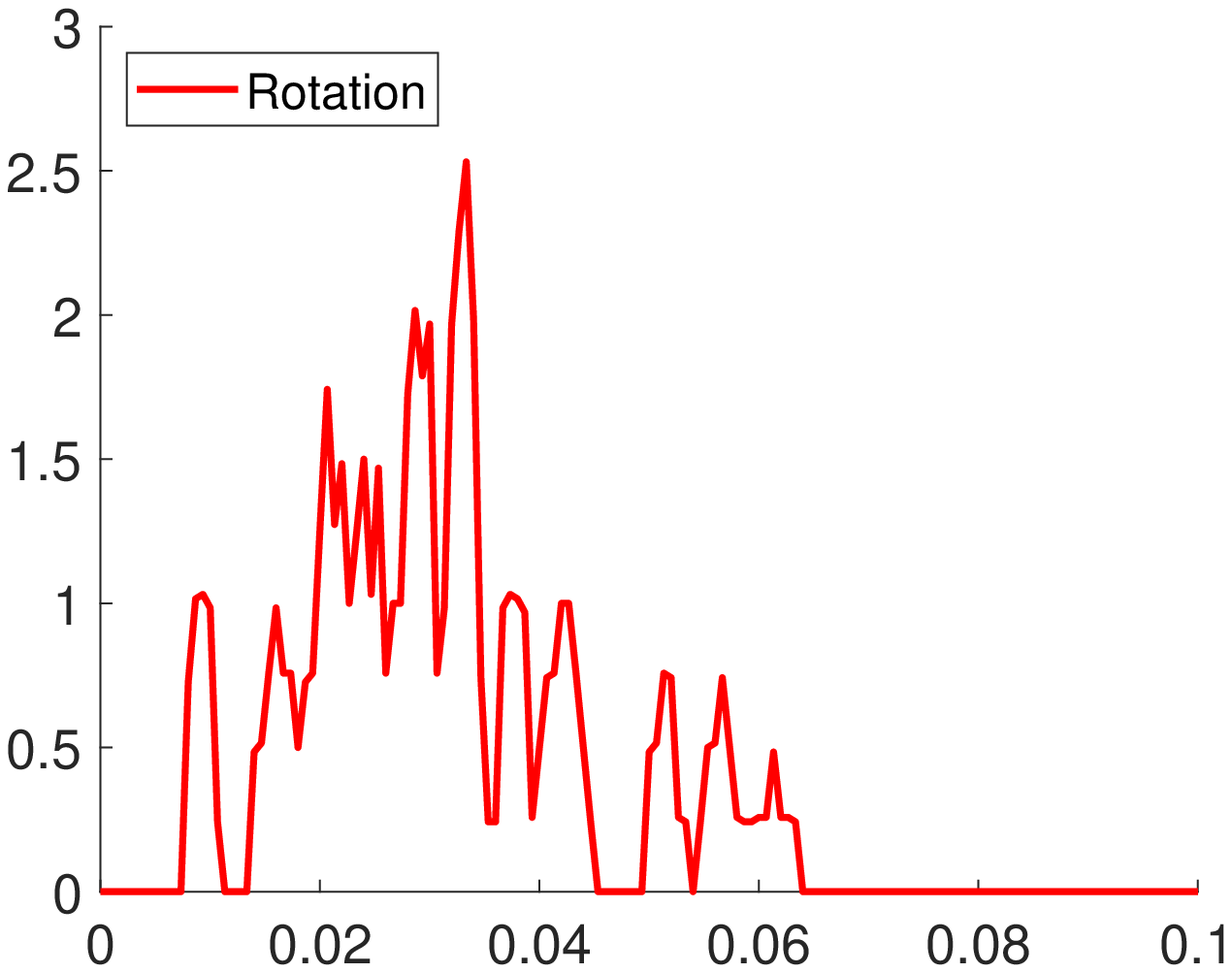}
\caption{Top row: Example of an input image (left). Undistorted image using the proposed resultant-based P4Pfr solver (middle). Input 3D point cloud and an example of registered camera (right).
Bottom row:  Histograms of errors for 62 images. The measured errors are (left) the $Log{10}$ relative focal length $|f-f_{GT}|/f_{GT}$, radial distortion  $|k-k_{GT}|/|k_{GT}|$, and the relative translation error $\| \vec{t}-\vec{t}_{GT} \| / \|\vec{t}_{GT}\|$, and (right) the rotation error in degrees.}
\vspace{-0.2cm}
\label{fig:rotunda}
\end{figure}

\begin{table*}[h!]
\centering
\scalebox{1}{
\begin{tabular}{l c c c | c c c} \toprule
Solver &  \multicolumn{3} {c} {\normalsize Our P4Pfr $28 \times 40$ } &   \multicolumn{3} {c} {\normalsize P4Pfr $40 \times 50$ SOTA} \\ \cmidrule(r){2-4} \cmidrule(r){5-7}
& avg. & med. & max & avg. & med. & max \\ \midrule
Focal (\%) & 0.080 & 0.063 & 0.266 & 0.08 & 0.07 & 0.29     \\
Distortion  (\%)  & 0.522 & 0.453 & 1.651 & 0.51 & 0.45 & 1.85      \\
Rotation  (degree)  &0.031 & 0.029 & 0.062 &0.03 & 0.03 & 0.10    \\
Translation (\%) & 0.066 & 0.051 & 0.210 & 0.07 & 0.07 & 0.26   \\ \bottomrule
\end{tabular}}
~
\caption{Errors for the real Rotunda dataset. The errors are relative to the ground truth for all except rotation which is shown in degrees. The results for the SOTA P4Pfr solver of size $40 \times 50$, are taken from~\citep{larsson2017making}.}
\label{tbl:real_data}
\end{table*}

\textbf{P4Pfr solver on real images:} \label{sec:real_images}
\noindent We evaluated our \srs-based solver for a practical problem of estimating the absolute pose of camera with unknown focal length and radial distortion from four 2D-to-3D point correspondences, i.e. the P4Pfr solver, on real data. We consider the \textit{Rotunda} dataset, which was proposed in \citep{kukelova2015efficient}, and in~\citep{larsson2017making} it was used for evaluating P4Pfr solvers. This dataset consists of 62 images captured by a GoPro Hero4 camera. 
Example of an input image from this dataset (left) as well as undistorted (middle) and registered image (right) using our new solver, is shown in Figure~\ref{fig:rotunda} (top). The Reality Capture software \citep{realitycapture} was used to build a 3D reconstructions of this scene. 
We used the 3D model to estimate the pose of each image using the new P4Pfr resultant-based solver ($28\times 40$) in a RANSAC framework. Similar to \citep{larsson2017making}, we used the camera and distortion parameters obtained from \cite{realitycapture} as ground truth for the experiment.
Figure~\ref{fig:rotunda} (bottom)
shows the errors for the focal length, radial distortion, and the camera pose. Overall, the errors are quite small, \eg~most of the focal lengths are within $0.1\%$ of the ground truth and almost all rotation errors are less than $0.1$ degrees, which shows that our new solver works well for real data. 
We have summarized these results in Tab.~\ref{tbl:real_data} where we present the errors for the focal length, radial distortion, and the camera pose obtained using our proposed solver and for the sake of comparison we also list the errors, which were reported in ~\citep{larsson2017making}, where the P4Pfr (40x50) solver was tested on the same dataset. Overall the errors are quite small, \eg~most of the focal lengths are within $0.1\%$ of the ground truth and almost all rotation errors are less than $0.1$ degrees, which shows that our new solver as well as the original solver work well for real data. The results of both solvers are very similar. However, note that the slightly different results reported in~\citep{larsson2017making} are due to RANSAC's random nature and a slightly different P4Pfr formulation (40x50) used in~\citep{larsson2017making}.

\section{Conclusion}
\noindent In this paper, we have proposed a \srs method for generating efficient solvers for minimal problems in computer vision. It uses a polynomial with a special form to augment the initial polynomial system $\F$ in~\eqref{eq:eq_system} and construct a \srs matrix $\Mres$~\eqref{eq:res_mat_form}, using the theory of polytopes~\citep{Cox-Little-etal-05}. The special form enables us to decompose the resultant matrix constraint~\eqref{eq:res_mat_sing}, into a regular eigenvalue problem~\eqref{eq:eig_prb} (or~\eqref{eq:eig_prb_alt}) using the Schur complement. As demonstrated in the Sec.~\ref{sec:experiments}, our \srs method leads to minimal solvers with comparable speed and stability w.r.t. to the SOTA solvers based on the action matrix method~\citep{Kukelova-thesis,larsson2017efficient,DBLP:conf/cvpr/LarssonOAWKP18,MartyushevVP2022}. Note that the way our \srs method is designed, for some minimal problems, it leads to solvers with larger eigenvalue problems but performing a smaller matrix inverse and with comparable or better stability compared to that of the \gb-based solvers. 

While the action matrix method and the \srs method are based on different mathematical theories, we have observed that the resulting solvers involve similar numerical operations, such as eigenvalue computation. 
This raises the question, ``Under what conditions for a given minimal problem (a given system of polynomial equations), are the two solvers the same?'' In the Sec.~\ref{sec:am_vs_res}, we have attempted to answer this question. Specifically, if we begin with an action matrix-based solver for a given minimal problem, then we propose a list of changes to be made to the steps in our \srs method so that it leads to an equivalent \srs-based solver. In the opposite direction, we also study the case when we begin with a \srs-based solver and establish a list of changes to the steps performed by an action matrix method such that it leads to an equivalent solver. In other words, if we begin with an action matrix-based solver, it determines the extra polynomial $f_{m+1}$ and the favourable monomial set, to be used in the \srs method. Or, if we begin with a \srs-based solver, it determines the monomial ordering, the extended polynomial set $\F^{\prime}$, and the basis of the quotient ring to be used for computing an action matrix-based solver. 

We hope that this discussion paves a path towards unifying the two approaches for generating minimal solvers, \ie, the action matrix methods and the \srs method based on an extra polynomial. It is known that both these approaches are examples of the normal form methods for a given polynomial system $\F$, recently studied in~\citep{TelenM2018,TelenMB2018,MourrainP2008}. To the best of our knowledge, normal forms have yet not been incorporated in the automatic generators for minimal solvers in computer vision. This may be an interesting future topic to investigate.

\backmatter

\bmhead{Acknowledgments}
Zuzana Kukelova was supported by the OP VVV funded project CZ.02.1.01/0.0/0.0/16$\_$019/0000765 “Research Center for Informatics”.



\begin{appendices}

\section{Action matrix}\label{app:am_steps}
Here, we describe the important steps typically performed by an action matrix method based on the theory of the \gb. Note that for a given minimal problem, the \textbf{offline} steps are performed only once, while the \textbf{online} steps are repeated for each instance of the input data/measurements for the given minimal problem.
\begin{enumerate}
    \item \textbf{[Offline]} We begin with a system of $m$ polynomial equations $\F=0$~\eqref{eq:eq_system} in $n$ variables. Let us denote the ideal generated by $\F$, as $I$. Let us also assume that $\F=0$ has $r$ solutions. It is well known that the quotient ring $A = \mathbb{C}[X]/I$ has the structure of a finite-dimensional vector space over the field $\mathbb{C}$ (see \citep{Cox-Little-etal-05} for details). Let $A$ be a $l$-dimensional vector space. Then, if $I$ is a radical ideal, \ie, $I = \sqrt I$, we have $l=r$. 
    
    \item \textbf{[Offline]} This method computes a linear basis of $A$, denoted as $\mathcal{B}_A = \lbrace \left[ \mon{\alpha_1}\right],\dots,\left[\mon{\alpha_r} \right] \rbrace$. Here, \textit{some} monomial ordering is assumed, to define the division of a polynomial $f \in \mathbb{C}[X]$ w.r.t. the ideal $I$. This polynomial division is denoted with the operator $\left[ f \right]$. The set of monomials, corresponding to the elements in $\mathcal{B}_A$, are $B_a = \lbrace \mon{\alpha_1},\dots,\mon{\alpha_r}  \rbrace $.
    \item \textbf{[Offline]} Some $f \in \mathbb{C}[X]$ is assumed as an \textit{action polynomial}, which sends each $\left[ \mon{\alpha} \right] \in A$ to $\left[ f \mon{\alpha} \right] \in A$. We thus have the following linear map
    \begin{equationarray}{l}
     T_f : A \rightarrow A, \ T_f([\mon{\alpha}]) = \left[ f \mon{\alpha} \right].
    \end{equationarray} Fixing a linear basis $\mathcal{B}_{A}$ of $A$, allows us to represent the linear map $T_f$, with an $r \times r$ matrix, $\M{M}_{f} = (m_{ij})$. Thus, for each $ \mon{\alpha_j} \in B_a$, we have
    \begin{equationarray}{l}\label{eq:tf_map}
      T_f ([\mon{\alpha_j}]) = [f\mon{\alpha_j} ] = \sum_{i=1}^r m_{ij}[\mon{\alpha_i}].
    \end{equationarray} In other words, the matrix $\M{M}_f$ represents, what we call a multiplication or an action matrix. Assuming $B_{fa} = \lbrace f\mon{\alpha_j} \mid \mon{\alpha_j} \in B_a \rbrace$, we have 
    \begin{equation}\label{eq:def_action_matrix}
        \M{M}_f \V{b}_a = \V{b}_{fa},
    \end{equation} where $\V{b}_{a} = \text{vec}(B_{a})$ and $\V{b}_{fa} = \text{vec}(B_{fa})$. 
    We can find polynomials $q_j \in I$, such that~\eqref{eq:tf_map} becomes
    \begin{equationarray}{l}\label{eq:am_lin_exp_red_mon}
      f \mon{\alpha_j} =  \sum_{i=1}^r m_{ij} \mon{\alpha_i} + q_j.
    \end{equationarray}
     Note that $q_j \in I \implies q_j = \sum_{i=1}^{m} h_{ij} f_i$. Various action matrix methods in the literature adopt different approaches for computing these polynomials $q_{j} \in \mathbb{C}\left[ X \right]$. Here, we assume the action polynomial $f = x_k$, for some variable $x_k \in X$.
    \item \textbf{[Offline]} In all action matrix methods, we basically need to compute a set $T_j$ of monomial multiples for each of the input polynomials $f_j \in \F$. This gives us an expanded set of polynomials, which we denote as $\F^{\prime} = \lbrace \mon{\alpha} f_i \mid f_i \in \F,  \mon{\alpha} \in T_i \rbrace $. This extended set of polynomials $\F^\prime$, is constructed such that each $q_{j}$ can be computed as a linear combination of the polynomials $\F^\prime$\footnote{The coefficients of each polynomial $q_{j}$ are found through a G-J elimination of the matrix $\M{C}$ representing the extended set of polynomials $\F^\prime$.}.
    \item \textbf{[Offline]} Let $B = \text{mon}(\F^\prime)$, which is partitioned~\citep{byrod-etal-ijcv-2009} as
    \begin{equationarray}{rl}\label{eq:am_B_partn}
     B =& B_e \sqcup B_r \sqcup B_a, \\
     B_r =& \lbrace x_k \mon{\alpha} \mid \mon{\alpha} \in B_a \rbrace \setminus B_a, \\
     B_e =& B \setminus (B_r \sqcup B_a),
    \end{equationarray} where $B_r$ and $B_e$ are respectively, what we call the \textit{reducible} monomials and the \textit{excess} monomials.
    \item \textbf{[Offline]} The set of equations $\F^{\prime}=0$, is expressed in a matrix form, as 
    \begin{equationarray}{l}\label{eq:am_C}
      \M{C} \ \V{b} = \begin{bmatrix}
          \M{C}_{11} & \M{C}_{12} & \M{C}_{13} \\
          \M{C}_{21} & \M{C}_{22} & \M{C}_{23}
      \end{bmatrix} \begin{bmatrix}
          \V{b}_e \\ \V{b}_r \\ \V{b}_a
      \end{bmatrix} = \V{0},
    \end{equationarray} where $\V{b}_e=\text{vec}(B_e)$ and $ \V{b}_r=\text{vec}(B_r)$. The rows of $\M{C}$ are assumed to be partitioned such that $\M{C}_{22}$ is a square invertible matrix. The matrix $\M{C}$ is known as an \textit{elimination template}. 
    As $\M{C}$ represents a coefficient matrix of $\F^\prime$ constructed as described in step $4$, if we perform a G-J elimination of $\M{C}$, we obtain the following 
    \begin{equationarray}{l}
      \begin{bmatrix}
          \M{C}^{\prime}_{11} & \M{0} & \M{C}^{\prime}_{13} \\
          \M{0} & \M{I} & \M{C}^{\prime}_{23}
      \end{bmatrix}.
    \end{equationarray} Note that, $\M{C}_{22}$ is a square invertible matrix. The submatrix $\M{C}^{\prime}_{11}$, may not be square. This implies that some columns in $\begin{bmatrix}
          \M{C}_{11} \\
         \M{C}_{21}
      \end{bmatrix}$ are linearly dependent, which can be removed~\citep{Kukelova-thesis}, along with the corresponding monomials from the set of excess monomials $B_e$. Let the resulting monomial set be denoted as $\hat{B}_e$ and the reduced column block as $\begin{bmatrix}
          \hat{\M{C}}_{11} \\
          \hat{\M{C}}_{21} 
      \end{bmatrix}$.
       Then,~\eqref{eq:am_C} becomes
        \begin{equationarray}{l}\label{eq:am_C_hat}
      \hat{\M{C}} \ \hat{\V{b}} = \begin{bmatrix}
          \hat{\M{C}}_{11} & \M{C}_{12} & \M{C}_{13} \\
          \hat{\M{C}}_{21} & \M{C}_{22} & \M{C}_{23}
      \end{bmatrix} \begin{bmatrix}
          \hat{\V{b}}_e \\ \V{b}_r \\ \V{b}_a
      \end{bmatrix} = \V{0},
    \end{equationarray} where $\hat{\M{C}}$ and $\hat{\V{b}}$ respectively denote the reduced elimination template and the monomial vector. 
       
    \item \textbf{[Online]} A Gauss-Jordan(G-J) elimination\footnote{Usually, G-J elimination is performed through a step of LU or QR factorization~\citep{byrod-etal-ijcv-2009}.} of $\hat{\M{C}}$ leads to
    \begin{equationarray}{l}\label{eq:LU_C}
      \begin{bmatrix}
          \M{I} & \M{0} & \hat{\M{C}}^{\prime}_{13} \\
          \M{0} & \M{I} & \M{C}^{\prime}_{23}
      \end{bmatrix} \begin{bmatrix}
          \hat{\V{b}}_e \\ \V{b}_r \\ \V{b}_a
      \end{bmatrix} = \V{0}.
    \end{equationarray} 
    The lower block row is rewritten as
    \begin{equationarray}{l}
      \begin{bmatrix}\label{eq:LU_C_lower}
          \M{I} &  \M{C}^{\prime}_{23}
      \end{bmatrix} \begin{bmatrix}
          \V{b}_r \\ \V{b}_a
      \end{bmatrix} = \V{0}.
    \end{equationarray}
    The entries of the action matrix $\M{M}_f$ can be then read off from the entries of the matrix $\M{C}^{\prime}_{23}$~\citep{byrod-etal-ijcv-2009,Kukelova-thesis}.
    For $\mon{\alpha_j} \in B_a$, if $x_k \mon{\alpha_j} \in B_a$ for some $\mon{\alpha_{i_1}} \in B_a$, then $j$-th row of $\M{M}_f$ contains $1$ in $i_1$-th column and $0$ in the remaining columns.
    But if $x_k \mon{\alpha_j} \notin B_a$, then $x_k \mon{\alpha_j} \in B_r$ for some $\mon{\alpha_{i_2}} \in B_r$. In this case, $j$-th row of $\M{M}_f$ is the $i_2$-th row of $-\M{C}^\prime_{23}$.
    
    We recover solutions to $\F=0$, from the eigenvalues (and eigenvectors) of the action matrix $\M{M}_{f}$. Specifically, $u_0 \in \mathbb{C}$ is an eigenvalue of the matrix $\M{M}_{f}$, iff $u_0$ is a value of the function $f$ on the variety $V$. We refer to the book~\citep{Cox-Little-etal-05} for further details. In other words, if $f=x_k$, then the eigenvalues of $\M{M}_{f}$ are the $x_{k}$-coordinates of the solutions of~\eqref{eq:eq_system}. The solutions to the remaining variables can be obtained from the eigenvectors of $\M{M}_{f}$. This means that after finding the multiplication matrix $\M{M}_{f}$, we can recover the solutions by its eigendecompostion, for which efficient algorithms exist.
\end{enumerate}

\noindent The output of the offline stage is the elimination template $\hat{\M{C}}$ in step $6$. In the online stage, step $7$, for a given instance of the minimal problem, we fill in the coefficients of the elimination template $\hat{\M{C}}$ using the input data, and perform its G-J elimination to compute the action matrix $\M{M}_f$. Eigendecomposition of $\M{M}_f$, then gives us the solutions to the given instance of the minimal problem.

The first automatic approach for generating elimination templates and \gb solvers was presented in~\citep{Kukelova-ECCV-2008}. Recently, an improvement to the automatic generator~\citep{Kukelova-ECCV-2008} was proposed in~\citep{larsson2017efficient}. It exploits the inherent relations between the input polynomial equations and it results in more efficient solvers than~\citep{Kukelova-ECCV-2008}. The automatic method from~\citep{larsson2017efficient} was later extended by a method for dealing with saturated ideals~\citep{larsson2017polynomial} and a method for detecting symmetries in polynomial systems~\citep{larsson2016uncovering}.

\section{Algorithms}\label{subsec:polytope_extra_poly_alg}
\noindent Here, we provide the algorithms for our \srs method in Sec.~\ref{sec:extra_poly_sparse_res}. 
\subsection{Extracting a favourable set of monomials}
\noindent The Alg.~\ref{alg:favourablebasisextraction} computes a favourable set of monomials $B$ and a block partition of the corresponding coefficient matrix $\M{C}(u_0)$, for an instance of a minimal problem, \ie, a system of $m$ polynomial equations $\F=0$, in $n$ unknown variables $X$~\eqref{eq:eq_system}. The output of the algorithm also contains a set of monomial multiples $T$. 
Our approach for computing a favourable monomial set, is described in Sec.~\ref{subsec:basis_using_polytopes} and our approach for partitioning the coefficient matrix, is described in Sec.~\ref{subsec:block_partn}. In Sec.~\ref{subsec:block_partn}, we have considered two ways of partitioning the favourable monomial set $B$, \ie, as in~\eqref{eq:res_mon_partn} and in~\eqref{eq:res_mon_partnalt}, in order to block partition $\M{C}(u_0)$. However, in Alg.~\ref{alg:favourablebasisextraction}, we consider the first one. For the alternate partition, all steps remain the same, except the step $15$, where we use the alternative partition in~\eqref{eq:res_mon_partnalt}.
\begin{equationarray*}{l}
   B^{\prime}_{1} \gets \lbrace \mon{\alpha} \in  B^\prime  \mid  \dfrac{\mon{\alpha}}{x_i}  \in  T_{m+1} \rbrace$, $B^{\prime}_{2} \gets B^{\prime} \setminus B^{\prime}_{1}.
\end{equationarray*}

\begin{algorithm}[h!]
    \caption{Computing a favourable set of monomials $B$ and block partitioning the coefficient matrix $\M{C}(u_0)$}
    \label{alg:favourablebasisextraction}
    \textbf{Input} $:\F = \lbrace f_{1}(\V{x}),\dots ,f_{m}(\V{x}) \rbrace$, $\V{x} = \left[ x_{1},\dots ,x_{n} \right]$ \\
    \textbf{Output} $:B, T, \M{C}(u_0)$
    \begin{algorithmic}[1]
    \STATE{$B \gets \phi, T \gets \phi$}
    \FOR{$k \in \lbrace 1,\dots ,n \rbrace$}
    \STATE{$ \F_a \gets \lbrace f_{1},\dots ,f_{m+1} \rbrace$, $f_{m+1} = x_{k} - u_0$}
    \STATE{Calculate the support of the input polynomials: \\ $A_{j} \gets \text{supp}(f_{j}), j = 1,\dots ,m+1$}
    \STATE{Construct Newton polytopes: \\ $NP_j \gets \text{conv}(A_{j}), j = 1, \dots, m+1$ as well as a unit simplex $NP_0 \subset \mathbb{Z}^{n}$.}
    \STATE{Enumerate combinations of indices of all possible sizes: \\ $K \gets \lbrace \lbrace k_{0},\dots ,k_{i} \rbrace \mid\! \forall 0 \! \leq\! i \leq (m+1); k_{0},\dots ,k_{i} \in \lbrace 0,\dots,m+1 \rbrace; k_{j} < k_{j+1}  \rbrace$}
    \STATE{Let $\Delta \gets \lbrace \lbrace \delta_{1},\dots ,\delta_{n+1} \rbrace \mid \delta_{i} \in \lbrace -0.1, 0, 0.1 \rbrace; i = 1,\dots ,(n+1) \rbrace$ }
    \FOR{$I \in K$}
        \STATE{Compute the Minkowski sum: $Q \gets \sum_{j \in I} (NP_{j})$}
        \FOR{$\delta \in \Delta$}
            \STATE{$B^{\prime} \gets \lbrace \mon{\alpha} \mid \V{\alpha} \in \mathbb{Z}^{n} \cap (Q + \delta) \rbrace$}
            \STATE{
            $\F_a \overset{B^\prime}{\rightarrow} (\F^\prime_a, T^{\prime})$}
            \STATE{$T^{\prime}$ contains $\lbrace T^{\prime}_{1} \dots T^{\prime}_{m+1} \rbrace$}
            \STATE{Compute $\M{C}(u_0)^{\prime}$ from $B^{\prime}$ and $T^{\prime}$}
            \STATE{$B^{\prime}_{1} \gets \! B^{\prime} \cap T^{\prime}_{m+1}$, $B^{\prime}_{2} \gets B^{\prime} \setminus B^{\prime}_{1}$}
            \IF{$\Sigma_{j=1}^{m+1}|T^{\prime}_{j}| \! \geq \! |B^{\prime}|$ and $\min\limits_{j}|T^{\prime}_{j}| \! > \! 0$ and $\textit{rank}(\M{C}(u_0)^{\prime}) \! = \! |B^{\prime}|$}
                \STATE{$\M{A_{12}} \gets$ submatrix of $\M{C}(u_0)^{\prime}$ column-indexed by $B^{\prime}_{2}$ and row-indexed by $T^{\prime}_{1} \cup \dots \cup T^{\prime}_{m}$}
                \IF{$\text{rank}(\M{A_{12}}) = |B^{\prime}_{2}|$ and $ | B | \geq |B^{\prime}|$}
                        \STATE{$B \gets B^{\prime}, T \gets T^{\prime}$}
                \ENDIF
            \ENDIF
        \ENDFOR
    \ENDFOR
    \ENDFOR
    \STATE{Compute $\M{C}(u_0)$ from $B$ and $T$}
    \end{algorithmic}
\end{algorithm}

\subsection{Row-column removal}\label{subsec:row_col_removal}
\noindent The next step in the proposed method is to reduce the favourable monomial set $B$, by removing columns from $\M{C}(u_0)$ along with a corresponding set of rows, described in Sec.~\ref{subsec:col_red}. The Alg.~\ref{alg:basisred} achieves this. Its input is the favourable monomial set $B$, the corresponding set of monomial multiples $T$, computed by Alg.~\ref{alg:favourablebasisextraction} and the output is a reduced monomial set $B_{\text{red}}$ and a reduced set of monomial multiples, $T_{\text{red}}$ that index the columns and rows of the reduced matrix $\M{C}_{\text{red}}(u_0)$ respectively. We note that this algorithm is the same irrespective of the version of partition of the monomial set $B$, \ie~\eqref{eq:res_mon_partn} or~\eqref{eq:res_mon_partnalt}.
\begin{algorithm}[h!]
    \caption{Row-column removal}
    \label{alg:basisred}
    \textbf{Input} $:B, T$ \\
    \textbf{Output} $:B_{\text{red}}, T_{\text{red}},\M{C}_{\text{red}}(u_0)$\\
    \begin{algorithmic}[1]
        \STATE{$B^{\prime} \gets B, T^{\prime} \gets T$}
        \REPEAT
        \STATE{$\text{stopflag} \gets$ TRUE}
        \STATE{Compute $\M{C}(u_0)^{\prime}$ from $B^{\prime}$ and $T^{\prime}$}
        \FOR{column $c$ in $\M{C}(u_0)^{\prime}$}
            \STATE{Copy $\M{C}(u_0)^{\prime}$ to $\M{C}(u_0)^{\prime \prime}$}
            \STATE{Remove rows $r_{1},\dots,r_{s}$ containing $c$ from $\M{C}(u_0)^{\prime \prime}$}
            \STATE{Remove columns $c_{1},\dots,c_{l}$ of $\M{C}(u_0)^{\prime \prime}$ present in $r_{1},\dots,r_{s}$}
            \IF{$\M{C}(u_0)^{\prime \prime}$ satisfies Prop.~\ref{prop:eigendecomp}}
                \STATE{Remove monomials from $B^{\prime}$ indexing columns $c_{1},\dots,c_{l}$ }
                \STATE{Remove monomials from $T^{\prime}$ indexing rows $r_{1},\dots,r_{s}$}
                \STATE{$\text{stopflag} \gets$ FALSE}
                \STATE{\textbf{break}}
            \ENDIF
        \ENDFOR
        \UNTIL{$\text{stopflag}$ is TRUE}
        \STATE{$B_{\text{red}} \gets B^{\prime}, T_{\text{red}} \gets T^{\prime}$}
        \STATE{Compute $\M{C}_{\text{red}}(u_0)$ from $B_{\text{red}}$ and $T_{\text{red}}$}
    \end{algorithmic}
    \end{algorithm}

\subsection{Row removal}\label{subsubsec:rowremoval}
\noindent It may happen that the reduced matrix $\M{C}_{\text{red}}(u_0)$ still has more rows than columns. Therefore, we propose an approach to remove the excess rows from $\M{C}_{\text{red}}(u_0)$, to transform it into a square matrix, which will be our \srs matrix $\Mres$. Towards this, we provide Alg.~\ref{alg:excessrowremoval} to remove the extra rows from $\M{C}_{\text{red}}(u_0)$ by removing some monomial multiples from $T_{\text{red}}$. It accepts the favourable monomial set $B_{\text{red}}$ and its corresponding monomial multiples $T_{\text{red}}$, as input and returns a reduced set of monomial multiples, $T_{\text{red}}$ such that, along with the basis $B_{\text{red}}$, leads to a square matrix. If we partitioned $B_{\text{red}}$ using the alternative approach~\eqref{eq:res_mon_partnalt}, we just need to change step $1$7 in Alg.~\ref{alg:excessrowremoval} to
\begin{equationarray*}{l}
   B^{\prime}_{1} \gets \lbrace \mon{\alpha} \! \in\!  B^\prime  \mid  \dfrac{ \mon{\alpha}}{x_i}  \in  T_{m+1} \rbrace$, $B^{\prime}_{2} \gets B^{\prime} \setminus B^{\prime}_{1}.
\end{equationarray*}

\begin{algorithm}[h!]
    \caption{Removal of the extra rows}
    \label{alg:excessrowremoval}
    \textbf{Input} $B_{\text{red}}, T_{\text{red}}$ \\
    \textbf{Output} $T_{\text{red}}, \Mres$ 
    \begin{algorithmic}[1]
    \STATE{$T_{\text{red}}$ contains $\lbrace T^{\prime}_{1},\dots ,T^{\prime}_{m+1} \rbrace$}
    \STATE{$B_{N} \gets |B_{\text{red}}|, T_{N} \gets \Sigma_{j=1}^{m+1}|T^{\prime}_{j}|, t_{\text{chk}} \gets \phi$}
    \WHILE{$T_{N} > B_{N}$}
        \STATE{$B^{\prime} \gets B_{\text{red}}, T^{\prime} \gets T_{\text{red}}$}
        \STATE{$T^{\prime} $ contains $\lbrace T^{\prime}_{1},\dots ,T^{\prime}_{m+1} \rbrace$}
        \STATE{Randomly select $t \in\! \lbrace t_{m} \in T^{\prime}_{m+1}\! \mid\! (t_m,m+1) \notin t_{\text{chk}} \rbrace$}
        \IF{$t$}
            \STATE{$T_{m+1}^{\prime} \gets T^{\prime}_{m+1} \setminus \lbrace t \rbrace$} \STATE{$T^{\prime} \gets  \lbrace T^{\prime}_{1},\dots, T_{m+1}^{\prime} \rbrace$}
            \STATE{$t_{\text{chk}} \gets t_{\text{chk}} \cup \lbrace (t,m+1) \rbrace$}
        \ELSE
            \STATE{Randomly select $i \in \lbrace 1,\dots ,m \rbrace$}
            \STATE{Randomly select $t \in  \lbrace t_{i} \in T^{\prime}_{i} \mid (t_i,i) \notin t_{\text{chk}} \rbrace$}
            \STATE{$T_{i}^{\prime} \gets T^{\prime}_{i} \setminus \lbrace t \rbrace, T^{\prime} \gets  \lbrace T^{\prime}_{1},\dots, T^{\prime}_{m+1} \rbrace$}
            \STATE{$t_{\text{chk}} \gets t_{\text{chk}} \cup \lbrace (t,i) \rbrace$}
        \ENDIF
        \STATE{$B_1^{\prime} \gets B^{\prime} \cap T^{\prime}_{m+1}$, $B_2^{\prime} \gets B^{\prime} \setminus B_1^{\prime}$}
        \STATE{Compute $\M{C}(u_0)^{\prime}$ from $B^{\prime}$ and $T^{\prime}$}
        \IF{$\M{C}(u_0)^{\prime}$ satisfies Prop.~\ref{prop:eigendecomp}}
        \STATE{$T_{\text{red}} \gets T^{\prime}, T_{N} \gets \Sigma^{m+1}_{j= 1} |T^{\prime}_{j}|$}
    \ENDIF
\ENDWHILE
\STATE{Compute $\Mres$ from $B_{\text{red}}$ and $T_{\text{red}}$}
\end{algorithmic}
\end{algorithm}

\section{Proofs}
In this appendix, we provide proofs for the propositions discussed in Sec.~\ref{sec:am_vs_res}, regarding the equivalence of an action matrix-based solver with a \srs-based solver, in three situations.
   \begin{proposition}\label{prop:X_frm_Mf}
     For a given system of polynomial equations $\F=0$~\eqref{eq:eq_system}, let us assume to have generated an action matrix-based solver using the method in Sec.~\ref{subsec:action_matrix_outline}. The action matrix-based solver is assumed to be generated for the action variable, $f=x_1$.  Also, let us consider a \srs-based solver, generated after applying the changes C1R-C6R in Sec.~\ref{subsec:res_frm_actn_mat}, to the steps $1$-$5$ in Sec.~\ref{subsec:sparse_res_outline}, \ie~the offline stage. Then, the two solvers are equivalent, as defined in Def.~\ref{def:solver_equiv}.
    \end{proposition}
    \textit{Proof:}
    Note from~\eqref{eq:invertible_left_block_C}, that the elimination template can be written as
    \begin{equationarray}{l}
        \prfx{a}\hat{\M{C}} = \begin{bmatrix}
    \hat{\M{C}}_{11} & \M{C}_{12} & \M{C}_{13} \\
    \hat{\M{C}}_{21} & \M{C}_{22} & \M{C}_{23}  
    \end{bmatrix} = \begin{bmatrix}
    \hat{\M{A}}_{12} & \hat{\M{A}}_{11} \end{bmatrix}.
    \end{equationarray} Therefore, G-J elimination of $\prfx{a}\hat{\M{C}}$ can be considered as computing the matrix product, $\hat{\M{A}}_{12}^{-1} \ \prfx{a}\hat{\M{C}} = \begin{bmatrix}
    \M{I} & \hat{\M{A}}_{12}^{-1} \hat{\M{A}}_{11} \end{bmatrix}$. Comparing it with the G-J eliminated form of the matrix $\prfx{a}\hat{\M{C}}$ in~\eqref{eq:LU_C}, we have \begin{equationarray}{rl}\label{eq:gj=inv}
                \hat{\M{A}}_{12}^{-1} \hat{\M{A}}_{11} &=    \begin{bmatrix} \hat{\M{C}}^\prime_{13} \\ \M{C}^\prime_{23} \end{bmatrix}. 
               \end{equationarray}
               
    \noindent 
        Note that for each $\mon{\alpha_j}$ in $\prfx{r}T_{m+1} (\!=\! B_1)$,
        the $j$-th row of $\begin{bmatrix} \M{A}_{21} - u_0 \M{I} & \M{A}_{22} \end{bmatrix} \prfx{r}\V{b}$~\eqref{eq:des_C_partn_repeated} represents the multiple $\mon{\alpha_j} f_{m+1} = \mon{\alpha_j} x_1  - \mon{\alpha_j} u_0$.
        Then we have the following two cases, for each $\mon{\alpha_j} \in B_1$:
    \begin{enumerate}[label=Case \arabic*:]
        \item If $x_1 \mon{\alpha_j} = \mon{\alpha_{i_1}} \in B_1$, the $j$-th row of $\M{A}_{22}$ is $\V{0}$. The $j$-th row of $\M{A}_{21}$, and hence that of $\M{X}$ has $1$ in $i_1$-th column and $0$ in the remaining columns. Moreover, in this case, the $j$-th row of the action matrix $\M{M}_f$ should also have $1$ in $i_1$-th column and $0$ in the remaining columns, (see step $7$ of the action matrix-based method in Sec.~\ref{subsec:action_matrix_outline}).
        
        \item If $x_1 \mon{\alpha_j} \notin B_1$, then $x_1 \mon{\alpha_j} =  \mon{\alpha_{i_2}}$, for some $ \mon{\alpha_{i_2}} \in B_2$. Then the $j$-th row of $\M{A}_{21}$ is $\V{0}$, and $j$-th row of $\M{A}_{22}$ has $1$ in $i_2$-th column and $0$ in the remaining columns. Therefore, the $j$-th row of the matrix product $-\M{A}_{22} \hat{\M{A}}_{12}^{-1} \hat{\M{A}}_{11}$, and hence that of the matrix $\M{X}$, is actually the $i_2$-th row of $-\hat{\M{A}}_{12}^{-1} \hat{\M{A}}_{11}$. From~\eqref{eq:gj=inv}, this is actually a row from the lower block\footnote{$i_2$-th row of $\begin{bmatrix} \hat{\M{C}}^\prime_{13} \\ \M{C}^\prime_{23} \end{bmatrix}$ will always be a row from the lower block, indexed by the monomials in $\V{b}_r$. This is because $x_1 \mon{\alpha_j} =  \mon{\alpha_{i_2}} \in B_2 \implies \mon{\alpha_{i_2}} \in B_r$, and from the change C4R, $B_2 = \hat{B}_e \sqcup B_r$.}, $-\M{C}^\prime_{23}$.
        However, from the discussion in the step $7$ in Sec.~\ref{subsec:action_matrix_outline}, this is the same as the $j$-th row of the action matrix $\M{M}_f$.
    \end{enumerate} Thus in both cases, the rows of the matrices $\M{M}_f$ and $\M{X}$ are the same and therefore, $\M{X}=\M{M}_f$. Moreover,~\eqref{eq:gj=inv} implies that computing the matrix product $\hat{\M{A}}_{12}^{-1} \hat{\M{A}}_{11}$ can be replaced by the step of G-J elimination of the matrix $\prfx{a}\hat{\M{C}}$. Therefore, as both the conditions in the Def.~\ref{def:solver_equiv} are satisfied, the action matrix-based solver is equivalent to the \srs-based solver.
    $\qedsymbol$

    \begin{proposition}\label{prop:Mf_frm_X}
     For a given system of polynomial equations $\F=0$~\eqref{eq:eq_system}, let us assume to have generated a \srs-based solver. Let us also consider an action matrix-based solver, generated after applying the changes C1A-C6A in Sec.~\ref{subsec:Mf_frm_X} to the offline stage, \ie~steps $1$-$6$ in Sec.~\ref{subsec:action_matrix_outline}. Then the two solvers are equivalent as defined in the Def.~\ref{def:solver_equiv}.
\end{proposition}
\textit{Proof:}
From~\eqref{eq:C_frm_M}, note that the elimination template $\prfx{a}\hat{\M{C}}$ can be expressed as
$\prfx{a}\hat{\M{C}} = \begin{bmatrix}
\hat{\M{C}}_{11} & \M{C}_{12} & \M{C}_{13} \\
\hat{\M{C}}_{21} & \M{C}_{22} & \M{C}_{23} \end{bmatrix} = 
\begin{bmatrix}
\hat{\M{A}}_{12} & \hat{\M{A}}_{11} \end{bmatrix} $. Therefore a G-J elimination of $\prfx{a}\hat{\M{C}}$ can be achieved by computing the matrix product  $\hat{\M{A}}_{12}^{-1} \ \prfx{a}\hat{\M{C}} = \begin{bmatrix}
\M{I} & \hat{\M{A}}_{12}^{-1} \hat{\M{A}}_{11} \end{bmatrix}$. Comparing it with the G-J eliminated form of the matrix $\prfx{a}\hat{\M{C}}$ in~\eqref{eq:LU_C}, we can write \begin{equationarray}{rl}\label{eq:inv=gj}
    \begin{bmatrix} \hat{\M{C}}^\prime_{13} \\ \M{C}^\prime_{23} \end{bmatrix} &= \hat{\M{A}}_{12}^{-1} \hat{\M{A}}_{11}.
\end{equationarray}
This is exactly the same situation, as in the Prop.~\eqref{prop:X_frm_Mf}. Specifically, the Schur complement $\M{X}$, and the action matrix $\M{M}_f$, are exactly the same matrices, with each row, either being a row from $-\M{C}^{\prime}_{23}$ or a row containing $1$'s or $0$'s.
Moreover~\eqref{eq:inv=gj} implies that  G-J elimination of $\prfx{a}\hat{\M{C}}$ can be replaced by the matrix product, $\hat{\M{A}}_{12}^{-1} \hat{\M{A}}_{11}$.
Therefore, as both the conditions in the Def.~\ref{def:solver_equiv} are satisfied, the action matrix-based solver is equivalent to the \srs-based solver.
$\qedsymbol$

\begin{proposition}\label{prop:Mf_frm_X_alt}
        For a given system of polynomial equations $\F=0$~\eqref{eq:eq_system}, let us assume to have generated a \srs-based solver using the alternative partition of $\prfx{r}\!B$~\eqref{eq:des_mon_partnalt} and $\prfx{r}\M{C}(u_0)$~\eqref{eq:des_C_partnalt}. Let us also consider an action matrix-based solver, generated after applying the changes C1A$^\prime$-C6A$^\prime$ in Sec.~\ref{subsec:Mf_frm_Xalt} to the steps $1$-$6$ in Sec.~\ref{subsec:action_matrix_outline}, \ie~the offline stage. Then the two solvers are equivalent as defined in the Def.~\ref{def:solver_equiv}.
    \end{proposition}
    \textit{Proof:}
        In the step C4A$^\prime$, the monomial multiples are set to be the same as those in the \srs method, \ie~$\prfx{a}T_i = \prfx{r}T_i, i=1,\dots,m+1$. Therefore, the upper block of the elimination template $\prfx{a}\M{C}$ in~\eqref{eq:am_C}, is fixed by setting
        \begin{equationarray}{rlrlrl}
        \M{C}_{11} =& \hat{\M{A}}_{12}, & \M{C}_{12} =& \M{0}, & \M{C}_{13} =& \hat{\M{A}}_{11},
        \end{equationarray} where $\begin{bmatrix} \hat{\M{A}}_{11} & \hat{\M{A}}_{12} \end{bmatrix}$~\eqref{eq:des_C_partnalt} is the upper block of the \srs matrix $\Mres$. 
        
        Note that each monomial $\mon{\alpha_j}$ in $\prfx{a}T_{m+1}$ provides the $j$-th row in the lower block of $\prfx{a}\M{C}$, as a vector form of the polynomial $\mon{\alpha_j}  \  \prfx{a}\!f_{m+1}=\mon{\alpha_j} (x_1 \lambda-1)$. Similarly, each monomial $\mon{\alpha_j}$ in $\prfx{r}T_{m+1}$ provides the $j$-th row in the lower block of $\prfx{r}\M{C}(u_0)$ in~\eqref{eq:des_C_partnalt}, as a vector form of the polynomial $\mon{\alpha_j}  \  \prfx{r}\!f_{m+1}=\mon{\alpha_j} (x_1 -u_0)$.
        The lower block of $\prfx{a}\M{C}$ in~\eqref{eq:am_C} is of the form $\begin{bmatrix} \M{C}_{21} & \M{I} & \M{C}_{23} \end{bmatrix}$, and the lower block of $\prfx{r}\M{C}(u_0)$ in~\eqref{eq:des_C_partnalt} is of the form $\begin{bmatrix} \M{I}+u_0 \M{B}_{21} & u_0 \M{B}_{22} \end{bmatrix}$. 
        Note that the columns of $\M{C}_{21}$ and $\M{C}_{23}$ are respectively indexed by $\V{b}_e$ and $\V{b}_a$, whereas the columns of $\M{B}_{21}$ and $\M{B}_{22}$ are respectively indexed by $\V{b}_1$ and $\V{b}_2$. From the change C6A$^\prime$, note that $\V{b}_a=\V{b}_1$, $\V{b}_e=\V{b}_2$ and $\prfx{a}T_{m+1}=\prfx{r}T_{m+1}$. Therefore, $\M{C}_{21} = \M{B}_{22}$ and $\M{C}_{23} = \M{B}_{21}$.

        Substituting the upper and the lower blocks of $\prfx{a}\M{C}$, in~\eqref{eq:am_C_hat}, we obtain
        \begin{equationarray}{l}\label{eq:am_C_from_X_alt}
            \prfx{a}\M{C} \ \prfx{a}\V{b} = \begin{bmatrix}
                  \hat{\M{A}}_{12} & \M{0} & \hat{\M{A}}_{11} \\
                  \M{B}_{22} & \M{I} & \M{B}_{21}
              \end{bmatrix} \begin{bmatrix}
                  \V{b}_e \\ \V{b}_r \\ \V{b}_a
              \end{bmatrix} = \V{0}.
        \end{equationarray} Note that, in the \srs approach, the matrix $ \hat{\M{A}}_{12}$ has to be square invertible. Therefore the matrix $\begin{bmatrix}
                  \hat{\M{A}}_{12} & \M{0} \\
                  \M{B}_{22} & \M{I}
              \end{bmatrix}$ is also invertible and there are no extra columns to be removed in the step $6$ in  Sec.~\ref{subsec:action_matrix_outline}. This means that we have $\prfx{a}\hat{\M{C}} = \prfx{a}\M{C}$, $\prfx{a}\hat{\V{b}}=\prfx{a}\V{b},\prfx{a}\hat{\V{b}}_e=\prfx{a}\V{b}_e, \hat{\M{C}}_{11} = \hat{\M{A}}_{12}$ and $\hat{\M{C}}_{21} = \M{B}_{22}$, in~\eqref{eq:am_C_hat}. Note that \begin{equationarray}{rl}
            \begin{bmatrix}
                  \hat{\M{A}}_{12} & \M{0} \\
                  \M{B}_{22} & \M{I}
              \end{bmatrix}^{-1} =& \begin{bmatrix}
                  \hat{\M{A}}^{-1}_{12} & \M{0} \\
                  -\M{B}_{22}\hat{\M{A}}^{-1}_{12} & \M{I}
              \end{bmatrix}.
        \end{equationarray}
        Therefore, a G-J elimination of the matrix $\prfx{a}\hat{\M{C}}$ is of the form \begin{equationarray}{l}\label{eq:inv=gj_alt}
             \begin{bmatrix}
              \M{I} & \M{0} & \hat{\M{A}}^{-1}_{12}\hat{\M{A}}_{11} \\
              \M{0} & \M{I} & \M{B}_{21} - \M{B}_{22} \hat{\M{A}}_{12}^{-1} \hat{\M{A}}_{11}
              \end{bmatrix}.
        \end{equationarray} Comparing it with the G-J eliminated form of the matrix $\prfx{a}\hat{\M{C}}$ in~\eqref{eq:LU_C}, the submatrix $\M{C}_{23}^\prime$ is
        \begin{equationarray}{l}
            \M{C}_{23}^\prime =\M{B}_{21} - \M{B}_{22} \hat{\M{A}}_{12}^{-1} \hat{\M{A}}_{11}.
        \end{equationarray}
         From the lower block~\eqref{eq:LU_C_lower}, we have  $\V{b}_r = -\M{C}_{23}^\prime \V{b}_a$. In this case $B_r \cap B_a = \emptyset$, which means that $-\M{C}_{23}^\prime$ is the action matrix $\M{M}_f$~\eqref{eq:def_action_matrix}, for $f=\lambda$.
         However, from~\eqref{eq:eig_prb_alt}, $\M{X}=\M{B}_{21} - \M{B}_{22} \hat{\M{A}}_{12}^{-1} \hat{\M{A}}_{11}$. Thus, the action matrix $\M{M}_f$ is equal to the Schur complement $\M{X}$. 
         Moreover from~\eqref{eq:inv=gj_alt}, it can be seen that a step of G-J elimination of the matrix $\prfx{a}\hat{\M{C}}$ can be replaced by the step of computing the matrix product $\hat{\M{A}}_{12}^{-1} \hat{\M{A}}_{11}$. Therefore, as both the conditions in the Def.~\ref{def:solver_equiv} are satisfied, the action matrix-based solver is equivalent to the \srs-based solver.
        $\qedsymbol$
\end{appendices}

\bibliography{manuscript}
\end{document}